\documentclass{article}
\usepackage{multirow}

\usepackage{microtype}
\usepackage{graphicx}
\usepackage{subcaption}
\usepackage{booktabs} 
\usepackage{xcolor}      
\usepackage{colortbl}    

\usepackage{hyperref}

\usepackage[accepted]{icml2026}

\usepackage{amsmath}
\usepackage{amssymb}
\usepackage{mathtools}
\usepackage{amsthm}

\theoremstyle{plain}

\theoremstyle{definition}

\theoremstyle{remark}

\icmltitlerunning{GENEB: Why Genomic Models Are Hard to Compare}

\begin{document}

\twocolumn[
  \icmltitle{GENEB: Why Genomic Models Are Hard to Compare}




  \begin{icmlauthorlist}
    \icmlauthor{Daria Ledneva}{yyy}
    \icmlauthor{Mikhail Nuridinov}{comp}
    \icmlauthor{Denis Kuznetsov}{yyy}
  \end{icmlauthorlist}

  \icmlaffiliation{yyy}{Moscow Independent Research Institute of Artificial Intelligence, Moscow,
Russia}
  \icmlaffiliation{comp}{
Moscow State Institute of Steel and Alloys, Moscow, Russia}

  \icmlcorrespondingauthor{Daria Ledneva}{a.ledn2026@gmail.com}

  \icmlkeywords{Machine Learning, ICML}

  \vskip 0.3in
]



\printAffiliationsAndNotice{}  

\begin{abstract}
Progress in genomic foundation models is difficult to assess due to fragmented benchmarks, incompatible evaluation protocols, and task-specific reporting. As a result, claims of superiority or generality across models are often not directly comparable. We introduce GENEB, a large-scale diagnostic benchmark that evaluates frozen representations from 40 genomic foundation models across 100 tasks spanning 13 functional categories under a unified probing-based protocol, including few-shot regimes. GENEB enables controlled comparison across model scale, architecture, tokenization, and pretraining data while explicitly exposing task-level trade-offs. Our analysis shows that aggregate leaderboards are unstable: model rankings vary sharply across task categories, scale provides only modest and inconsistent gains, and architectural and pretraining alignment frequently outweigh parameter count. These results highlight limitations of current evaluation practices and position GENEB as a reference framework for principled comparison and category-aware model selection in genomic machine learning.
\end{abstract}

\section{Introduction}

\begin{figure}[h]
    \centering
    \includegraphics[width=0.97\linewidth]{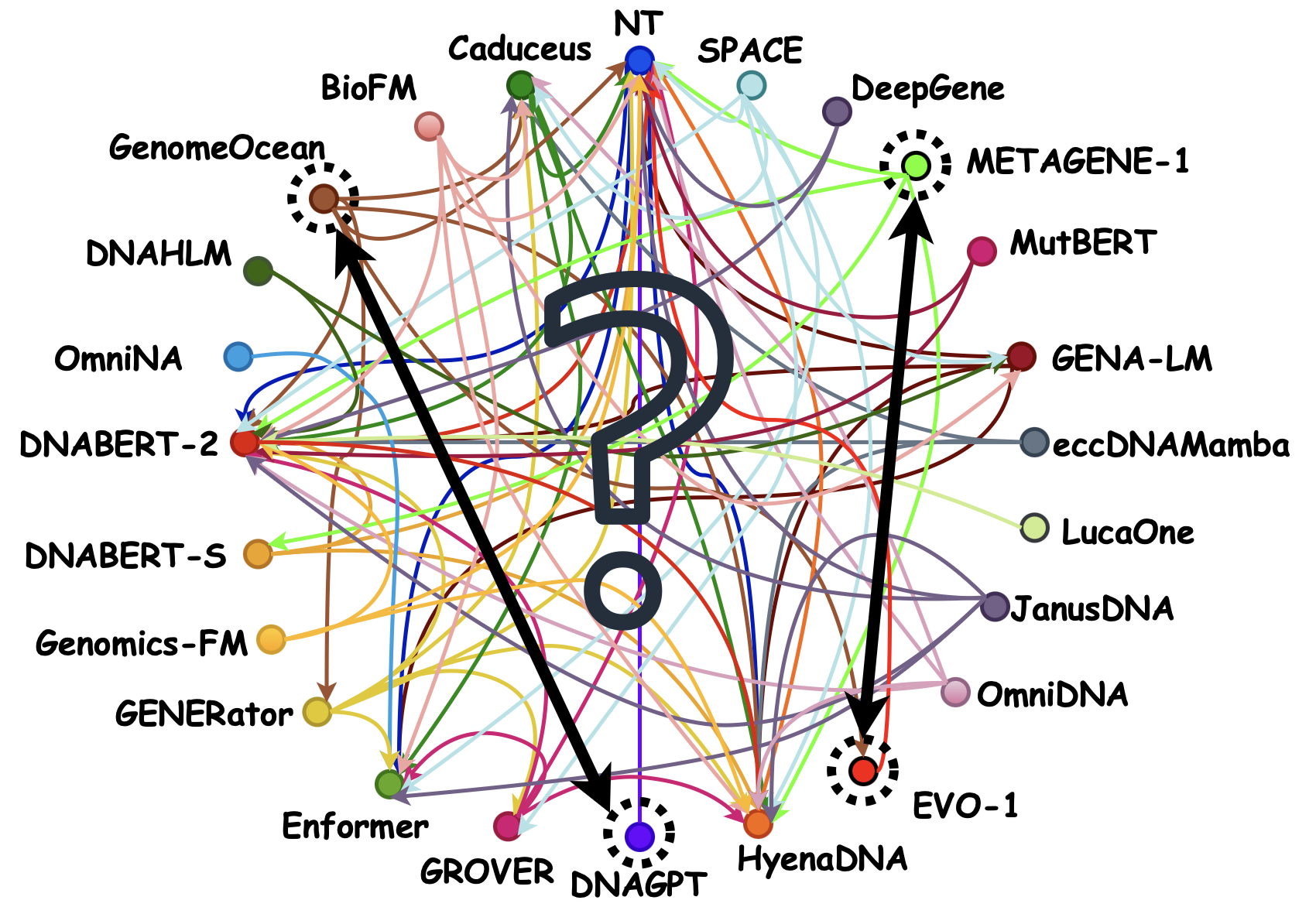}
\caption{\textbf{Fragmented comparison landscape of genomic foundation
models.} Each node represents a published model; directed edges denote
models explicitly used as baselines or comparators in the corresponding
paper. The sparse, disconnected graph reflects the absence of unified
cross-model evaluation in genomic machine learning.}
    \label{fig:chaotic_models}
\end{figure}

The genomic machine learning landscape has expanded rapidly over the
past decade, producing a large and heterogeneous ecosystem of models,
architectures, and training paradigms. This expansion has not been
accompanied by commensurate methodological infrastructure for
comparison. Figure~\ref{fig:chaotic_models} illustrates the present
state of the field: models are evaluated on disjoint benchmarks,
compared under incompatible protocols, and frequently reported as
state-of-the-art within narrowly defined settings, leaving unclear how
different models relate to one another or whether reported
improvements reflect genuine progress.

This fragmentation makes even basic questions difficult to answer.
Principled comparison between widely discussed models such as
\textsc{DNA-GPT}~\cite{zhang2023dnagptgeneralizedpretrainedtool}, \textsc{GenomeOcean}~\cite{Zhou2025.01.30.635558}, and 
\textsc{Evo}~\cite{nguyen2024sequence} is currently not feasible: each is evaluated on different task
sets, preprocessing pipelines, and evaluation protocols. The same
model is sometimes characterised as a major breakthrough in one
context and as underperforming in another, reflecting not contradictory
evidence but the absence of a common evaluation framework. 

This problem is compounded by the rapid growth in scale and 
visibility of genomic foundation models. As models become larger 
and more expressive, claims of superiority and generality have grown 
correspondingly bolder, yet the methodological basis for adjudicating 
such claims has not kept pace. The result is a widening gap between 
what is asserted about model capability and what can be reliably 
established through reproducible cross-model evaluation.

We introduce \textbf{GENEB}, a large-scale benchmark evaluating 40
genomic foundation models on 100 tasks spanning 13 functional
categories under a unified probing protocol. GENEB is designed to
enable controlled, systematic comparison and to expose performance
trade-offs obscured by fragmented evaluation practices; in spirit, it
plays a role analogous to MTEB~\citep{muennighoff2023mtebmassivetextembedding}
in natural language processing, providing a unified evaluation
framework rather than a single-task leaderboard. By making results
directly comparable across models and tasks, GENEB establishes a
shared reference point for assessing progress in genomic machine
learning.

\paragraph{Conflict of Interest Disclosure.}
The authors declare no financial conflicts of interest. None of the
40 evaluated models was developed by the authors or their funders.

\section{Related Work}


The rapid growth of genomic foundation models has produced a 
heterogeneous landscape spanning diverse architectures, tokenization 
schemes, and pretraining strategies.

\paragraph{Architectures.}
Early genomic models predominantly adopted Transformer encoders trained with masked language modeling \citep{zhou2024dnabert2efficientfoundationmodel,dalla2023nucleotide,GENA_LM,sanabria2024grover}. More recent work explored decoder-only and generative architectures for unified sequence modeling and long-context processing \citep{zhang2023dnagptgeneralizedpretrainedtool,nguyen2024sequence,wu2025generatorlongcontextgenerativegenomic,Zhou2025.01.30.635558,li2025omnidnaunifiedgenomicfoundation}. To reduce attention complexity, alternative designs based on long convolutions and state-space models have been proposed, alongside hybrid architectures combining multiple paradigms \citep{nguyen2023hyenadnalongrangegenomicsequence,schiff2024caduceusbidirectionalequivariantlongrange,liu2025eccdnamambapretrainedmodelultralong,duan2025janusdnapowerfulbidirectionalhybrid,vishniakov2025gene42longrangegenomicfoundation}.

\paragraph{Tokenization and Pretraining.}
Tokenization strategies range from single-nucleotide and $k$-mer representations to learned BPE vocabularies, each offering different trade-offs between resolution and efficiency \citep{zhou2024dnabert2efficientfoundationmodel,Zhou2025.01.30.635558}. Pretraining data similarly varies from human-only and species-specific corpora to broad multi-species and domain-focused datasets, with prior studies suggesting potential benefits of both diversity and specialization depending on the task \citep{dalla2023nucleotide,wu2025generatorlongcontextgenerativegenomic,avsec2021enformer,doi:10.1073/pnas.2421738122}.

\paragraph{Benchmarks.}
Several benchmarks evaluate genomic foundation models, including Nucleotide Transformer tasks~\citep{dalla2023nucleotide}, GUE/GUE+~\citep{zhou2024dnabert2efficientfoundationmodel}, Genomic Benchmarks~\citep{Gresova2022.06.08.495248}, BEND~\citep{marin2024bendbenchmarkingdnalanguage}, and DNALongBench~\citep{Cheng2025.01.06.631595}. While these resources cover important regulatory, epigenetic, and cross-species tasks, they differ in task design and evaluation protocols and typically assess only a limited subset of models, making cross-paper comparison difficult.

\paragraph{Comparative Benchmarking Studies.}
Recent studies have explored broader comparisons of genomic foundation models, but usually evaluate a small number of representative architectures. For example, \citet{wang2025genomic} focus on approximately ten model families and predominantly human-centric tasks. Platform-based efforts such as OmniGenBench~\citep{wang2025omnigenbenchbenchmarkomnipotentmultimodal} provide dynamic leaderboards, but currently include a limited and evolving set of baselines, leaving many recent DNA-specific models unevaluated.

\paragraph{Positioning of GENEB.}
GENEB addresses these gaps by providing a large-scale, controlled benchmark covering 40 genomic foundation models evaluated on 100 DNA classification tasks across 13 functional categories. By evaluating all models on the full task suite under a unified probing-based protocol, GENEB enables matched comparisons across architecture, tokenization, and pretraining data and yields a complete performance matrix that exposes task-dependent trade-offs. We plan to release GENEB as a public benchmark with evaluations hosted on Hugging Face, serving as a community reference analogous to MTEB in NLP~\citep{muennighoff2023mtebmassivetextembedding}.

\paragraph{Extended Related Work.}
A detailed discussion of prior benchmarks, comparative studies, and architectural trends is provided in Appendix~\ref{app:related_work}.

\section{Methodology}

GENEB evaluates genomic foundation models (see Appendix~\ref{app:model_summary}, Table~\ref{tab:model_summary}) using an embedding-based \emph{probing} protocol: frozen sequence representations are assessed with lightweight classifiers, isolating representation quality and enabling controlled comparison across architectures and training regimes. The benchmark covers diverse genomic prediction tasks spanning multiple functional categories; full task definitions are provided in Appendix~\ref{subsec:task_taxonomy}.

\textbf{Probing setup.}
For each task, frozen embeddings are used as features for logistic regression (\texttt{max\_iter=1000}) and evaluated in 1-shot, 10-shot, and full-data regimes (Figure ~\ref{fig:fewshot}). Results are averaged over five fixed random seeds $\{13, 17, 42, 123, 997\}$. The stability of model rankings under non-linear probing is verified
empirically in Appendix~\ref{subsec:probe_stability}, and sensitivity of
few-shot conclusions to the choice of regularization strength is
analyzed in Appendix~\ref{subsec:fewshot_sensitivity}.

\textbf{Metric and data.} We report Matthews Correlation Coefficient (MCC), which is robust to class imbalance and standard in genomic evaluation. Tasks exceeding $10^5$ sequences are subsampled. An empirical analysis using \textsc{GenomeOcean} embeddings shows
that MCC stabilizes beyond this size, motivating $10^5$ as a
practical upper bound.

\section{Aggregate Performance Analysis Across 100 Genomic Tasks}
\label{sec:results}

\begin{figure}[h]
    \centering
\includegraphics[width=1\linewidth]{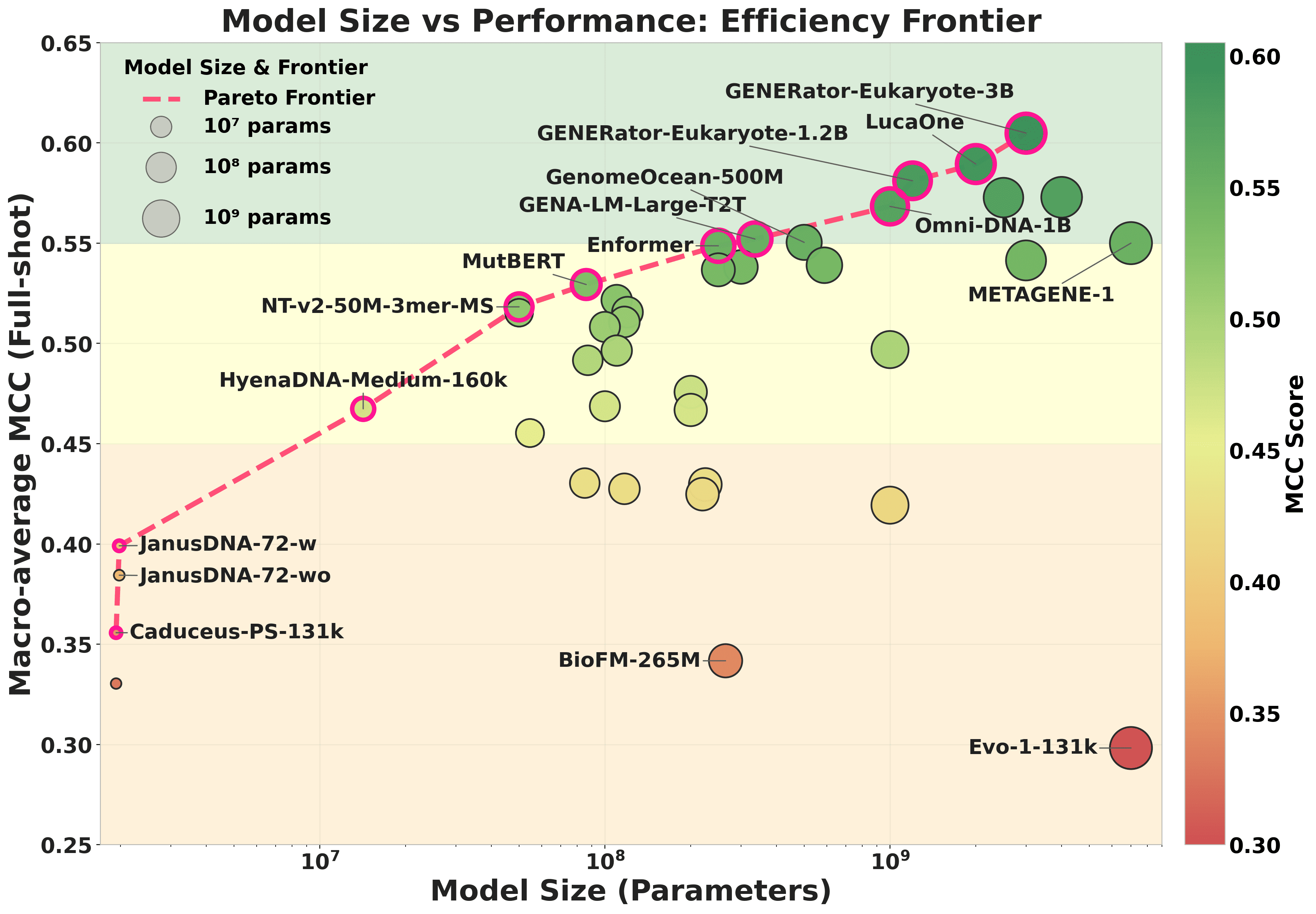}
    \caption{\textbf{Pareto frontier of model efficiency: macro-MCC vs.\
parameter count.} Each point represents one of the 40 genomic
foundation models, with parameter count on a logarithmic x-axis and
full-shot macro-average MCC on the y-axis. Marker size and color both
encode macro-MCC. The dashed line marks the Pareto frontier of best
performance--size trade-offs. Spearman correlation between
$\log(\text{params})$ and macro-MCC is $\rho = 0.573$ ($p < 0.001$);
excluding the prokaryotic-only outlier \textsc{Evo-1-131k} raises this
to $\rho = 0.694$ ($p < 0.001$). While scale is a substantial
predictor of aggregate performance, several large models fall below
the frontier, indicating that architecture and pretraining choices
can offset substantial scale differences (see
Section~\ref{sec:results}, Table~\ref{tab:per_cat_scaling}).}
\label{fig:frontier}
\end{figure}

We present a systematic analysis of 40 DNA foundation models evaluated
on 100 genomic prediction tasks spanning 13 functional categories. Our
goal is to characterize how model scale (Figure~\ref{fig:frontier}),
architecture, tokenization, and pretraining data interact under a
unified evaluation protocol, and to extract practically relevant
patterns for model selection. Unless stated otherwise, all statistics
refer to MCC aggregated within the GENEB benchmark.

\paragraph{The Scale--Performance Disconnect.}
Model size exhibits a statistically significant and substantial
association with aggregate performance ($\rho = 0.573$, $p < 0.001$;
Figure~\ref{fig:frontier}), which strengthens further to
$\rho = 0.694$ ($p < 0.001$) once the prokaryotic-only outlier
\textsc{Evo-1-131k} is excluded (see Domain Mismatch paragraph below).
Even with this strong aggregate trend, however, model selection is
not reducible to parameter count: among the 36 in-domain models
(excluding prokaryotic-only, microbial-only, and plant-specific
pretraining), we identify 25 instances in which a model at least
$5\times$ smaller outperforms a larger counterpart in aggregate MCC,
with this count being identical under micro- and macro-averaging.
A representative example is \textsc{MutBERT} (86M, Transformer-encoder),
which exceeds \textsc{eccDNAMamba} (537M, Mamba) by $+0.110$
macro-MCC despite a $6.2$-fold size difference, illustrating that
non-scale design choices can offset substantial scale gaps in GENEB.
Category-level scaling correlations are reported in
Table~\ref{tab:per_cat_scaling}. We additionally verify that these
aggregate statistics are robust to the choice of averaging scheme:
macro-averaged MCC (weighting all 13 categories equally) yields
rankings that correlate with the micro-averaged rankings reported
here at $\rho = 0.988$ (Appendix~\ref{subsec:macro_micro}).

\begin{table}[t]
\centering
\caption{\textbf{Per-category scaling correlations.} Spearman rank
correlation $\rho$ between $\log_{10}(\text{parameter count})$ and
macro-MCC within each functional category ($n=40$ models). $\rho$
near $+1$ indicates that larger models systematically outperform
smaller ones; $\rho$ near $0$ indicates no monotonic relationship
between size and performance. The $p$-value tests whether the
observed $\rho$ differs from zero; bolded values are significant at
$p<0.05$. Rows sorted by $\rho$ descending. Scaling is significant
in 11 of 13 categories, with $\rho$ ranging from $0.346$ (DNA
methylation) to $0.587$ (histone modifications).}
\label{tab:per_cat_scaling}
\small
\begin{tabular}{lcc}
\toprule
\textbf{Category} & \textbf{$\rho$} & \textbf{$p$} \\
\midrule
Histone modifications      & $\mathbf{0.587}$ & $<0.001$ \\
lncRNA                     & $\mathbf{0.575}$ & $<0.001$ \\
Splice sites               & $\mathbf{0.547}$ & $<0.001$ \\
Enhancers                  & $\mathbf{0.497}$ & $0.001$ \\
Promoters                  & $\mathbf{0.493}$ & $0.001$ \\
Coding/non-coding          & $\mathbf{0.486}$ & $0.002$ \\
Mouse enhancers            & $\mathbf{0.483}$ & $0.002$ \\
Virus/phage                & $\mathbf{0.439}$ & $0.005$ \\
Regulatory                 & $\mathbf{0.387}$ & $0.014$ \\
TF binding                 & $\mathbf{0.361}$ & $0.022$ \\
DNA methylation            & $\mathbf{0.346}$ & $0.029$ \\
Species classification     & $0.308$          & $0.054$ \\
Chromatin accessibility    & $0.240$          & $0.136$ \\
\bottomrule
\end{tabular}
\end{table}

\paragraph{Architecture Comparison Under Controlled Conditions.}
To isolate architectural effects, we compare models matched by
pretraining corpus (multi-species) and tokenization (BPE), and focus on
pairs where the remaining configuration differences are minimized.
Under these controlled conditions, Transformer models show substantial advantages over the state-space 
model available in this controlled setting. Specifically,
\textsc{GenomeOcean-500M} (Transformer-decoder) exceeds \textsc{eccDNAMamba}
(Mamba) by $+0.131$ macro-MCC (0.550 vs.\ 0.419), and
\textsc{Omni-DNA-1B} shows a comparable $+0.149$ gap (0.568 vs.\
0.419). Within Transformers, encoder models exceed decoders across all six
matched pairs (Appendix~\ref{subsec:controlled_pairs}), with the
strongest gap of $+0.127$ MCC for \textsc{GENA-LM-Large-T2T} over
\textsc{OmniNA-220M} (0.552 vs.\ 0.425) under matched multi-species/BPE
conditions. These gaps hold under both micro- and macro-averaging
(Appendix~\ref{subsec:macro_micro}); at the per-category level, however,
the encoder-decoder ranking is task- and setting-dependent.

\begin{figure*}[h]
    \centering
    \includegraphics[width=1\linewidth]{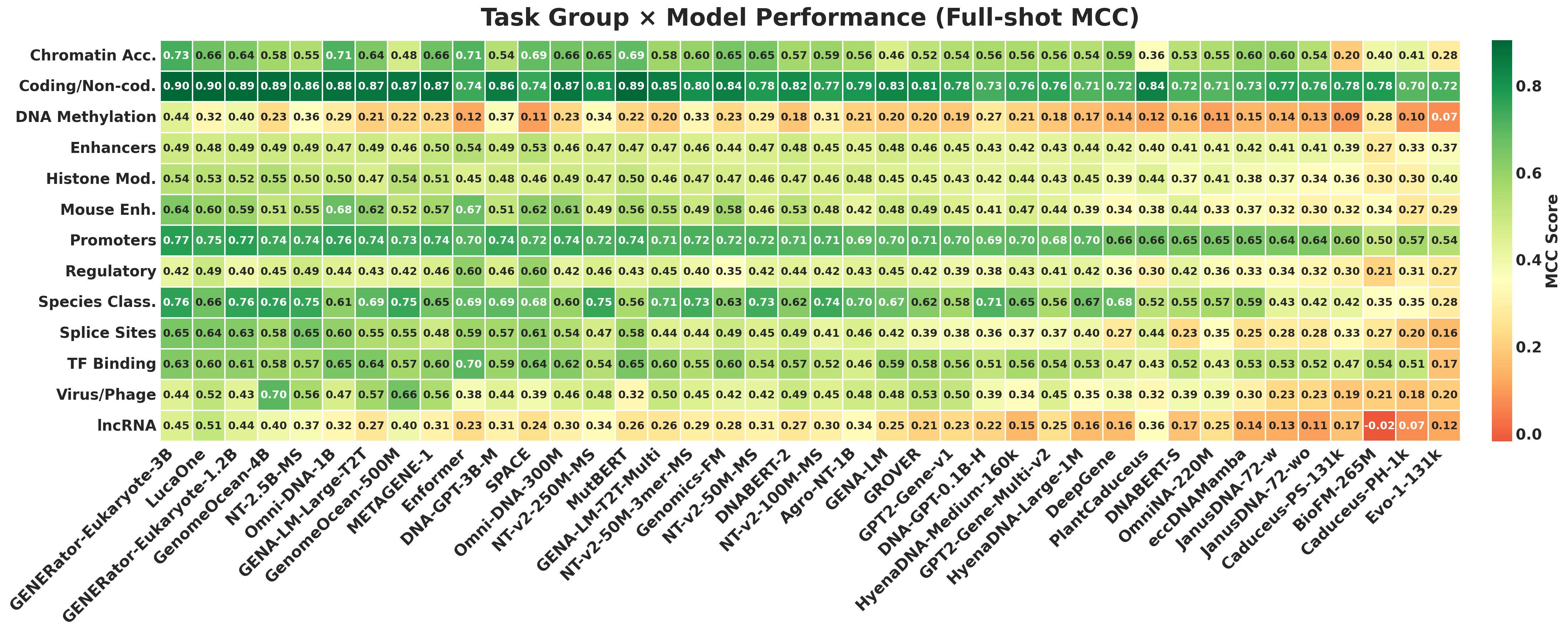}
\caption{\textbf{Model performance across task groups.} 
Heatmap shows full-shot MCC averaged within each task group for 40 genomic foundation models, sorted by overall full-shot macro-average MCC. 
Cell values report category-level mean MCC, with colors ranging from red/orange for lower scores to green for higher scores. 
The results reveal substantial task-level heterogeneity: some categories, such as promoter, coding/non-coding, and species-classification tasks, are consistently easier, whereas DNA methylation, lncRNA, virus/phage, and regulatory tasks remain challenging. 
This category-specific structure shows that aggregate model rankings can hide important differences in downstream behavior.}
\label{fig:category_summary}    
\end{figure*}

\begin{figure*}[t]
    \centering
\includegraphics[width=1\linewidth]{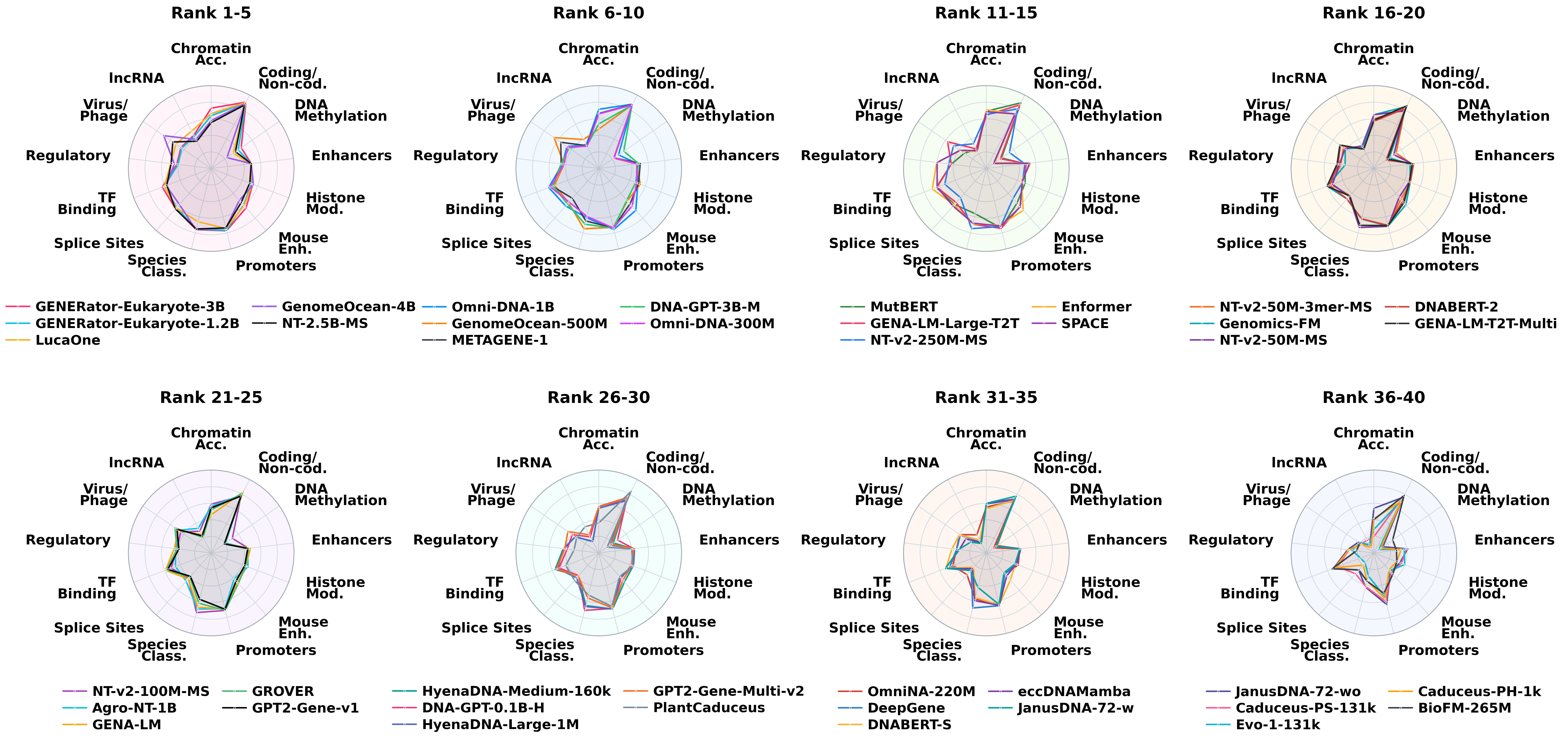}
\caption{\textbf{Radar plots for category-aware model selection.}
Each subplot shows full-shot macro-MCC across the 13 GENEB task
categories for a group of five models, grouped by overall macro-MCC
rank from strongest to weakest. The plots expose category-specific
strengths not captured by aggregate rankings: \textsc{Enformer} has
a moderate overall rank but leads on TF binding ($0.698$), enhancers
($0.539$), and regulatory tasks ($0.604$), and ranks second on mouse
enhancers ($0.674$) and third on chromatin accessibility ($0.711$);
the \textsc{GenomeOcean} family is particularly strong on virus/phage
tasks; and plant-oriented models such as \textsc{PlantCaduceus} and
\textsc{Agro-NT-1B} show relative strength on lncRNA tasks. These
profiles motivate task-specific model selection over global
leaderboard position.}
\label{fig:radar}
\end{figure*}

\paragraph{Architecture Gaps Are Largest on Cross-Species Regulatory Tasks.}
Architecture-dependent gaps are particularly pronounced on tasks
requiring cross-species generalization (Figure~\ref{fig:category_summary};
Figure~\ref{fig:radar}). On virus/phage,
\textsc{GenomeOcean-500M} (Transformer-decoder) exceeds
\textsc{eccDNAMamba} (Mamba) by $+0.355$ macro-MCC ($0.657$ vs.\
$0.302$). On mouse enhancers, \textsc{Omni-DNA-1B} exceeds
\textsc{eccDNAMamba} by $+0.305$ ($0.675$ vs.\ $0.370$), and
\textsc{GENA-LM-Large-T2T} exceeds \textsc{OmniNA-220M} by $+0.284$
under matched multi-species/BPE conditions. These category-level gaps are several-fold larger than the
aggregate parameter-tier gain ($+0.075$ macro-MCC between models
above $1$B and below $200$M parameters), reinforcing that architecture
and pretraining alignment can exceed scale-related differences on
several categories.

\paragraph{Chromatin Accessibility: A Domain Where SSM Models Become Competitive.}
Chromatin accessibility provides a notable exception to the general
pattern of Transformer dominance. \textsc{eccDNAMamba} (Mamba)
exceeds \textsc{GenomeOcean-500M} (Transformer-decoder) by $+0.124$
macro-MCC ($0.599$ vs.\ $0.475$) on this category. Both \textsc{JanusDNA-72-w} and \textsc{eccDNAMamba} also exhibit a
substantial within-model advantage on chromatin accessibility
relative to their overall performance ($+0.200$ and $+0.179$ MCC
above their respective aggregate macro-MCC), a clear instance of
category-level specialization (see Appendix~\ref{subsec:specialization}) within the SSM model family. The controlled pretraining-corpus
analysis additionally shows a consistent $+0.076$ macro-MCC advantage
of multi-species over human-only pretraining for chromatin
accessibility ($6$/$6$ pairs;
Table~\ref{tab:transfer_human_multi}). Within GENEB, these results
suggest that chromatin accessibility may benefit from inductive biases
captured by the evaluated SSM model family when combined with
taxonomically diverse pretraining.

\begin{table}[h]
\centering
\caption{\textbf{Transfer learning: human vs.\ multi-species pretraining.}
Per-category $\Delta$ MCC (multi-species $-$ human) averaged across
6 controlled model pairs matched on architecture, tokenization, and
size ($\pm 2\times$). For each pair, the within-category MCC is
computed by averaging over tasks; the table reports the cross-pair
average. ``Wins'' indicates the number of pairs in which multi-species
exceeds human in that category. ``Overall'' is the macro-averaged
aggregate ($\Delta$ MCC averaged across the 13 categories).}
\label{tab:transfer_human_multi}
\small
\begin{tabular}{lcc}
\toprule
\textbf{Task Category} & \textbf{$\Delta$ MCC} & \textbf{Wins} \\
\midrule
\textbf{Overall (macro)}   & $\mathbf{+0.018}$ & \textbf{5/6} \\
\midrule
\rowcolor{green!15} Chromatin Acc.    & $+0.076$ & 6/6 \\
\rowcolor{green!15} Splice Sites      & $+0.065$ & 5/6 \\
\rowcolor{green!15} Mouse Enh.        & $+0.041$ & 5/6 \\
\rowcolor{green!15} lncRNA            & $+0.034$ & 6/6 \\
\rowcolor{gray!10}  Coding/Non-cod.   & $+0.018$ & 4/6 \\
\rowcolor{gray!10}  Histone Mod.      & $+0.017$ & 5/6 \\
\rowcolor{gray!10}  Species Class.    & $+0.016$ & 3/6 \\
\rowcolor{gray!10}  Enhancers         & $+0.005$ & 4/6 \\
\rowcolor{gray!10}  Promoters         & $+0.004$ & 3/6 \\
\rowcolor{gray!10}  Regulatory        & $+0.003$ & 3/6 \\
\rowcolor{gray!10}  TF Binding        & $-0.003$ & 3/6 \\
\rowcolor{gray!10}  DNA Methylation   & $-0.014$ & 2/6 \\
\rowcolor{red!15}   Virus/Phage       & $-0.025$ & 3/6 \\
\bottomrule
\end{tabular}
\vspace{1mm}
\begin{flushleft}
\footnotesize
\textcolor{green!50!black}{Green}: multi-species advantage ($\Delta > +0.02$).
\textcolor{gray}{Gray}: parity ($|\Delta| \leq 0.02$).
\textcolor{red!70!black}{Red}: human advantage ($\Delta < -0.02$).
\end{flushleft}
\end{table}

\paragraph{Tokenization Strategy Effects.}
Holding architecture and pretraining corpus fixed, we isolate the
effect of tokenization scheme across 11 matched controlled pairs
(Appendix~\ref{subsec:tokenization_controlled}). No global ordering
emerges; the preferred scheme varies with model family and
pretraining setting. The single available Transformer-decoder pair
under matched multi-species pretraining shows BPE exceeding $k$-mer
by $+0.032$ macro-MCC (\textsc{GenomeOcean-4B} vs.\
\textsc{DNA-GPT-3B-M}). Within Transformer-encoders under matched
multi-species pretraining, BPE and $k$-mer perform comparably
($+0.006$ across 5 pairs, with all pair-level gaps within $\pm 0.02$
MCC). Under matched human pretraining, single-nucleotide
tokenization (\textsc{MutBERT}) exceeds BPE baselines in both
available comparisons: $+0.033$ over \textsc{GENA-LM} and $+0.038$
over \textsc{GROVER}. Non-standard vocabularies show mixed effects: BPE exceeds
\textsc{BioFM-265M}'s BioToken framework by $+0.134$ and $+0.125$
in two Transformer-decoder/human pairs, while \textsc{LucaOne}'s
mixed nucleotide--amino-acid vocabulary exceeds $k$-mer by $+0.017$
over a matched Transformer-encoder (\textsc{NT-2.5B-MS}). Tokenization thus interacts with
architecture, scale, and task structure rather than admitting a
single global ordering in GENEB.

\paragraph{Transfer Learning Analysis: Isolating Pretraining Corpus Effects.}
To disentangle pretraining data effects from architectural confounds,
we restrict analysis to controlled pairs matched on architecture,
tokenization, and size (within $\pm 2\times$;
Appendix~\ref{subsec:controlled_pairs}).

\textit{Human vs.\ multi-species pretraining} (6 controlled pairs
across Transformer-encoders and -decoders, all BPE-tokenized) shows
a small aggregate effect: multi-species pretraining yields an
average $+0.018$ macro-MCC improvement over human-only. However,
the effect is structured by task category
(Table~\ref{tab:transfer_human_multi}). Multi-species pretraining
shows its most consistent advantage on chromatin accessibility
(6/6 pairs; $\Delta=+0.076$), with additional positive shifts on
splice sites ($\Delta=+0.065$), mouse enhancers ($\Delta=+0.041$),
and lncRNA ($\Delta=+0.034$, 6/6). In contrast, virus/phage tasks favor human-only pretraining on average
($\Delta=-0.025$). Remaining categories show small or near-parity shifts
under the $|\Delta|<0.02$ criterion (smallest $\Delta=-0.014$ on DNA
methylation).

\begin{figure*}
    \centering
    \includegraphics[width=1.0\linewidth]{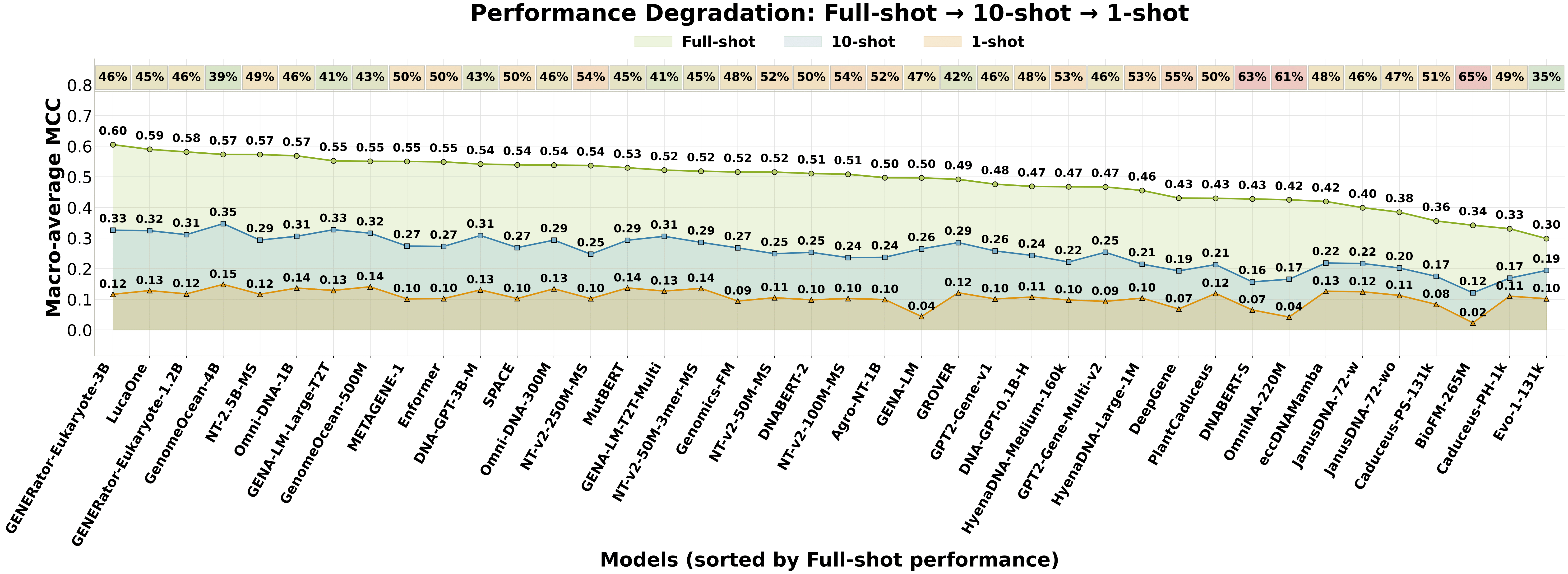}
\caption{\textbf{Few-shot performance degradation.} Macro-average MCC of genomic foundation models under full-data, 10-shot, and 1-shot evaluation regimes. Models are sorted by full-data performance. The top band reports the relative performance drop from full-data to 10-shot evaluation, highlighting the sensitivity of each model to limited supervision.}
    \label{fig:fewshot}
\end{figure*}

\textit{Multi-species vs.\ multi-species-microbial} (2 controlled
pairs; Transformer-encoder; $k$-mer) yields the largest corpus effect
observed in GENEB controlled comparisons. Models pretrained on
general multi-species data that include eukaryotic genomes
(\textsc{NT-v2-100M-MS}, \textsc{Genomics-FM}) exceed
microbial-focused pretraining (\textsc{DNABERT-S}) by $+0.084$
macro-MCC on average (2/2 pairs). The largest category gaps align
with known biological differences between microbial and eukaryotic
genomes: splice sites ($\Delta=+0.222$), species classification
($\Delta=+0.130$), lncRNA ($\Delta=+0.116$), and DNA methylation
($\Delta=+0.108$). In GENEB, this is consistent with
microbial-focused pretraining being insufficient for eukaryotic
genomic prediction tasks even under matched architectures and
tokenization.

\textit{Eukaryotic-genes vs.\ multi-species} (1 controlled pair;
3B Transformer-decoder; $k$-mer) shows that
\textsc{GENERator-Eukaryote-3B} (curated eukaryotic genes) exceeds
\textsc{DNA-GPT-3B-M} (broad multi-species) by $+0.063$ macro-MCC
overall, with the largest advantages on chromatin accessibility
($+0.191$), lncRNA ($+0.142$), and mouse enhancers ($+0.124$). The
only category favoring broad multi-species in this comparison is
regulatory element prediction ($-0.040$). This conclusion rests on a single pair and may reflect model-specific
training choices (Appendix~\ref{subsec:controlled_pairs}), but is
consistent with corpus curation yielding gains beyond broad sequence
diversity.

\textit{What cannot be concluded from controlled comparisons.}
Several pretraining corpus types lack matched architectural controls
in the current model set and therefore cannot be attributed cleanly
to data effects: plant-genome models (unique architectures and scale
points), human-mouse epigenomic profile models (\textsc{Enformer},
\textsc{SPACE}; CNN-Transformer hybrids), and the prokaryotic model
(\textsc{Evo-1-131k}; StripedHyena). Apparent advantages or deficits
for these groups should be interpreted as potentially confounded by
architecture and training procedure rather than pretraining data
alone.

\paragraph{Few-Shot Robustness Reveals an Inverse Performance Pattern.}
Across all 40 models, mean macro-MCC degrades from $0.488$
(full-shot) to $0.253$ (10-shot) to $0.106$ (1-shot)
(Figure~\ref{fig:fewshot}), corresponding to relative reductions of
$48.2\%$ and $78.2\%$, respectively. Per-model relative 10-shot
drops vary substantially, from $35\%$ for \textsc{Evo-1-131k} to
$65\%$ for \textsc{BioFM-265M} (median $48\%$), reflecting both
category-level and model-level effects that the aggregate trend
does not capture.

\textit{First}, the degradation is structured by task category
(per-category breakdown in Appendix~\ref{app:results}). Promoter
prediction retains $38.8\%$ of full-shot macro-MCC at 1-shot and
species classification retains $30.1\%$, consistent with broadly
distributed sequence-composition signals (GC content, codon usage,
$k$-mer enrichment) that pretraining captures and 1-shot
supervision preserves. In contrast, three categories collapse to
within $0.03$ MCC of random performance at 1-shot: virus/phage
($\text{MCC}_{\text{1-shot}}=0.027$, $93.5\%$ drop), DNA methylation
($0.015$, $93.2\%$), and lncRNA ($0.022$, $91.3\%$). These collapses
align with under-representation of viral genomes in eukaryotic
pretraining corpora, the position-specific nature of DNA methylation
signals, and the low-level sequence similarity of lncRNA to the
broader non-coding genome.

\textit{Second}, beyond this category-level structure, few-shot
robustness is inversely aligned with full-shot performance at the
model level. The five smallest absolute drops occur in
\textsc{Evo-1-131k} ($\Delta = 0.196$), \textsc{Caduceus-PH-1k}
($0.220$), \textsc{JanusDNA-72-wo} ($0.272$),
\textsc{Caduceus-PS-131k} ($0.272$), and \textsc{JanusDNA-72-w}
($0.275$) -- all among the weakest models in full-shot. Top
full-shot performers exhibit the opposite pattern:
\textsc{GENERator-Eukaryote-3B} ($\Delta=0.489$),
\textsc{GENERator-Eukaryote-1.2B} ($0.463$), \textsc{LucaOne}
($0.461$), and \textsc{NT-2.5B-MS} ($0.456$) all exceed $0.42$ in
absolute drop. This pattern does not indicate greater robustness in
weaker models: a small absolute drop reflects a low full-shot
ceiling that leaves limited room for further degradation, not
recovery of useful signal at 1-shot. \textsc{Evo-1-131k} is
illustrative -- its low ceiling stems from domain mismatch
(see Domain Mismatch paragraph below), not from few-shot regime itself.

These observations expose a structural limitation of aggregate
few-shot leaderboards: they conflate category tractability with
model quality, and reward models whose absolute drop is small
because their full-shot performance is already low. Category-resolved
evaluation (Appendix~\ref{app:results}) is therefore the
appropriate diagnostic in low-supervision regimes.

\paragraph{The Hard Frontier of GENEB: Where Scaling Does Not Help.}
A substantial fraction of GENEB tasks remains far from saturation:
$28$ of $100$ tasks have mean MCC below $0.35$
(Figure~\ref{fig:category_summary}), dominated by 4mC methylation
prediction (\emph{G.\ subterraneus} $0.061$;
\emph{E.\ coli} $0.103$; \emph{G.\ pickeringii} $0.107$) and plant
lncRNA identification (\emph{S.\ lycopersicum} $0.221$;
\emph{G.\ max} $0.228$; \emph{T.\ aestivum} $0.238$). Even the
strongest model on these categories,
\textsc{GENERator-Eukaryote-3B}, reaches only $0.206$--$0.477$ on 4mC
tasks, and \textsc{LucaOne} reaches $0.417$--$0.629$ on plant lncRNA.

Category-level scaling analysis (Spearman $\rho$ between
$\log(\text{params})$ and per-category macro-MCC across all 40
models; Table~\ref{tab:per_cat_scaling}) shows that this hardness
is not purely a parameter-count problem. Eleven of the 13
categories show positive scaling significant at $p<0.05$, but with
substantial variation in strength ($\rho$ ranging from $0.346$ for
DNA methylation to $0.587$ for histone modifications). Two
categories show no significant scaling: species classification
($\rho=0.308$, $p=0.054$) and chromatin accessibility ($\rho=0.240$,
$p=0.136$). Even where scaling is statistically significant, the
hard frontier of GENEB (4mC methylation, plant lncRNA) shows that
scaling alone does not close the absolute performance gap; progress
on these tasks will require complementary advances in
pretraining-corpus design, inductive biases, or task-specific
supervision.

\begin{figure}[t]
    \centering
\includegraphics[width=1\linewidth]{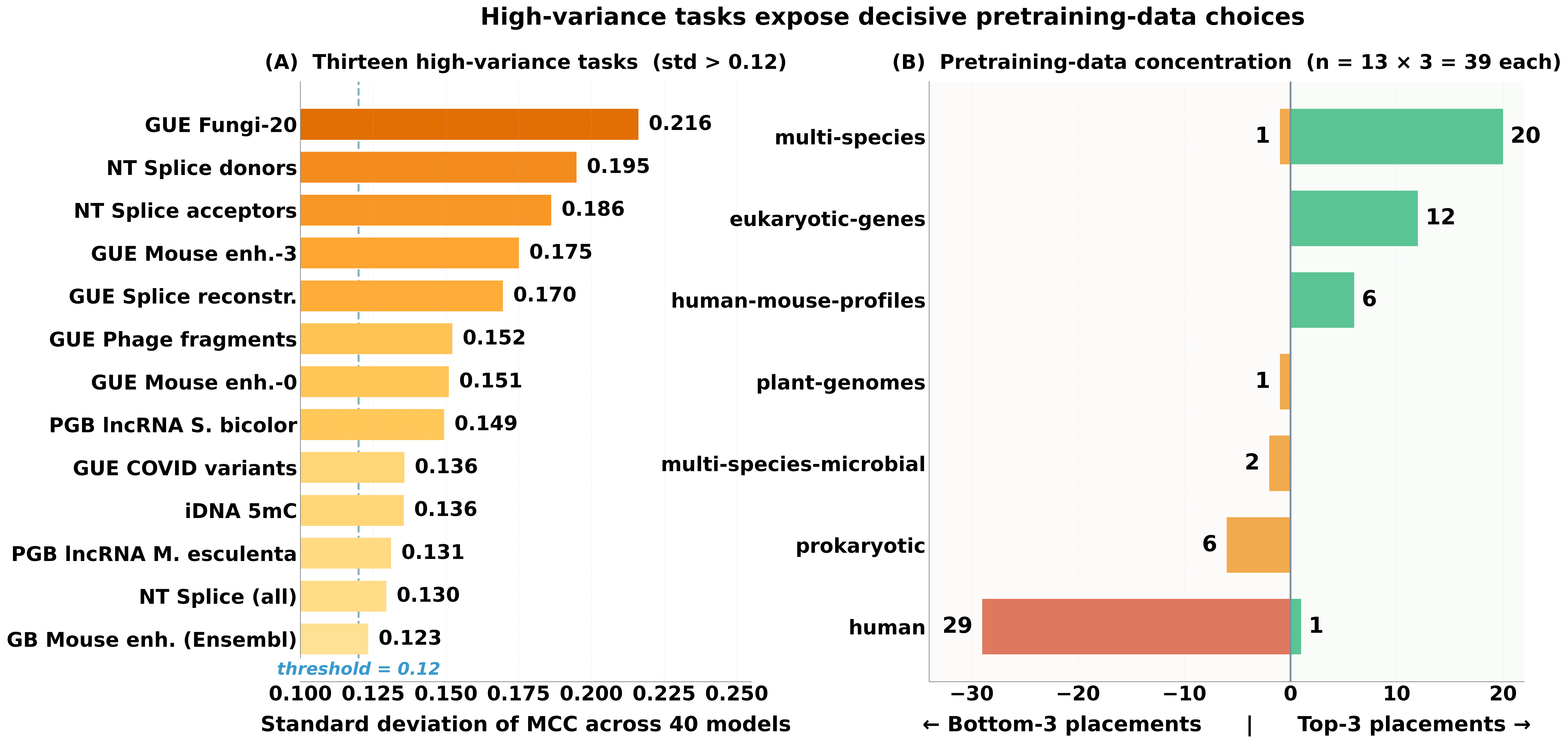}
   \caption{\textbf{High-variance tasks reveal the role of pretraining data.}
\textbf{(A)} GENEB tasks with cross-model standard deviation above $0.12$, corresponding to settings where model selection most strongly affects downstream performance. 
\textbf{(B)} Pretraining-data composition of top-3 and bottom-3 placements across these tasks. Multi-species and eukaryotic-gene pretraining dominate top placements, while human-only, prokaryotic, and microbial pretraining are concentrated among bottom placements. The result indicates that high-variance tasks expose biologically meaningful differences in pretraining scope that are obscured by aggregate leaderboards.}
    \label{fig:high-var}
\end{figure}

\paragraph{High-Variance Tasks Reveal Decisive Design Patterns.}
The architectural and pretraining effects identified above are most
pronounced on tasks where models disagree most: 13 GENEB tasks
exhibit cross-model standard deviation above $0.12$, indicating
settings in which model choice has outsized impact on downstream
performance (Figure~\ref{fig:high-var}A). Aggregating top-3 and
bottom-3 placements across these tasks ($39$ slots in each band)
reveals concentration effects (Figure~\ref{fig:high-var}B). By
pretraining-data type, multi-species and eukaryotic-gene pretraining
together capture $32/39$ top-3 placements ($20$ and $12$,
respectively), while human-only pretraining is concentrated almost
entirely in the bottom band ($29/39$ bottom-3, only $1/39$ top-3).
By architecture, the same imbalance holds: Transformer-decoders and
Transformer-encoders together account for $33/39$ top-3 placements
($18$ and $15$, respectively), whereas Mamba ($17/39$ bottom-3),
Hybrid-Mamba-MoE ($7/39$), and StripedHyena ($6/39$) dominate
the bottom. These high-variance tasks thus operationalise the
findings above into a practical diagnostic: in settings where
GENEB models disagree most, pretraining scope and architectural
family -- not scale -- predict whether a model will land in the top
or bottom tier.

\paragraph{Domain Mismatch and Out-of-Domain Models.}
The poor aggregate performance of \textsc{Evo-1-131k} -- one of the largest
models in GENEB ($7$B parameters), yet ranked among the weakest on TF binding, mouse enhancers,
chromatin accessibility, species classification, and splice sites
(Figure~\ref{fig:radar}) -- is best understood as a symptom of a broader issue: the GENEB task suite is heavily skewed toward eukaryotic genomic phenomena
($12$ of $13$ categories), creating a structural disadvantage for
models pretrained on prokaryotic or microbial corpora. This disadvantage is distinct from -- and should
not be conflated with -- the effects of scale, architecture, or
tokenization. Recomputing the scale--performance correlation without \textsc{Evo-1-131k}
raises Spearman $\rho$ from $0.573$ to $0.694$ ($p<0.001$) and widens
the tier gap between models above $1$B and below $200$M parameters
from $+0.075$ to $+0.102$ macro-MCC.
\textsc{DNABERT-S}, pretrained on multi-species microbial genomes,
exhibits a milder version of the same pattern: overall macro-MCC
$0.427$ versus $0.512$ for the two eukaryotic multi-species models
under matched architecture and tokenization
(\textsc{NT-v2-100M-MS}, \textsc{Genomics-FM}; see Transfer Learning
Analysis above). We therefore treat aggregate rankings as a poor
proxy for performance in prokaryotic or viral genomics, and
encourage users of GENEB working in those domains to consult
per-task results directly rather than relying on the overall
leaderboard.

\paragraph{Practitioner Recommendations.}
The findings above motivate per-category model selection rather than
reliance on aggregate rankings (Figure~\ref{fig:radar}). The
per-category recommendations below are organised around four
recurring patterns. For compact deployment under tight compute
budgets, \textsc{MutBERT} (86M, Transformer-encoder, single-nucleotide
tokenization, human pretraining) is the top sub-100M model in 8 of 13
categories and the strongest $\le 100$M model on overall macro-MCC
($0.529$). For epigenomic-profile tasks (TF binding, regulatory,
enhancers), the CNN--Transformer hybrids \textsc{Enformer} and
\textsc{SPACE} are consistently among the top models, with
\textsc{Enformer} matching the top model \textsc{Omni-DNA-1B}
on mouse enhancers. At the opposite end, hard
regimes (DNA methylation, plant lncRNA) remain unsolved within GENEB:
the best macro-MCC is $0.440$ for DNA methylation
(\textsc{GENERator-Eukaryote-3B}) and $0.508$ for plant lncRNA
(\textsc{LucaOne}), with scaling weak for DNA methylation ($\rho=0.346$,
Table~\ref{tab:per_cat_scaling}). Finally, few-shot evaluation
reranks the best model in 8 of 13 categories, so the model winning
under full supervision is not necessarily the model to deploy with
only $\sim 10$ labelled examples per task.

\textit{Coding/non-coding.} \textsc{GENERator-Eukaryote-3B} ($0.904$)
and \textsc{LucaOne} ($0.901$) lead under full supervision; the most
compact strong alternative is \textsc{MutBERT} ($0.894$, 86M), which
nearly matches the 3B-parameter leader. In 10-shot deployments,
\textsc{MutBERT} becomes the top performer ($0.748$ vs.\ $0.694$ for
\textsc{GENERator-Eukaryote-3B}).

\textit{Promoter prediction.} \textsc{GENERator-Eukaryote-3B}
($0.774$) leads, with \textsc{GENERator-Eukaryote-1.2B} ($0.768$) and
\textsc{Omni-DNA-1B} ($0.759$) close behind. Under tighter compute
budgets, \textsc{Omni-DNA-300M} ($0.740$) and \textsc{MutBERT}
($0.739$, 86M) provide strong compact options.

\textit{Species classification.} The top three models -- %
\textsc{GenomeOcean-4B} ($0.762$), \textsc{GENERator-Eukaryote-3B}
($0.761$), and \textsc{GENERator-Eukaryote-1.2B} ($0.757$) -- all
exceed $0.75$ and use multi-species or eukaryotic-gene pretraining.
For compute-constrained deployment, the \textsc{NT-v2} family
(\textsc{NT-v2-250M-MS} $0.747$, \textsc{NT-v2-100M-MS} $0.741$,
\textsc{NT-v2-50M-3mer-MS} $0.734$) offers near-equivalent performance
at one-tenth the parameter count.

\textit{Chromatin accessibility.} \textsc{GENERator-Eukaryote-3B}
($0.728$), \textsc{Omni-DNA-1B} ($0.714$), and \textsc{Enformer}
($0.711$) all exceed $0.71$; \textsc{Enformer} (252M) is the strongest
sub-300M option, and \textsc{MutBERT} (86M, $0.691$) the strongest
sub-100M option. The category also stands out as a setting in which
the evaluated SSM model is competitive (\textsc{eccDNAMamba} $0.599$).
Under 10-shot supervision the ranking reranks substantially:
\textsc{GENA-LM-Large-T2T} becomes the leader ($0.567$, down from
$0.645$ at full-shot), while \textsc{GENERator-Eukaryote-3B} drops
to $0.540$.

\textit{TF binding.} \textsc{Enformer} ($0.698$) leads by a wide
margin, followed by \textsc{Omni-DNA-1B} ($0.647$) and \textsc{MutBERT}
($0.646$, 86M). \textsc{MutBERT} is therefore the recommended compact
choice.

\textit{Virus/phage classification.} \textsc{GenomeOcean-4B}
($0.697$) and \textsc{GenomeOcean-500M} ($0.657$) dominate this
category, with a sharp drop to \textsc{GENA-LM-Large-T2T}
($0.569$); below 200M parameters performance drops further
(\textsc{GROVER} $0.532$). \textsc{GenomeOcean-4B} remains the
10-shot leader ($0.377$), though absolute scores collapse
substantially.

\textit{Mouse enhancers.} \textsc{Omni-DNA-1B} ($0.675$) and
\textsc{Enformer} ($0.674$) are effectively tied at full supervision;
\textsc{Enformer} (252M) is therefore the recommended compact choice. Under 10-shot supervision, however,
\textsc{SPACE} becomes the leader ($0.379$, from a full-shot score of
$0.618$), while \textsc{Omni-DNA-1B} drops to $0.253$ -- a substantial
rerank that practitioners working with limited cross-species labels
should account for.

\textit{Splice sites.} \textsc{NT-2.5B-MS} ($0.652$) and
\textsc{GENERator-Eukaryote-3B} ($0.648$) lead under full supervision.
The strongest sub-300M alternative is \textsc{Enformer} ($0.586$);
the strongest sub-100M alternative is \textsc{MutBERT} ($0.579$).
Microbial-focused pretraining transfers poorly to this category,
losing $0.222$ macro-MCC to multi-species in controlled comparisons
(see Transfer Learning Analysis above). Under 10-shot supervision, \textsc{LucaOne} becomes
the top model ($0.298$, from a full-shot score of $0.636$), while
\textsc{NT-2.5B-MS} drops to $0.242$.

\textit{Regulatory.} \textsc{Enformer} ($0.604$) and \textsc{SPACE}
($0.598$) lead substantially, reflecting their epigenomic-profile
pretraining. The strongest sub-100M option is \textsc{HyenaDNA-Medium-160k}
($0.432$). In 10-shot, \textsc{DNABERT-2} unexpectedly leads
($0.216$, from a full-shot score of $0.435$), while \textsc{Enformer}
drops to $0.184$; we view this rerank as protocol-fragile given the
small absolute scores in this low-data regime.

\textit{Histone modifications.} The \textsc{GenomeOcean} family leads
(\textsc{GenomeOcean-4B} $0.545$, \textsc{GenomeOcean-500M} $0.537$),
along with \textsc{GENERator-Eukaryote-3B} ($0.537$). With 30 tasks,
this is the largest category in GENEB and shows the strongest
within-category scaling ($\rho = 0.587$; Table~\ref{tab:per_cat_scaling}).
The strongest sub-100M option is \textsc{MutBERT} ($0.501$). In
10-shot, \textsc{MutBERT} also leads ($0.300$), making it the
recommended choice across regimes when compute is constrained.

\textit{Enhancers.} \textsc{Enformer} ($0.539$) and \textsc{SPACE}
($0.526$) lead, followed by \textsc{METAGENE-1} ($0.505$). The
strongest sub-300M alternative is \textsc{Enformer} itself.
Under 10-shot, \textsc{GENA-LM-Large-T2T} becomes the top model
($0.372$, from $0.488$ at full-shot), a switch worth noting for
low-data deployments.

\textit{lncRNA.} \textsc{LucaOne} ($0.508$) leads by a substantial
margin, followed by \textsc{GENERator-Eukaryote-3B} ($0.453$) and
\textsc{GENERator-Eukaryote-1.2B} ($0.438$). The strongest sub-300M
option is \textsc{PlantCaduceus} ($0.357$), reflecting its
plant-specific pretraining. This is a hard regime: no per-task score
exceeds $0.63$ across any of the 6 plant lncRNA tasks, and 10-shot
performance is at or below $0.21$ for every model.

\textit{DNA methylation.} \textsc{GENERator-Eukaryote-3B} ($0.440$)
leads, with \textsc{GENERator-Eukaryote-1.2B} ($0.397$) and
\textsc{DNA-GPT-3B-M} ($0.367$) following. This is the hardest
category in GENEB: $6$ of $8$ tasks have mean MCC below $0.25$;
although scaling is statistically significant
($\rho = 0.346$, $p = 0.029$; Table~\ref{tab:per_cat_scaling}),
no sub-300M model exceeds $0.34$. Few-shot performance collapses
to near-random levels across all models, with no 1-shot score
exceeding $0.04$.

\section{Conclusion}

We introduced GENEB, a benchmark evaluating 40 genomic foundation
models across 100 tasks from 13 functional categories under a
unified linear-probing protocol with macro-MCC as the principal
aggregation metric. Our results show that while model scale shows a substantial
aggregate association with performance ($\rho = 0.573$), it remains
an imperfect predictor of category-level outcomes: architecture and
pretraining alignment frequently offset substantial scale
differences, and the model that wins under full supervision reranks
under 10-shot evaluation in 8 of 13 categories. Transformer-based models generally outperform
the evaluated state-space alternative, though domain-specific
exceptions exist (e.g., chromatin accessibility); tokenization
effects interact with architecture rather than admitting a single
global ordering; and microbial-only corpora transfer poorly to
eukaryotic tasks.

Overall, these findings argue for category-aware, controlled
evaluation rather than aggregate leaderboards. GENEB provides a
reference framework to support principled model comparison and
selection in genomic machine learning.

\section{Limitations}

GENEB has several limitations.

\textit{Long-range tasks.} GENEB underrepresents tasks requiring
explicit modeling of very long-range regulatory interactions
($>10$\,kb). As a result, models with explicit long-context capability
(\textsc{HyenaDNA-Large-1M}, \textsc{Caduceus-PS-131k},
\textsc{Evo-1-131k}) are not exercised on the regime where their
architectural priors would most likely yield differentiating gains.
A detailed enumeration of considered-but-excluded long-range datasets
and model context-length constraints is provided in
Appendix~\ref{subsec:excluded_long_range_tasks}.

\textit{Task selection and curation.} Task selection is constrained
by available datasets and existing benchmarks. Some constituent tasks
may be noisy or weakly defined, particularly in the hard regimes
identified in Section~\ref{sec:results} (DNA methylation, plant
lncRNA), where label quality and supervision signal vary across
sources. Further task curation and refinement is needed as genomic
benchmarks mature.

\textit{Model coverage.} Not all genomic foundation models could be
included due to unavailable weights, incompatible pipelines, or
computational constraints. Excluded models and inclusion criteria are
discussed in Appendix~\ref{app:excluded_models}.

\textit{Prokaryotic and viral task gap.} Of the 13 GENEB categories,
only virus/phage classification reflects a non-eukaryotic domain;
prokaryotic gene prediction, microbial genome assembly verification,
and CRISPR system characterization are not currently represented. As
a result, aggregate GENEB rankings are an unreliable proxy for
performance in prokaryotic or viral genomics (see Domain Mismatch
paragraph, Section~\ref{sec:results}).

\textit{Frozen representations and pooling-tokenization interactions.}
GENEB evaluates frozen representations using linear probing, which
enables controlled comparison of embedding quality across the
40-model set but may underestimate the performance achievable with
task-specific fine-tuning. Empirical analysis in
Appendix~\ref{subsec:probe_stability} shows that model rankings under
linear probing are highly consistent with those under non-linear MLP
probes; whether this stability extends to full task-specific
fine-tuning remains an open question. Additionally, single-nucleotide
and $k$-mer tokenizations produce substantially longer token
sequences than BPE for a fixed input window, so the choice of pooling
(mean, attention-weighted, or final-token) may favor different
schemes. We use mean pooling throughout; pooling-tokenization
interactions are not fully disentangled.

\textit{Aggregate metric.} GENEB reports both micro- and
macro-averaged MCC; histone modifications (30 tasks) and promoters
(22 tasks) jointly account for over half of the per-task evaluations,
so micro-averaging is structurally biased toward these two categories
(see Appendix~\ref{subsec:macro_micro}). We treat macro-MCC as the
principal aggregation, and recommend that category-level results,
rather than any single overall ranking, drive model selection.

\section{Use of Large Language Models}
\label{app:llm_usage}

Large language models were used as writing and editing assistants
during the preparation of this manuscript. Specifically, they were
employed to improve clarity, organization, and phrasing of the
text, as well as to assist with LaTeX formatting. All experimental
design, data processing, analysis, and interpretation of results
were performed by the authors, and all reported findings were
verified against the underlying benchmark outputs.

\section*{Acknowledgements}

This work was supported by the Ministry of Economic Development of the Russian Federation (agreement No. 139-15-2025-013, dated June 20, 2025, subsidy identifier 000000C313925P4B0002).

\section*{Impact Statement}

GENEB is intended to improve the rigor of model comparison in
genomic representation learning by replacing heterogeneous,
single-paper evaluations with a unified protocol across 40 models
and 100 tasks. We expect several positive effects on the field.
First, category-aware evaluation reduces the risk that
practitioners select models based on aggregate leaderboards that
mask substantial heterogeneity across biological task types,
particularly in clinically and agriculturally relevant domains
(e.g., regulatory element prediction, plant lncRNA, viral
classification). Second, our controlled comparisons isolate the
contributions of architecture, tokenization, and pretraining
corpus, which we hope will inform principled design of future
genomic foundation models rather than indiscriminate scaling.

We are mindful of several potential concerns. Genomic models can
in principle be applied to dual-use research, including the design
of pathogenic sequences; we believe a benchmark of representation
quality on standard prediction tasks does not meaningfully shift
this risk surface, but we note that responsible release practices
for individual models remain the responsibility of their authors.
Additionally, GENEB inherits biases from its constituent datasets:
the task suite is skewed toward eukaryotic and, within eukaryotes,
toward human and well-studied model organisms, which may
under-represent biologically and clinically important
non-model-organism settings. Users of GENEB should consult
per-task and per-category results when applying findings to
domains beyond those directly represented in the benchmark.

\bibliography{example_paper}
\bibliographystyle{icml2026}

\newpage
\appendix
\onecolumn

\section{Related Work}
\label{app:related_work}

\begin{figure}[h]
    \centering
    \includegraphics[width=0.6\linewidth]{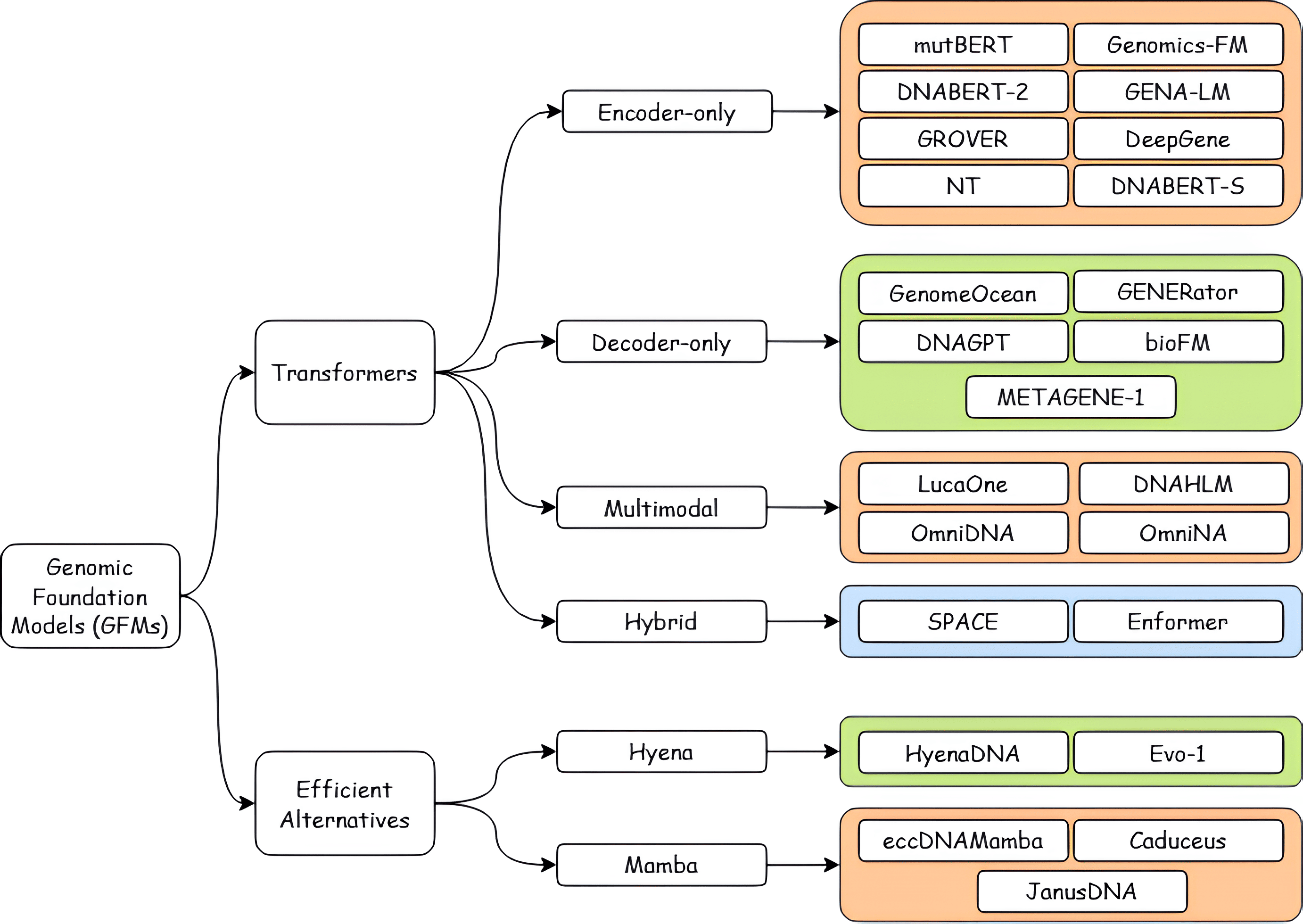}
    \caption{Architectural taxonomy of genomic foundation models.}
    \label{fig:models}
\end{figure}

The rapid development of foundation models for genomics has led to diverse architectural and methodological innovations (see  Figure~\ref{fig:models}). We organize prior work by architectural design, tokenization, 
pretraining strategy, and benchmark scope.

\paragraph{Transformer-Based Encoder Models.}
Early DNA language models predominantly adopted BERT-style encoder architectures with masked language modeling (MLM) objectives. DNABERT-2~\citep{zhou2024dnabert2efficientfoundationmodel} addressed computational inefficiencies of $k$-mer tokenization by replacing it with Byte-Pair Encoding (BPE), achieving comparable performance with 21$\times$ fewer parameters. The Nucleotide Transformer~\citep{dalla2023nucleotide} scaled this paradigm to 2.5B parameters, demonstrating that multi-species pretraining improves cross-species generalization. GENA-LM~\citep{GENA_LM} extended context lengths to 36kb using sparse attention mechanisms (BigBird), while GROVER~\citep{sanabria2024grover} introduced frequency-balanced BPE vocabularies to mitigate the ``rare word problem'' in genomic sequences.

\paragraph{Autoregressive and Generative Models.}
Decoder-only architectures have gained prominence for their generative capabilities and unified training objectives. DNA-GPT~\citep{zhang2023dnagptgeneralizedpretrainedtool} introduced a comprehensive token language incorporating sequence, numerical, and instruction tokens for multi-task genomic analysis. Evo~\citep{nguyen2024sequence} achieved genome-scale sequence modeling using the StripedHyena architecture, processing up to 131kb contexts with single-nucleotide resolution, with Evo~2~\citep{Brixi2025.02.18.638918} extending this to all domains of life. GENERator~\citep{wu2025generatorlongcontextgenerativegenomic} demonstrated that gene-focused pretraining on eukaryotic sequences outperforms whole-genome approaches for both understanding and generation tasks. GenomeOcean~\citep{Zhou2025.01.30.635558} pioneered training on metagenomic assemblies, capturing rare biosphere diversity across 645Gbp of environmental sequences. Omni-DNA~\citep{li2025omnidnaunifiedgenomicfoundation} unified cross-modal tasks (DNA-to-text, DNA-to-image) within a single autoregressive framework through vocabulary expansion.

\paragraph{State Space Models and Efficient Architectures.}
To address the quadratic complexity of attention mechanisms, several works have explored sub-quadratic alternatives. HyenaDNA~\citep{nguyen2023hyenadnalongrangegenomicsequence} employed implicit long convolutions via the Hyena operator, enabling processing of sequences up to 1M nucleotides with $\mathcal{O}(L\log L)$ complexity. Caduceus~\citep{schiff2024caduceusbidirectionalequivariantlongrange} built upon the Mamba selective state space model, introducing bidirectional processing and reverse-complement (RC) equivariance to respect DNA's double-stranded nature. eccDNAMamba~\citep{liu2025eccdnamambapretrainedmodelultralong} adapted the Mamba-2 architecture specifically for extrachromosomal circular DNA, incorporating circular topology awareness through specialized augmentation strategies.

\paragraph{Hybrid Architectures.}
Recent work has combined the strengths of different architectural paradigms. JanusDNA~\citep{duan2025janusdnapowerfulbidirectionalhybrid} introduced a bidirectional Mamba-Attention-MoE hybrid that achieves both autoregressive efficiency and bidirectional understanding, processing up to 1M base pairs at single-nucleotide resolution. Gene42~\citep{vishniakov2025gene42longrangegenomicfoundation} extended dense attention to 192kb through incremental pretraining with adjusted RoPE frequencies.

\paragraph{Multi-Modal and Multi-Species Models.}
Several recent models aim for broader biological understanding through multi-modal integration. LucaOne~\citep{He2024.05.10.592927} unified nucleic acid and protein modeling within a single 2B-parameter encoder using a shared 39-token vocabulary, demonstrating emergent understanding of the central dogma. OmniNA~\citep{Shen2024.01.14.575543} integrated nucleotide sequences with textual annotations, enabling natural language-based genomic analysis across 17 diverse tasks.

\paragraph{Long-Range Genomic Modeling.}
Modeling distal regulatory interactions requires extended context lengths. Enformer~\citep{avsec2021enformer} combined convolutional towers with 11 transformer layers to capture interactions up to 100kb, significantly improving gene expression prediction from sequence. DNALongBench~\citep{Cheng2025.01.06.631595} provides a comprehensive benchmark for evaluating long-range dependencies up to 1M base pairs. SPACE~\citep{yang2025spacegenomicprofilepredictor} proposed supervised pretraining on genomic profile prediction as an alternative to self-supervised approaches, employing Mixture-of-Experts (MoE) for cross-species modeling and profile-grouped decoders for capturing inter-track dependencies.

\paragraph{Tokenization Strategies.}
Tokenization choices significantly impact model performance and efficiency. Single-nucleotide tokenization~\citep{nguyen2023hyenadnalongrangegenomicsequence, duan2025janusdnapowerfulbidirectionalhybrid} preserves fine-grained resolution critical for SNP analysis but increases sequence length. $K$-mer approaches~\citep{dalla2023nucleotide} capture local context but suffer from vocabulary explosion and sensitivity to mutations. BPE-based methods~\citep{zhou2024dnabert2efficientfoundationmodel, Zhou2025.01.30.635558} balance efficiency and biological relevance by learning data-driven vocabularies. BioToken~\citep{Medvedev2025.03.27.645711} introduced biologically-informed tokenization that explicitly encodes structural annotations (exon/intron boundaries) and variant information within the token vocabulary, achieving strong performance with only 265M parameters.

\paragraph{Specialized Domains.}
Beyond general-purpose models, specialized architectures address specific genomic applications. DNABERT-S~\citep{zhou2024dnabertspioneeringspeciesdifferentiation} targets species differentiation through contrastive learning with Manifold Instance Mixup, doubling clustering performance over baselines. DeepGene~\citep{Zhang2024.04.24.590879} incorporates pan-genome graph representations via Minigraph to capture population-level genetic diversity. MutBERT~\citep{Long2025.01.23.634452} represents the genome as probabilistic distributions over allele frequencies, explicitly modeling population-level SNP variation. Genomics-FM~\citep{Ye2024.07.16.603653} employs an ensemble vocabulary strategy, pretraining on multiple tokenizations simultaneously and selectively activating task-appropriate vocabularies during fine-tuning. PlantCaduceus~\citep{doi:10.1073/pnas.2421738122} adapts the Caduceus architecture for plant genomics, demonstrating remarkable cross-species transferability across 160My of evolutionary divergence. AgroNT~\citep{mendoza2023foundational} provides a specialized foundation model for edible plant genomes.

\paragraph{Benchmarks for Genomic Foundation Models.}
Standardized evaluation is essential for assessing the generalization capabilities of genomic foundation models, yet existing benchmarks typically focus on limited task families or biological domains. The Nucleotide Transformer benchmark~\citep{dalla2023nucleotide} introduced a suite of DNA classification tasks spanning histone mark occupancy, regulatory element identification, and splice site prediction, establishing a common evaluation baseline for early Transformer-based models. The Genome Understanding Evaluation (GUE/GUE+)~\citep{zhou2024dnabert2efficientfoundationmodel} extended this paradigm to multi-species settings, including human, mouse, virus, and yeast, and to variable sequence lengths ranging from tens to thousands of base pairs. Genomic Benchmarks~\citep{Gresova2022.06.08.495248} further emphasized regulatory element classification, such as enhancers, promoters, coding/non-coding regions, and open chromatin, across diverse organisms.

Beyond these core resources, additional benchmarks target specific biological regimes. Plant-focused benchmarks evaluate long non-coding RNA classification and transcriptional activity in crop genomes~\citep{yang2025omnigenbenchmodularplatformreproducible}, while DNA methylation datasets assess epigenetic modifications (4mC, 5mC, 6mA) across a range of taxa~\citep{Feng2024.08.16.608288}. Collectively, these benchmarks have enabled substantial progress, but they remain fragmented in task scope, preprocessing assumptions, and evaluation protocols.

\paragraph{Limitations of Existing Benchmarking Practices.}
A common limitation across prior benchmarks is the restricted set of evaluated models. Most studies assess a small number of representative architectures, often selected for convenience or compatibility with a specific benchmark, rather than attempting exhaustive cross-model evaluation. As a result, comparisons across papers are difficult to interpret, and it is often unclear which models constitute appropriate baselines for a given task. Moreover, reported state-of-the-art results frequently depend on benchmark-specific design choices, obscuring task-level trade-offs and failure modes.

Recent benchmarking efforts such as BEND~\citep{marin2024bendbenchmarkingdnalanguage} and DNALongBench~\citep{Cheng2025.01.06.631595} have highlighted important evaluation issues, including biologically meaningful task design and long-range dependency modeling. However, these benchmarks still evaluate limited model subsets and do not provide a unified performance matrix across a broad landscape of genomic foundation models.

\paragraph{Comparative Benchmarking Studies.}
Several recent works have attempted broader comparative analysis of genomic foundation models. For example, \citet{wang2025genomic} evaluate approximately ten representative model families, primarily centered on human functional annotation tasks. While such studies provide useful snapshots, the evaluated model sets remain limited and exclude many recently proposed architectures. In addition, results are typically reported in static tables without unified leaderboards, making it difficult to place new models in a broader performance context. As a consequence, researchers often lack clear guidance on which baselines are appropriate for a given task and how to position new models relative to existing work.

Platform-centric initiatives such as OmniGenBench~\citep{wang2025omnigenbenchbenchmarkomnipotentmultimodal} have made important strides toward reproducibility by providing modular infrastructure and dynamic leaderboards. However, their reported baselines are often limited to a small set of established models (e.g., DNABERT-2, HyenaDNA, RNA-FM) and place substantial emphasis on RNA modalities. Consequently, a large and rapidly evolving portion of the DNA-specific foundation model landscape remains underexplored.

\paragraph{Positioning of GENEB.}
GENEB is designed to complement and extend prior benchmarking efforts by enforcing exhaustive and controlled evaluation across a substantially broader model and task space. GENEB aggregates 100 DNA classification tasks drawn from multiple established benchmarks, including Nucleotide Transformer tasks, Genomic Benchmarks, GUE/GUE+, plant genomics benchmarks, and DNA methylation datasets, spanning 13 functional categories. In contrast to prior work, GENEB evaluates all included models on the full task suite under a unified probing-based protocol, regardless of the model’s original domain or anticipated strengths.

This exhaustive cross-evaluation yields a complete, static performance matrix over 40 genomic foundation models, including recent architectures such as BioFM, GENERator, JanusDNA, and DeepGene that are absent from many existing benchmarks. By eliminating selective reporting and enforcing matched comparisons across architecture, tokenization, and pretraining data, GENEB exposes task-dependent trade-offs and failure modes that are often obscured in partial or benchmark-specific evaluations.

Beyond the static analysis presented in this work, we intend to release GENEB as a public benchmark with model evaluations hosted on Hugging Face, enabling transparent comparison and reproducible evaluation of future genomic foundation models. In this sense, GENEB is conceptually aligned with MTEB \cite{muennighoff2023mtebmassivetextembedding} in natural language processing: a community-facing benchmark that provides a shared reference point for evaluating representation quality across diverse tasks. We hope that GENEB can serve a similar role for genomic foundation models, supporting more principled model comparison and task-aware model selection.

\section{Task Taxonomy and Benchmark Composition}
\label{subsec:task_taxonomy}

GENEB aggregates tasks from nine widely used genomic benchmarks
to ensure broad biological coverage and comparability to prior
work: \textbf{NT} (Nucleotide Transformer original, $18$ tasks)
and \textbf{NT-rev} (its revised release, $18$ tasks) covering
histone marks, enhancers, promoters, and splice sites;
\textbf{GUE} ($31$ tasks) covering empirically-derived histone
marks, TF binding, mouse enhancers, promoters, splice sites,
phage fragments, and species classification; \textbf{GB}
(Genomic Benchmarks, $9$ tasks) covering enhancers, promoters,
coding/non-coding, regulatory elements, and open chromatin;
\textbf{PGB} (Plant Genomic Benchmark, $7$ tasks) covering plant
lncRNA and transcriptional activity; \textbf{iPro-WAEL}
($8$ tasks) covering bacterial, plant, and human
cell-type-specific promoter recognition; \textbf{deep4mc}
($6$ tasks) and \textbf{iDNA-ABF} ($2$ tasks) covering DNA
methylation (4mC, 5mC, 6mA) across taxa; and \textbf{iDHS-EL}
($1$ task) for chromatin accessibility.

Our benchmark comprises $100$ classification tasks organized
into $13$ functional categories spanning regulatory element
prediction, epigenomic modification detection, and evolutionary
sequence analysis. This taxonomic organization reflects the
diverse challenges facing DNA foundation models and enables
systematic analysis across distinct biological domains.
Table~\ref{tab:task_groups} summarizes the benchmark composition,
with task counts ranging from single-task categories
(Coding/Non-coding, Chromatin Accessibility) to large multi-task
groups (Histone Modifications with $30$ tasks, Promoters with
$22$ tasks).

\begin{table*}[h]
\centering
\caption{Benchmark task organization across 13 functional categories. 
Tasks span regulatory element prediction, epigenetic modification 
detection, and evolutionary sequence analysis across bacterial, 
plant, and mammalian systems. Within each row, the source prefix 
(e.g., NT, GUE, GB) is given once and applies to all listed tasks.}
\label{tab:task_groups}
\small
\begin{tabular}{@{}llcp{9.5cm}@{}}
\toprule
\textbf{Category} & \textbf{Subcategory} & \textbf{$n$} & \textbf{Representative Tasks} \\
\midrule
\multirow{3}{*}{Histone Modifications} & Original (NT) & 10 & H3, H3K14ac, H3K36me3, H3K4me1/2/3, H3K79me3, H3K9ac, H4, H4ac \\
& Empirical (GUE EMP) & 10 & H3, H3K14ac, H3K36me3, H3K4me1/2/3, H3K79me3, H3K9ac, H4, H4ac \\
& Revised (NT-rev) & 10 & H2AFZ, H3K27ac, H3K27me3, H3K36me3, H3K4me1/2/3, H3K9ac, H3K9me3, H4K20me1 \\
\midrule
\multirow{4}{*}{Promoters} & Human cell-type (iPro) & 4 & iPro: GM12878, HUVEC, HeLa-S3, NHEK \\
 & Plant & 3 & iPro Arabidopsis (TATA / no TATA), PGB Promoter \emph{M.\ esculenta} \\
 & Bacterial (iPro) & 2 & iPro: \emph{B.\ amyloliquefaciens}, \emph{R.\ capsulatus} \\
 & General & 13 & GUE Prom 300 (all / TATA / no TATA), GUE Prom core (all / TATA / no TATA), NT Promoter (all / TATA / no TATA), NT-rev Promoter (all / TATA / no TATA), GB Promoter (no TATA) \\
\midrule
\multirow{2}{*}{Enhancers} & Species-specific (GB) & 4 & GB: Drosophila enh.\ (STARK), Mouse enh.\ (Ensembl), Human enh.\ (Cohn), Human enh.\ (Ensembl) \\
 & Type classification & 4 & NT Enhancers, NT Enhancers (types), NT-rev Enhancers, NT-rev Enhancers (types) \\
\midrule
\multirow{2}{*}{DNA Methylation} & 5mC / 6mA (iDNA) & 2 & iDNA: 5mC, 6mA \\
 & 4mC (6 species) & 6 & 4mC: \emph{A.\ thaliana}, \emph{C.\ elegans}, \emph{D.\ melanogaster}, \emph{E.\ coli}, \emph{G.\ pickeringii}, \emph{G.\ subterraneus} \\
\midrule
\multirow{2}{*}{Splice Sites} & Site-specific & 4 & NT Splice (donors / acceptors), NT-rev Splice (donors / acceptors) \\
 & Combined & 3 & NT Splice (all), NT-rev Splice (all), GUE Splice reconstr. \\
\midrule
lncRNA & Plant (6 species; PGB) & 6 & PGB lncRNA: \emph{G.\ max}, \emph{M.\ esculenta}, \emph{S.\ bicolor}, \emph{S.\ lycopersicum}, \emph{T.\ aestivum}, \emph{Z.\ mays} \\
\midrule
Mouse Enhancers & Tissue-specific (GUE) & 5 & GUE Mouse enh.: -0, -1, -2, -3, -4 \\
\midrule
TF Binding & Human TFs (GUE) & 5 & GUE TF: -0, -1, -2, -3, -4 \\
\midrule
\multirow{2}{*}{Species Classification} & Binary & 1 & GB Human-or-worm \\
 & Multi-class (GUE) & 2 & GUE: Fungi-20, Virus-40 \\
\midrule
Regulatory & Human regulatory (GB) & 2 & GB: Ensembl regulatory, OCR Ensembl \\
\midrule
Virus/Phage & Viral sequences (GUE) & 2 & GUE: Phage fragments, COVID variants \\
\midrule
Coding/Non-coding & Sequence type & 1 & GB Coding/Non-coding \\
\midrule
Chromatin Accessibility & DNase-seq & 1 & iDHS DNase-I \\
\bottomrule
\end{tabular}
\vspace{1mm}
\begin{flushleft}
\footnotesize\textit{Source prefixes:} NT = Nucleotide Transformer; NT-rev = Nucleotide Transformer revised; GUE = Genome Understanding Evaluation; GUE EMP = GUE empirically-validated subset; GB = Genomic Benchmarks; iPro = iPro-WAEL promoter dataset; PGB = Plant Genomics Benchmark; iDNA = iDNA-ABF; iDHS = iDHS-EL; 4mC = deep4mc.
\end{flushleft}
\end{table*}

\paragraph{Epigenomic and Chromatin Tasks.} The largest task
category, Histone Modifications ($n=30$), encompasses prediction
of post-translational modifications across distinct histone
marks including H3K4me1/2/3, H3K27ac, H3K36me3, H3K9ac, and H4
acetylation states. The tasks draw from three sources: \textbf{NT}
($10$ tasks: H3, H4, H3K4me1/2/3, H3K9ac, H3K14ac, H3K36me3,
H3K79me3, H4ac), \textbf{NT-rev} ($10$ tasks: revised versions
of H3K4me1/2/3, H3K9ac, and H3K36me3 plus newly added H2AFZ,
H3K27ac, H3K27me3, H3K9me3, and H4K20me1), and \textbf{GUE}
($10$ tasks: empirically derived subsets of the
NT histone marks). Chromatin Accessibility ($n=1$) is the
\textbf{iDHS-EL DNase-I} task, providing a complementary measure
of chromatin state.

\paragraph{Regulatory Element Tasks.} Promoter recognition
($n=22$) draws from six sources: \textbf{NT} and \textbf{NT-rev}
($6$ tasks total: TATA-containing, TATA-less, and combined human
promoter classification), \textbf{GUE} ($6$ tasks: core and 300-bp promoter variants with TATA-containing,
TATA-less, and combined splits), \textbf{GB} ($1$ task:
human non-TATA promoters), \textbf{iPro-WAEL} ($8$ tasks spanning
bacterial \textit{B.\ amyloliquefaciens} and
\textit{R.\ capsulatus}, plant \textit{Arabidopsis} TATA/non-TATA,
and human cell-type-specific lines GM12878, HUVEC, HeLa-S3,
NHEK), and \textbf{PGB} ($1$ task: \textit{M.\ esculenta}
transcriptional activity). Enhancer prediction ($n=8$) draws
from \textbf{NT} and \textbf{NT-rev} ($4$ tasks: combined and
type-stratified enhancer classification) and \textbf{GB}
($4$ tasks: \textit{Drosophila} Stark, mouse enhancers (Ensembl),
and human enhancers (Cohn and Ensembl)). General regulatory
element tasks ($n=2$) are both from \textbf{GB}: human Ensembl
regulatory region prediction and human open chromatin region
prediction.

\paragraph{Transcription Factor and Splicing Tasks.} TF Binding
prediction ($n=5$) comprises five \textbf{GUE} human transcription
factor binding tasks. Splice Site detection ($n=7$) combines
\textbf{NT} ($3$ tasks: donors, acceptors, combined),
\textbf{NT-rev} ($3$ revised counterparts), and \textbf{GUE}
($1$ reconstructed splice site task, an artificially difficult
reconstruction of splice junctions).

\paragraph{Sequence Modification Tasks.} DNA Methylation
prediction ($n=8$) draws from \textbf{deep4mc} ($6$ tasks: 4mC
detection in \textit{A.\ thaliana}, \textit{C.\ elegans},
\textit{D.\ melanogaster}, \textit{E.\ coli},
\textit{G.\ pickeringii}, \textit{G.\ subterraneus}) and
\textbf{iDNA-ABF} ($2$ tasks: 5mC and 6mA modification site
prediction).

\paragraph{Non-coding RNA and Species Classification Tasks.}
Long non-coding RNA classification ($n=6$) comprises the six
\textbf{PGB} plant lncRNA tasks: soybean (\textit{G.\ max}),
cassava (\textit{M.\ esculenta}), sorghum
(\textit{S.\ bicolor}), tomato (\textit{S.\ lycopersicum}),
wheat (\textit{T.\ aestivum}), and maize (\textit{Z.\ mays}).
Species Classification ($n=3$) draws from \textbf{GB}
(human-versus-worm discrimination) and \textbf{GUE} ($20$-way
fungal species and $40$-way viral species classification).

\paragraph{Additional Tasks.} Mouse Enhancer prediction ($n=5$)
comprises five \textbf{GUE} mouse enhancer tasks (tissue-specific
enhancer classification). Virus/Phage detection ($n=2$) comprises
two \textbf{GUE} tasks: phage fragment identification and COVID-19 variant classification. Coding/Non-coding discrimination ($n=1$)
is the \textbf{GB} coding-vs-intergenomic-sequences task.
This diverse task composition enables evaluation of transfer
learning across taxonomic boundaries (human to plant, eukaryotic
to prokaryotic), assessment of specialization versus
generalization trade-offs, and identification of task-specific
model advantages. The predominance of eukaryotic tasks ($12$ of
$13$ categories) reflects current genomic research priorities
while creating systematic disadvantages for prokaryotic-focused
models, a limitation we address in our analysis.

\clearpage

\section{Model Summary}
\label{app:model_summary}

Table~\ref{tab:model_summary} summarizes the $40$ genomic foundation
models evaluated in GENEB. For each model we report its canonical
name, architecture family, tokenization scheme, parameter count, and
pretraining-corpus type.

\begin{table*}[h]
\centering
\caption{\textbf{Summary of the 40 genomic foundation models evaluated in GENEB.}
Sorted by parameter count (descending). Tokenization labels: SN = single-nucleotide;
BPE = byte-pair encoding; k-mer = fixed-length $k$-mer; mixed = mixed
nucleotide/amino-acid vocabulary; BioToken = model-specific tokenizer.
Architecture labels: T-enc = Transformer-encoder; T-dec = Transformer-decoder.}
\label{tab:model_summary}
\footnotesize
\setlength{\tabcolsep}{4pt}
\renewcommand{\arraystretch}{1.05}
\begin{tabular}{@{}lllrl@{}}
\toprule
\textbf{Model} & \textbf{Architecture} & \textbf{Tokenization} & \textbf{Params} & \textbf{Pretraining data} \\
\midrule
METAGENE-1                & T-dec               & BPE        & $7$B   & multi-species \\
Evo-1-131k                & StripedHyena        & SN         & $7$B   & prokaryotic \\
GenomeOcean-4B            & T-dec               & BPE        & $4$B   & multi-species \\
GENERator-Eukaryote-3B    & T-dec               & k-mer      & $3$B   & eukaryotic-genes \\
DNA-GPT-3B-M              & T-dec               & k-mer      & $3$B   & multi-species \\
NT-2.5B-MS                & T-enc               & k-mer      & $2.5$B & multi-species \\
LucaOne                   & T-enc               & mixed      & $2$B   & multi-species \\
GENERator-Eukaryote-1.2B  & T-dec               & k-mer      & $1.2$B & eukaryotic-genes \\
Omni-DNA-1B               & T-dec               & BPE        & $1$B   & multi-species \\
Agro-NT-1B                & T-enc               & k-mer      & $1$B   & plant-genomes \\
SPACE                     & CNN-Transformer-MoE & SN         & $589$M & human-mouse-profiles \\
eccDNAMamba               & Mamba           & BPE        & $537$M   & multi-species \\
GenomeOcean-500M          & T-dec               & BPE        & $500$M & multi-species \\
GENA-LM-Large-T2T         & T-enc               & BPE        & $336$M & multi-species \\
Omni-DNA-300M             & T-dec               & BPE        & $300$M & multi-species \\
BioFM-265M                & T-dec               & BioToken   & $265$M & human \\
Enformer                  & CNN-Transformer     & SN         & $252$M & human-mouse-profiles \\
NT-v2-250M-MS             & T-enc               & k-mer      & $250$M & multi-species \\
PlantCaduceus             & Mamba           & SN         & $225$M & plant-genomes \\
OmniNA-220M               & T-dec               & BPE        & $220$M & multi-species \\
GPT2-Gene-Multi-v2        & T-dec               & BPE      & $200$M & human \\
GPT2-Gene-v1              & T-dec               & BPE      & $200$M & human \\
Genomics-FM               & T-enc               & BPE+k-mer  & $120$M & multi-species \\
DNABERT-S                 & T-enc               & k-mer      & $117$M & multi-species-microbial \\
DNABERT-2                 & T-enc               & BPE        & $117$M & multi-species \\
GENA-LM-T2T-Multi         & T-enc               & BPE        & $110$M & multi-species \\
GENA-LM                   & T-enc               & BPE        & $110$M & human \\
DNA-GPT-0.1B-H            & T-dec               & k-mer      & $100$M & multi-species \\
NT-v2-100M-MS             & T-enc               & k-mer      & $100$M & multi-species \\
GROVER                    & T-enc               & BPE        & $87$M  & human \\
MutBERT                   & T-enc               & SN         & $86$M  & human \\
DeepGene                  & T-enc+Graph-Transformer   & BPE        & $85$M  & human \\
HyenaDNA-Large-1M         & Hyena               & SN         & $55$M  & human \\
NT-v2-50M-3mer-MS         & T-enc               & k-mer      & $50$M  & multi-species \\
NT-v2-50M-MS              & T-enc               & k-mer      & $50$M  & multi-species \\
HyenaDNA-Medium-160k      & Hyena               & SN         & $14$M  & human \\
Caduceus-PS-131k          & Mamba           & SN         & $8$M   & human \\
JanusDNA-72-w             & Hybrid-Mamba-MoE    & SN         & $2$M   & human \\
JanusDNA-72-wo            & Hybrid-Mamba-MoE    & SN         & $2$M   & human \\
Caduceus-PH-1k            & Mamba           & SN         & $2$M   & human \\
\bottomrule
\end{tabular}
\end{table*}

\clearpage

\section{Models Excluded from Evaluation}
\label{app:excluded_models}

During our comprehensive survey of genomic foundation models, we identified several promising models that could not be included in our benchmark evaluation due to various technical and practical limitations. We document these exclusions for transparency and to inform future benchmark efforts. Table~\ref{tab:excluded_models} summarizes the excluded models and their exclusion reasons.

\begin{table*}[h]
\centering
\caption{\textbf{Genomic foundation models excluded from GENEB evaluation.} Models are grouped by exclusion category. All models were considered for inclusion but could not be evaluated due to the stated limitations.}
\label{tab:excluded_models}
\small
\begin{tabular}{llp{7cm}}
\toprule
\textbf{Model} & \textbf{Category} & \textbf{Exclusion Reason} \\
\midrule
\multicolumn{3}{l}{\textit{Unavailable or Private Weights}} \\
Gene42~\cite{gene42} & Private weights & Model weights not publicly released \\
NTv3~\cite{ntv3} & Private weights & Model weights not publicly released despite HuggingFace placeholder \\
\midrule
\multicolumn{3}{l}{\textit{Code/Infrastructure Issues}} \\
NucleotideGPT~\cite{nucleotidegpt} & Broken code & Corrupted weights and buggy code; TPU-only with no GPU support documentation \\
EpiGePT~\cite{epigept} & Missing extraction code & No embedding extraction interface; requires extensive custom implementation \\
ENBED~\cite{enbed} & Missing extraction code & No embedding extraction code; unclear model checkpoint location \\
HAD~\cite{had} & No code & No public code repository available \\
C.La.P.~\cite{clap} & No code & No public code repository available \\
HybriDNA~\cite{hybridna} & No code & No public code repository available \\
MxDNA~\cite{mxdna} & Broken weights & Incompatible or corrupted model weights \\
VQDNA~\cite{vqdna}  & No code & No public code repository available \\
BMFM-DNA~\cite{bmfmdna} & Dependency conflicts & Library version conflicts prevent execution \\
\midrule
\multicolumn{3}{l}{\textit{Computational Constraints}} \\
Evo2~\cite{evo2} & Hardware requirements & Requires H100/H200 GPUs exceeding our computational budget \\
\midrule
\multicolumn{3}{l}{\textit{Architectural Limitations}} \\
ChatNT~\cite{chatnt} & Wrapper model & Uses Nucleotide Transformer as encoder; not an independent foundation model \\
\bottomrule
\end{tabular}
\end{table*}

\subsection{Detailed Exclusion Notes}

\paragraph{Private or Unavailable Weights.}
Two models -- Gene42 and NTv3 -- were excluded because their pretrained weights have not been publicly released. Gene42~\cite{gene42} introduces a long-range genomic foundation model with dense attention capable of processing sequences up to 192 kbp, representing a significant architectural advancement. NTv3~\cite{ntv3} extends the Nucleotide Transformer family with joint sequence-function modeling. Both models report strong benchmark results in their respective publications, but the lack of public weights prevents independent evaluation.

\paragraph{Missing or Broken Code.}
Several models suffer from incomplete or non-functional code releases. NucleotideGPT~\cite{nucleotidegpt} provides weights that appear corrupted and code that only supports TPU execution without GPU alternatives. EpiGePT~\cite{epigept} and ENBED~\cite{enbed} lack embedding extraction interfaces, making it infeasible to obtain sequence representations for downstream evaluation without substantial reverse-engineering effort. HAD~\cite{had}, C.La.P.~\cite{clap}, and HybriDNA~\cite{hybridna} have no public code repositories despite published papers describing their architectures.

\paragraph{Runtime and Dependency Issues.}
MxDNA~\cite{mxdna} provides code and weights but fail to execute correctly due to weight incompatibilities or undocumented runtime requirements. BMFM-DNA~\cite{bmfmdna} from IBM Research encounters library version conflicts that prevent successful model loading.

\paragraph{Computational Requirements.}
Evo2~\cite{evo2}, the successor to the original Evo model included in our benchmark, requires H100 or H200 GPUs for inference due to its 40B parameter scale. This exceeds our available computational resources (A100 GPUs) and represents a practical barrier for many research groups.

\paragraph{Wrapper Models.}
ChatNT~\cite{chatnt} was excluded because it uses a frozen Nucleotide Transformer v2 as its DNA encoder, making it a wrapper rather than an independent foundation model. Evaluating ChatNT would redundantly measure NTv2 performance while introducing confounding factors from the conversational interface.

\subsection{Implications for Reproducibility}

The prevalence of excluded models (13 out of 53 initially surveyed, approximately 25\%) highlights a reproducibility challenge in genomic foundation model research. We encourage future model releases to include: (i) publicly available pretrained weights, (ii) documented embedding extraction code, (iii) clear hardware requirements, and (iv) tested installation procedures across common environments.

\subsection{Excluded Long-Range Regulatory Tasks}
\label{subsec:excluded_long_range_tasks}

GENEB does not include tasks that require explicit modeling of very
long-range regulatory interactions ($>10$\,kb). We considered the
following candidate datasets but excluded them on the grounds outlined
below:

\paragraph{Enhancer--promoter interaction prediction.} Tasks based on
predicting physical or functional contacts between enhancers and
their target promoters at distances of $50$--$500$\,kb (e.g.,
ChIA-PET, BENGI, HiChIP-derived datasets). Most genomic foundation
models in GENEB accept context windows below $6$\,kb, making fair
evaluation impossible without arbitrary cropping.

\paragraph{Three-dimensional chromatin contact map prediction.} Tasks
based on predicting Hi-C contact frequencies or topologically
associating domain (TAD) boundaries from megabase-scale sequence
windows. These require context lengths that exceed the input limits
of all but a few of the evaluated models.

\paragraph{Distal eQTL effect prediction.} Tasks linking sequence
variants to gene expression changes when the variant lies $>100$\,kb
from the affected gene. Direct sequence-to-effect formulations
require long context; reduced formulations introduce confounds that
defeat the purpose of fair cross-model comparison.

\paragraph{Whole-locus expression quantification.} Tasks predicting
tissue-specific expression from $>10$\,kb genomic windows around a
gene of interest, including locus-level enhancer collections.

\paragraph{Models affected by these exclusions.} Models with
explicit long-context capability -- specifically
\textsc{HyenaDNA-Large-1M} ($1$M tokens),
\textsc{Caduceus-PS-131k} ($131$k tokens),
\textsc{Evo-1-131k} ($131$k tokens), and
\textsc{JanusDNA-72-w}/\textsc{JanusDNA-72-wo} (Hybrid-Mamba-MoE
architecture) -- are not exercised on the regime where their architectural
priors would most plausibly yield differentiating gains. We treat
this as a known limitation of the current benchmark snapshot;
extending GENEB with long-range regulatory tasks under a unified
protocol is an important direction for future work.

\newpage


\newpage

\section{Probe Stability and Protocol Sensitivity Analysis}
\label{app:probe_stability}

GENEB evaluates frozen representations with linear probing, which provides
a controlled and interpretable measure of representation quality but in
principle could obscure information accessible only via non-linear readouts.
To verify that the rankings and conclusions reported in the main paper are
not artifacts of this choice, we conduct two complementary stability
analyses: (i) replacing the linear probe with a non-linear MLP probe
(Section~\ref{subsec:probe_stability}), and (ii) varying the regularization
strength of the linear probe across few-shot regimes
(Section~\ref{subsec:fewshot_sensitivity}). Both analyses are performed on
a representative subset of $11$ models and $13$ tasks, where each task is
drawn from a distinct functional category to ensure coverage of the
benchmark's diversity. The selected models span the full range of
architectures, tokenization schemes, pretraining corpora, and parameter
scales evaluated in GENEB; the model and task subsets are summarized in
Tables~\ref{tab:stability_models} and~\ref{tab:stability_tasks}.

\begin{table}[h]
\centering
\caption{Representative model subset used for probe stability and protocol
sensitivity analyses. The subset spans all major architectural families,
tokenization schemes, pretraining corpora, and three orders of magnitude
in parameter scale.}
\label{tab:stability_models}
\small
\begin{tabular}{@{}lcllc@{}}
\toprule
\textbf{Model} & \textbf{Params} & \textbf{Architecture} & \textbf{Tokenization} & \textbf{Pretraining} \\
\midrule
\textsc{GENERator-Eukaryote-3B}    & 3B    & Transformer-decoder & $k$-mer           & Eukaryotic genes \\
\textsc{LucaOne}                    & 2B    & Transformer-encoder & Other (custom)    & Multi-species \\
\textsc{GENERator-Eukaryote-1.2B}  & 1.2B  & Transformer-decoder & $k$-mer           & Eukaryotic genes \\
\textsc{GenomeOcean-4B}            & 4B    & Transformer-decoder & BPE               & Multi-species \\
\textsc{Omni-DNA-1B}                & 1B    & Transformer-decoder & BPE               & Multi-species \\
\textsc{GenomeOcean-500M}          & 500M  & Transformer-decoder & BPE               & Multi-species \\
\textsc{GENA-LM-Large-T2T}         & 336M  & Transformer-encoder & BPE               & Multi-species \\
\textsc{GROVER}                     & 87M   & Transformer-encoder & BPE               & Human \\
\textsc{MutBERT}                    & 86M   & Transformer-encoder & Single-nucleotide & Human \\
\textsc{HyenaDNA-Large-1M}         & 55M & Hyena               & Single-nucleotide & Human \\
\textsc{NT-v2-50M-MS}              & 50M   & Transformer-encoder & $k$-mer           & Multi-species \\
\bottomrule
\end{tabular}
\end{table}

\begin{table}[h]
\centering
\caption{Representative task subset used for probe stability and protocol
sensitivity analyses. One task is selected from each of the 13 functional
categories in GENEB.}
\label{tab:stability_tasks}
\small
\begin{tabular}{@{}ll@{}}
\toprule
\textbf{Category} & \textbf{Task} \\
\midrule
Histone Modifications     & GUE EMP H3K4me1 \\
Promoters                 & iPro GM12878 \\
Enhancers                 & GB Human enh.\ (Cohn) \\
DNA Methylation           & 4mC \emph{E.\ coli} \\
Splice Sites              & NT Splice (all) \\
lncRNA                    & PGB lncRNA \emph{Z.\ mays} \\
Mouse Enhancers           & GUE Mouse enh.-0 \\
TF Binding                & GUE TF-0 \\
Species Classification    & GUE Virus-40 \\
Regulatory                & GB Ensembl regulatory \\
Virus/Phage               & GUE Phage fragments \\
Coding/Non-coding         & GB Coding/Non-coding \\
Chromatin Accessibility   & iDHS DNase-I \\
\bottomrule
\end{tabular}
\end{table}

\subsection{Probe Stability Analysis}
\label{subsec:probe_stability}

\paragraph{Setup.}
For each of the $11 \times 13 = 143$ model--task combinations, we compute
two MCC values: one using the linear probe (logistic regression) employed
throughout the main paper, and one using a non-linear MLP probe consisting
of a single hidden layer of 256 units with ReLU activation, trained with
early stopping. All other components of the evaluation pipeline -- feature
extraction, normalization, train/test splits, and random seeds -- are held
identical across the two probes. We then compare the resulting rankings
and absolute MCC values to assess whether the linear probe provides a
faithful proxy for representation quality under non-linear readouts.

\paragraph{Rankings are highly stable across probes.}
Spearman rank correlation between the two probes is $\rho = 0.964$
($p < 0.001$) across all 143 model--task pairs, and $\rho = 0.973$
($p < 0.001$) when computed on per-model average MCC. The top-3 and
top-5 models, ranked by mean MCC across the 13 tasks, are identical
under both probes (Table~\ref{tab:probe_rankings}). Per-task Spearman
correlations are positive for 12 of 13 tasks
(Table~\ref{tab:probe_per_task}), with a median of $\rho = 0.855$. The
sole exception is \textsc{GB Ensembl regulatory}, where MCC values
are tightly clustered across models, leaving rank ordering dominated by
stochastic noise rather than substantive differences in representation
quality.

\begin{table}[h]
\centering
\caption{Per-model average MCC under linear and MLP probes across the 13
representative tasks. Models are ordered by linear-probe MCC. The top-5
positions are identical under both probes. The largest absolute
difference (\textsc{HyenaDNA-Large-1M}, $+0.052$) does not alter the
model's relative rank.}
\label{tab:probe_rankings}
\small
\begin{tabular}{@{}lccc@{}}
\toprule
\textbf{Model} & \textbf{Linear MCC} & \textbf{MLP MCC} & $\boldsymbol{\Delta}$ \\
\midrule
\textsc{GENERator-Eukaryote-3B}    & 0.605 & 0.609 & $+0.004$ \\
\textsc{GENERator-Eukaryote-1.2B}  & 0.579 & 0.595 & $+0.016$ \\
\textsc{LucaOne}                    & 0.573 & 0.600 & $+0.027$ \\
\textsc{GenomeOcean-4B}            & 0.552 & 0.552 & $+0.000$ \\
\textsc{Omni-DNA-1B}                & 0.550 & 0.542 & $-0.009$ \\
\textsc{GenomeOcean-500M}          & 0.536 & 0.535 & $-0.001$ \\
\textsc{GENA-LM-Large-T2T}         & 0.530 & 0.535 & $+0.005$ \\
\textsc{MutBERT}                    & 0.516 & 0.517 & $+0.001$ \\
\textsc{NT-v2-50M-MS}              & 0.511 & 0.521 & $+0.010$ \\
\textsc{GROVER}                     & 0.466 & 0.477 & $+0.011$ \\
\textsc{HyenaDNA-Large-1M}         & 0.427 & 0.479 & $+0.052$ \\
\bottomrule
\end{tabular}
\end{table}

\begin{table}[h]
\centering
\caption{Per-task Spearman rank correlation between linear and MLP probe
rankings across the 11 representative models. Tasks are ordered by
correlation strength. Twelve of thirteen tasks exhibit strongly positive
rank correlation. The single negative value
(\textsc{GB Ensembl regulatory}) corresponds to a task where MCC values
are tightly clustered across models, leaving rank ordering dominated by
noise rather than substantive performance differences.}
\label{tab:probe_per_task}
\small
\begin{tabular}{@{}lc@{}}
\toprule
\textbf{Task} & $\boldsymbol{\rho}$ \\
\midrule
GB Coding/Non-coding        & $+0.982$ \\
GUE Phage fragments         & $+0.982$ \\
PGB lncRNA \emph{Z.\ mays}  & $+0.973$ \\
GUE Virus-40                & $+0.945$ \\
GUE EMP H3K4me1             & $+0.891$ \\
GUE Mouse enh.-0            & $+0.891$ \\
GB Human enh.\ (Cohn)       & $+0.855$ \\
NT Splice (all)             & $+0.818$ \\
GUE TF-0                    & $+0.809$ \\
iDHS DNase-I                & $+0.743$ \\
4mC \emph{E.\ coli}         & $+0.673$ \\
iPro GM12878                & $+0.391$ \\
GB Ensembl regulatory       & $-0.082$ \\
\bottomrule
\end{tabular}
\end{table}

\paragraph{Absolute MCC shifts are small.}
The mean signed difference between linear and MLP probe MCC across the
11 models is $+0.011$, indicating only a marginal aggregate benefit from
non-linear readouts. The largest single shift is observed for
\textsc{HyenaDNA-Large-1M} ($+0.052$), which is consistent with the
hypothesis that Hyena representations benefit modestly from non-linear
projection. However, this shift does not change the model's relative
rank, and the overall ordering remains intact.

\paragraph{Conclusion.}
These results indicate that the rankings reported in the main paper are
robust to probe choice within the representative subset evaluated here:
model orderings under linear probing constitute a reliable proxy for
those obtained with non-linear MLP probes. The main empirical
conclusions of GENEB -- including the scale--performance disconnect,
architectural dominance under matched conditions, and category-dependent
specialization -- are therefore unlikely to be artifacts of the linear
probing protocol. We note that the present analysis addresses the
stability of relative comparisons under frozen representations; the
question of whether rankings remain stable under full task-specific
fine-tuning is discussed in the main paper as a limitation.

\subsection{Few-Shot Protocol Sensitivity}
\label{subsec:fewshot_sensitivity}

\paragraph{Setup.}
A second potential source of artifact in our few-shot conclusions is the
choice of regularization strength in the linear probe, particularly in
the low-data 1-shot and 10-shot regimes where logistic regression
behavior can be sensitive to hyperparameter settings. To assess this, we
sweep the inverse regularization strength
$C \in \{0.01, 0.1, 1.0, 10.0, 100.0\}$
across all three shot regimes (1-shot, 10-shot, and full-data) for the
same $11 \times 13$ model--task subset used in
Section~\ref{subsec:probe_stability}. We sweep $C$ as the principal
regularization hyperparameter of logistic regression; other settings
(maximum iterations, solver, convergence tolerance) are held identical
to the main-paper protocol, as are feature normalization, train/test
splits, and random seeds. We then quantify both ranking stability and
absolute MCC sensitivity as functions of $C$.

\paragraph{Rankings are essentially invariant in the 1-shot regime.}
Pairwise Spearman correlations between rankings at different $C$ values
are summarized in Table~\ref{tab:fewshot_summary}. In the 1-shot regime,
mean pairwise $\rho = 0.993$, with a minimum of $0.982$ across all
$\binom{5}{2}=10$ pairs of $C$ values
(Table~\ref{tab:fewshot_1shot}). This near-invariance indicates that
rankings in the most data-constrained regime are dominated by intrinsic
representation quality rather than by regularization choice.

\begin{table}[h]
\centering
\caption{Ranking stability across regularization strengths
$C \in \{0.01, 0.1, 1, 10, 100\}$. Each row reports the mean, minimum,
and maximum pairwise Spearman correlation between model rankings under
different values of $C$ within a given shot regime. Rankings are highly
stable in the 1-shot regime and remain stable for adjacent values of
$C$ in the 10-shot and full-data regimes.}
\label{tab:fewshot_summary}
\small
\begin{tabular}{@{}lccc@{}}
\toprule
\textbf{Regime} & \textbf{Mean pairwise} $\boldsymbol{\rho}$ & \textbf{Min} $\boldsymbol{\rho}$ & \textbf{Max} $\boldsymbol{\rho}$ \\
\midrule
1-shot     & $0.993$ & $0.982$ & $1.000$ \\
10-shot    & $0.805$ & $0.582$ & $0.982$ \\
Full-data  & $0.766$ & $0.436$ & $0.955$ \\
\bottomrule
\end{tabular}
\end{table}

\begin{table}[h]
\centering
\caption{Pairwise Spearman correlation between model rankings at
different regularization strengths in the 1-shot regime. Rankings are
nearly invariant ($\rho \geq 0.982$) across all $C$ pairs.}
\label{tab:fewshot_1shot}
\small
\begin{tabular}{@{}lccccc@{}}
\toprule
 & $C=0.01$ & $C=0.1$ & $C=1$ & $C=10$ & $C=100$ \\
\midrule
$C=0.01$  &  --    & 0.982 & 0.982 & 0.982 & 0.982 \\
$C=0.1$   &       &  --    & 1.000 & 1.000 & 1.000 \\
$C=1$     &       &       &  --    & 1.000 & 1.000 \\
$C=10$    &       &       &       &  --    & 1.000 \\
$C=100$   &       &       &       &       &  --    \\
\bottomrule
\end{tabular}
\end{table}

\paragraph{Rankings remain stable for adjacent $C$ values in the 10-shot regime.}
At 10-shot, mean pairwise $\rho = 0.805$, with adjacent $C$ values
yielding $\rho \geq 0.9$ (Table~\ref{tab:fewshot_10shot}). Larger
divergences appear only between extreme settings (e.g., $C = 0.01$ vs.
$C = 100$, $\rho = 0.582$), reflecting the increased influence of
regularization when small but non-trivial amounts of supervision are
available. The full-data regime exhibits a similar pattern (mean
$\rho = 0.766$; min $\rho = 0.436$), indicating that ranking
fluctuations are driven primarily by extreme regularization choices
rather than by typical hyperparameter selections.

\begin{table}[h]
\centering
\caption{Pairwise Spearman correlation between model rankings at
different regularization strengths in the 10-shot regime. Adjacent
$C$ values yield $\rho \geq 0.9$, with substantial divergence appearing
only between extreme settings.}
\label{tab:fewshot_10shot}
\small
\begin{tabular}{@{}lccccc@{}}
\toprule
 & $C=0.01$ & $C=0.1$ & $C=1$ & $C=10$ & $C=100$ \\
\midrule
$C=0.01$  &  --    & 0.927 & 0.809 & 0.636 & 0.582 \\
$C=0.1$   &       &  --    & 0.900 & 0.727 & 0.664 \\
$C=1$     &       &       &  --    & 0.936 & 0.891 \\
$C=10$    &       &       &       &  --    & 0.982 \\
$C=100$   &       &       &       &       &  --    \\
\bottomrule
\end{tabular}
\end{table}

\paragraph{Absolute MCC sensitivity is bounded and concentrated in a small subset of models.}
Per-model MCC ranges across $C$ values in the full-data regime are
reported in Table~\ref{tab:fewshot_ranges}. For most models, the range
is well below $0.10$~MCC, with the smallest sensitivity observed for
\textsc{GenomeOcean-500M} (range $0.014$). The largest sensitivities
are observed for \textsc{LucaOne} ($0.199$),
\textsc{HyenaDNA-Large-1M} ($0.144$), and \textsc{NT-v2-50M-MS}
($0.108$), suggesting that representations from these models are
somewhat more dependent on regularization tuning than those of the
remaining models. We note that absolute MCC sensitivity does not
translate into ranking instability for typical regularization choices:
across adjacent $C$ values, the relative ordering of models remains
substantively unchanged
(Tables~\ref{tab:fewshot_1shot} and~\ref{tab:fewshot_10shot}).

\begin{table}[h]
\centering
\caption{Per-model MCC range across regularization strengths
$C \in \{0.01, 0.1, 1, 10, 100\}$ in the full-data regime. The largest
sensitivity is observed for \textsc{LucaOne}, \textsc{HyenaDNA-Large-1M},
and \textsc{NT-v2-50M-MS}; the smallest for \textsc{GenomeOcean-500M}.}
\label{tab:fewshot_ranges}
\small
\begin{tabular}{@{}lccc@{}}
\toprule
\textbf{Model} & \textbf{Min MCC} & \textbf{Max MCC} & \textbf{Range} \\
\midrule
\textsc{GenomeOcean-500M}          & 0.523 & 0.537 & 0.014 \\
\textsc{GROVER}                     & 0.457 & 0.473 & 0.017 \\
\textsc{GENERator-Eukaryote-1.2B}  & 0.563 & 0.583 & 0.021 \\
\textsc{GENERator-Eukaryote-3B}    & 0.572 & 0.603 & 0.031 \\
\textsc{GenomeOcean-4B}            & 0.517 & 0.557 & 0.040 \\
\textsc{GENA-LM-Large-T2T}         & 0.477 & 0.529 & 0.053 \\
\textsc{Omni-DNA-1B}                & 0.483 & 0.550 & 0.067 \\
\textsc{MutBERT}                    & 0.452 & 0.527 & 0.076 \\
\textsc{NT-v2-50M-MS}              & 0.404 & 0.512 & 0.108 \\
\textsc{HyenaDNA-Large-1M}         & 0.318 & 0.462 & 0.144 \\
\textsc{LucaOne}                    & 0.415 & 0.614 & 0.199 \\
\bottomrule
\end{tabular}
\end{table}

\paragraph{The principal few-shot finding is protocol-stable.}
Crucially, the central few-shot conclusion reported in the main
paper -- the sharp degradation of mean MCC from full-data to 1-shot -- is
replicated at every value of $C$ tested. The magnitude of this
degradation varies modestly across regularization choices, and its
direction and severity are preserved without exception.

\paragraph{Conclusion.}
These results indicate that the few-shot findings reported in the main
paper reflect properties of the evaluated representations under
data-constrained regimes, rather than artifacts of a particular
regularization choice. Rankings are essentially invariant under 1-shot
evaluation, remain stable for typical regularization settings under
10-shot and full-data evaluation, and the qualitative pattern of severe
low-data degradation persists uniformly across the regularization sweep.

\subsection{Controlled-Pair Comparisons and Residual Confounds}
\label{subsec:controlled_pairs}

\paragraph{Methodology.}
Throughout the main paper, comparative claims about architecture,
tokenization, and pretraining data are based on \emph{matched pairs} of
models that differ in exactly one factor of interest while holding others
as constant as the available model set permits. This controlled-pair
design substantially reduces the risk of attributing observed performance
differences to the wrong cause and is preferable to unmatched comparisons
across the full benchmark. However, perfect isolation is impossible in
practice: genomic foundation models differ along multiple correlated
axes (architecture, tokenization, training corpus, scale, training
duration, and pretraining objective), so even carefully matched pairs
retain residual confounds. To make these confounds explicit and to enable
readers to assess the strength of each claim, we enumerate the full set
of $31$ controlled pairs underlying the comparative analyses.

\begin{table}[h]
\centering
\caption{Summary of the $31$ controlled-pair comparisons used in the
analyses of architecture, pretraining data, and tokenization. Each pair
varies a single factor of interest while holding the others constant.}
\label{tab:pairs_summary}
\small
\begin{tabular}{@{}lc@{}}
\toprule
\textbf{Comparison Type} & \textbf{Pairs} \\
\midrule
\textbf{A. Architecture}                                  & 11 \\
\quad Decoder vs.\ Encoder                                & 6 \\
\quad Decoder vs.\ Mamba                                  & 3 \\
\quad Encoder vs.\ Graph-Transformer                      & 2 \\
\midrule
\textbf{B. Pretraining Data}                              & 9 \\
\quad Human vs.\ Multi-species                            & 6 \\
\quad Multi-species vs.\ Microbial                        & 2 \\
\quad Eukaryotic-genes vs.\ Multi-species                 & 1 \\
\midrule
\textbf{C. Tokenization}                                  & 11 \\
\quad BPE vs.\ $k$-mer                                    & 6 \\
\quad Single-nucleotide vs.\ BPE                          & 2 \\
\quad Other                                               & 3 \\
\midrule
\textbf{Total}                                            & \textbf{31} \\
\bottomrule
\end{tabular}
\end{table}

\paragraph{Summary of comparisons.}
Table~\ref{tab:pairs_summary} provides an overview of the matched pairs
by factor type. Architecture comparisons (11 pairs) hold tokenization and
pretraining corpus type constant while varying the architectural family.
Pretraining-data comparisons (9 pairs) hold architecture and tokenization
constant while varying the corpus type. Tokenization comparisons (11 pairs)
hold architecture and pretraining corpus constant while varying the
tokenization scheme.

\paragraph{Architecture comparisons.}
Table~\ref{tab:pairs_arch} enumerates the $11$ matched pairs used to
isolate architectural effects. In each pair, tokenization and
pretraining corpus type are held constant; the varied factor is
architectural family. For each pair we report the macro-averaged MCC
difference $\Delta = \text{MCC}_A - \text{MCC}_B$; positive values
indicate that the architectural family listed first in the ``Variable''
column wins. Residual confounds for this group of comparisons include
model size differences (constrained to within a factor of two where
possible), training duration, exact composition of the pretraining
data, pretraining objective (masked language modeling vs.\ causal
language modeling), and depth-to-width ratio.

\begin{table}[h]
\centering
\caption{Architecture-controlled pairs. Each pair holds tokenization
and pretraining corpus type constant and varies the architectural
family. $\Delta$~macro-MCC is the difference between model A and
model B; positive values indicate that the architectural family
listed first in the ``Variable'' column wins.}
\label{tab:pairs_arch}
\small
\begin{tabular}{@{}llllc@{}}
\toprule
\textbf{Model A} & \textbf{Model B} & \textbf{Controlled} & \textbf{Variable} & \textbf{$\Delta$} \\
\midrule
\textsc{Omni-DNA-1B}        & \textsc{eccDNAMamba}       & BPE, multi-species     & Decoder vs.\ Mamba         & $+0.149$ \\
\textsc{GenomeOcean-500M}   & \textsc{eccDNAMamba}       & BPE, multi-species     & Decoder vs.\ Mamba         & $+0.131$ \\
\textsc{Omni-DNA-300M}      & \textsc{eccDNAMamba}       & BPE, multi-species     & Decoder vs.\ Mamba         & $+0.119$ \\
\textsc{GenomeOcean-500M}   & \textsc{GENA-LM-Large-T2T} & BPE, multi-species     & Decoder vs.\ Encoder       & $-0.002$ \\
\textsc{Omni-DNA-300M}      & \textsc{GENA-LM-Large-T2T} & BPE, multi-species     & Decoder vs.\ Encoder       & $-0.014$ \\
\textsc{DNA-GPT-3B-M}       & \textsc{NT-2.5B-MS}        & $k$-mer, multi-species & Decoder vs.\ Encoder       & $-0.031$ \\
\textsc{DNA-GPT-0.1B-H}     & \textsc{NT-v2-100M-MS}     & $k$-mer, multi-species & Decoder vs.\ Encoder       & $-0.040$ \\
\textsc{DNA-GPT-0.1B-H}     & \textsc{Genomics-FM}       & $k$-mer, multi-species & Decoder vs.\ Encoder       & $-0.047$ \\
\textsc{OmniNA-220M}        & \textsc{GENA-LM-Large-T2T} & BPE, multi-species     & Decoder vs.\ Encoder       & $-0.127$ \\
\textsc{GROVER}             & \textsc{DeepGene}          & BPE, human             & Encoder vs.\ Graph-Trans.  & $+0.061$ \\
\textsc{DeepGene}           & \textsc{GENA-LM}           & BPE, human             & Graph-Trans.\ vs.\ Encoder & $-0.066$ \\
\bottomrule
\end{tabular}
\end{table}

\paragraph{Pretraining-data comparisons.}
Table~\ref{tab:pairs_pretraining} enumerates the $9$ matched pairs
used to isolate the effect of pretraining corpus type. In each pair,
architecture and tokenization are held constant; the varied factor is
the taxonomic composition of the pretraining data. For each pair we
report the macro-averaged MCC difference $\Delta = \text{MCC}_A -
\text{MCC}_B$; positive values indicate that the corpus type listed
first in the ``Variable'' column wins. Residual confounds include
exact corpus scale and diversity, training duration, and learning-rate
schedules. We additionally note that the \emph{Eukaryotic-genes vs.\
Multi-species} comparison is supported by a single matched pair
(\textsc{GENERator-Eukaryote-3B} vs.\ \textsc{DNA-GPT-3B-M}); claims
derived from this comparison are accordingly flagged in the main text
and should be interpreted with additional caution.

\begin{table}[h]
\centering
\caption{Pretraining-data-controlled pairs. Each pair holds
architecture and tokenization constant and varies the pretraining
corpus type. $\Delta$~macro-MCC is the difference between model A and
model B; positive values indicate the corpus type listed first in
``Variable'' wins.}
\label{tab:pairs_pretraining}
\small
\begin{tabular}{@{}llllc@{}}
\toprule
\textbf{Model A} & \textbf{Model B} & \textbf{Controlled} & \textbf{Variable} & \textbf{$\Delta$} \\
\midrule
\textsc{Genomics-FM}            & \textsc{DNABERT-S}          & Encoder, $k$-mer  & Multi vs.\ Microbial    & $+0.088$ \\
\textsc{NT-v2-100M-MS}          & \textsc{DNABERT-S}          & Encoder, $k$-mer  & Multi vs.\ Microbial    & $+0.081$ \\
\midrule
\textsc{GENERator-Eukaryote-3B} & \textsc{DNA-GPT-3B-M}       & Decoder, $k$-mer  & Eukaryotic vs.\ Multi   & $+0.063$ \\
\midrule
\textsc{DNABERT-2}              & \textsc{GENA-LM}            & Encoder, BPE      & Multi vs.\ Human        & $+0.014$ \\
\textsc{GENA-LM-T2T-Multi}      & \textsc{GENA-LM}            & Encoder, BPE      & Multi vs.\ Human        & $+0.025$ \\
\textsc{DNABERT-2}              & \textsc{GROVER}             & Encoder, BPE      & Multi vs.\ Human        & $+0.019$ \\
\textsc{GENA-LM-T2T-Multi}      & \textsc{GROVER}             & Encoder, BPE      & Multi vs.\ Human        & $+0.030$ \\
\textsc{Omni-DNA-300M}          & \textsc{GPT2-Gene-Multi-v2} & Decoder, BPE      & Multi vs.\ Human        & $+0.071$ \\
\textsc{OmniNA-220M}            & \textsc{GPT2-Gene-v1}       & Decoder, BPE      & Multi vs.\ Human        & $-0.051$ \\
\bottomrule
\end{tabular}
\end{table}

\paragraph{Tokenization comparisons.}
\label{subsec:tokenization_controlled}
Table~\ref{tab:pairs_tokenization} enumerates the $11$ matched pairs
used to isolate the effect of tokenization scheme. In each pair,
architecture and pretraining corpus type are held constant; the varied
factor is the tokenization strategy. For each pair we report the
macro-averaged MCC difference $\Delta = \text{MCC}_A -
\text{MCC}_B$, where positive values indicate that the tokenization
listed first in the ``Variable'' column wins. Residual confounds include
model size differences (within $\pm 2\times$), vocabulary size (tightly
coupled with tokenization scheme), and model-specific training regimes.

Aggregating across the 11 pairs reveals three regimes. (i)~In the
single available matched Transformer-decoder comparison with
multi-species pretraining, BPE exceeds $k$-mer by $+0.032$
(\textsc{GenomeOcean-4B} vs.\ \textsc{DNA-GPT-3B-M}).
(ii)~In matched Transformer-encoder comparisons with multi-species
pretraining, BPE and $k$-mer perform comparably ($+0.006$ across 5
pairs, with all gaps within $\pm 0.02$ MCC).
(iii)~In matched Transformer-encoder comparisons with human
pretraining, single-nucleotide tokenization (\textsc{MutBERT})
consistently outperforms BPE baselines ($+0.033$ and $+0.038$ over
\textsc{GENA-LM} and \textsc{GROVER}, respectively). The three
comparisons involving non-standard tokenization schemes
(\textsc{BioFM-265M}'s BioToken; \textsc{LucaOne}'s mixed vocabulary)
are reported for completeness but should be interpreted with caution,
as these schemes introduce confounds beyond a simple choice of
tokenization unit.

\begin{table}[t]
\centering
\caption{Tokenization-controlled pairs. Each pair holds architecture
and pretraining corpus type constant and varies the tokenization
scheme. $\Delta$~macro-MCC is the difference between model A and
model B; positive values indicate that the tokenization listed first
in the ``Variable'' column wins.}
\label{tab:pairs_tokenization}
\small
\begin{tabular}{@{}llllc@{}}
\toprule
\textbf{Model A} & \textbf{Model B} & \textbf{Controlled} & \textbf{Variable} & \textbf{$\Delta$} \\
\midrule
\textsc{GenomeOcean-4B}     & \textsc{DNA-GPT-3B-M}       & Decoder, multi-species & BPE vs.\ $k$-mer        & $+0.032$ \\
\midrule
\textsc{GENA-LM-Large-T2T}  & \textsc{NT-v2-250M-MS}      & Encoder, multi-species & BPE vs.\ $k$-mer        & $+0.015$ \\
\textsc{GENA-LM-T2T-Multi}  & \textsc{NT-v2-100M-MS}      & Encoder, multi-species & BPE vs.\ $k$-mer        & $+0.013$ \\
\textsc{GENA-LM-T2T-Multi}  & \textsc{Genomics-FM}        & Encoder, multi-species & BPE vs.\ $k$-mer        & $+0.006$ \\
\textsc{DNABERT-2}          & \textsc{NT-v2-100M-MS}      & Encoder, multi-species & BPE vs.\ $k$-mer        & $+0.002$ \\
\textsc{DNABERT-2}          & \textsc{Genomics-FM}        & Encoder, multi-species & BPE vs.\ $k$-mer        & $-0.005$ \\
\midrule
\textsc{MutBERT}            & \textsc{GENA-LM}            & Encoder, human         & Single-nt vs.\ BPE      & $+0.033$ \\
\textsc{GROVER}             & \textsc{MutBERT}            & Encoder, human         & BPE vs.\ Single-nt      & $-0.038$ \\
\midrule
\textsc{GPT2-Gene-v1}       & \textsc{BioFM-265M}         & Decoder, human         & BPE vs.\ Other          & $+0.134$ \\
\textsc{GPT2-Gene-Multi-v2} & \textsc{BioFM-265M}         & Decoder, human         & BPE vs.\ Other          & $+0.125$ \\
\textsc{LucaOne}            & \textsc{NT-2.5B-MS}         & Encoder, multi-species & Other vs.\ $k$-mer      & $+0.017$ \\
\bottomrule
\end{tabular}
\end{table}

\paragraph{Conclusion.}
Throughout the main paper, observations derived from these matched pairs
are framed as \emph{consistent with} or \emph{associated with} the
varied factor, rather than as \emph{caused by} it. Single-pair
comparisons -- most notably the \emph{Eukaryotic-genes vs.\ Multi-species}
contrast -- are flagged as such on first appearance, and conclusions
relying on them are stated with correspondingly reduced confidence. We
view this controlled-pair methodology as a principled middle ground
between unmatched comparisons across the full benchmark, which conflate
multiple factors, and fully randomized causal experiments, which are
infeasible at the scale of foundation-model pretraining.

\subsection{Micro- vs.\ Macro-Averaged Aggregate Performance}
\label{subsec:macro_micro}

The main paper reports macro-averaged MCC as the principal aggregation
metric, computed by first averaging within each of the 13 functional
categories and then averaging across categories with equal weight per
category. This choice avoids implicitly overweighting categories with
many tasks (e.g., histone modifications with 30 tasks, promoters with
22 tasks) and instead treats each functional category as a unit of
biological interest. To verify that the central findings of GENEB are
robust to this weighting choice, we additionally compute
\emph{micro-averaged} MCC by simple averaging across all 100 tasks and
compare the two aggregation schemes.

\begin{figure}[H]
    \centering
    \includegraphics[width=0.95\linewidth]{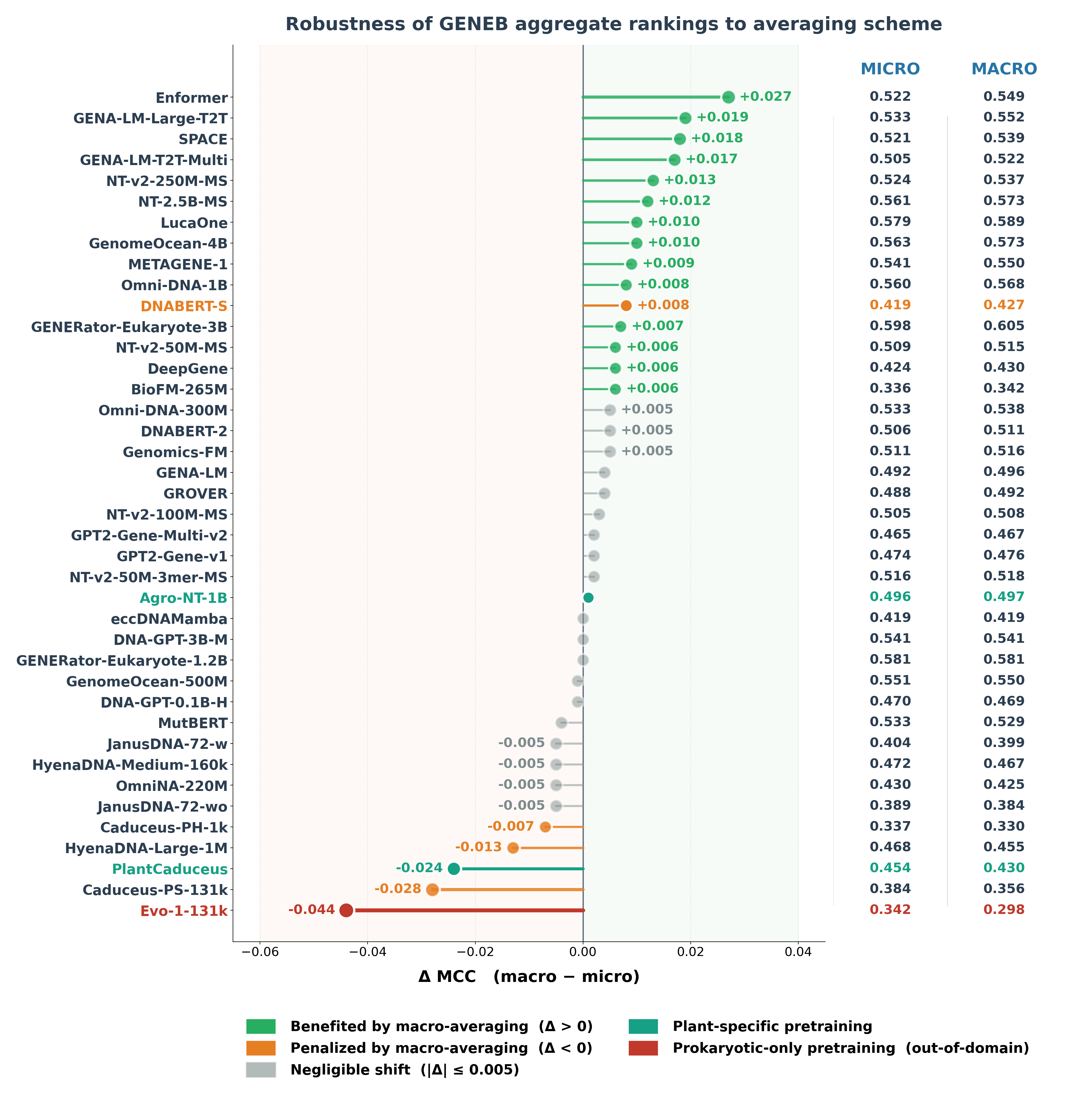}
    \caption{\textbf{Robustness of GENEB aggregate rankings to averaging
    scheme.} For each of the 40 models, $\Delta = \text{MCC}_{\text{macro}} - \text{MCC}_{\text{micro}}$
    is shown in the left panel; the side panel reports the underlying
    micro- and macro-averaged MCC values. Models are sorted from largest
    negative shift to largest positive shift. Out-of-domain models are
    highlighted: prokaryotic-only \textsc{Evo-1-131k} (red,
    $\Delta = -0.044$), microbial-only \textsc{DNABERT-S} (orange,
    $\Delta = +0.008$), and plant-specific \textsc{PlantCaduceus} and
    \textsc{Agro-NT-1B} (teal). Across all 40 models the Spearman rank
    correlation between micro- and macro-averaged orderings is
    $\rho = 0.988$ ($p < 0.001$); the mean absolute shift is
    $|\Delta| = 0.009$ MCC.}
    \label{fig:micro_macro}
\end{figure}

Figure~\ref{fig:micro_macro} compares the two aggregation schemes
across all 40 models. The Spearman rank correlation between micro-
and macro-averaged orderings is $\rho = 0.988$ ($p < 0.001$),
indicating that the relative ordering of models is largely preserved.
The mean absolute shift in aggregate MCC is $|\Delta| = 0.009$, and
the top-5 set is identical under both schemes (within the top-5,
\textsc{LucaOne} and \textsc{GENERator-Eukaryote-1.2B} swap positions).
The largest individual shifts are concentrated in out-of-domain or
specialized models: \textsc{Evo-1-131k} ($\Delta = -0.044$),
\textsc{Caduceus-PS-131k} ($\Delta = -0.028$), and
\textsc{PlantCaduceus} ($\Delta = -0.024$), all of which exhibit
highly uneven category-level performance that is amplified under
category-balanced averaging.

The central empirical findings of GENEB -- the substantial overall
correlation between scale and performance, the instability of category
rankings, and the dominance of architectural and pretraining alignment
over parameter count -- hold under both averaging schemes. We view the
sensitivity of out-of-domain models to weighting choice as itself
consistent with the broader argument of this paper: single-score
leaderboards are an unreliable basis for genomic model selection.

\newpage

\section{Results Analysis}
\label{app:results}

This section presents a systematic analysis of the benchmark results
across 40 DNA foundation models evaluated on 100 genomic tasks
spanning 13 functional categories. We structure our analysis around
three fundamental questions: (1) how do architectural choices influence
model performance, (2) what role does tokenization strategy play, and
(3) how does model scale interact with pretraining design decisions?

\subsection{Experimental Overview}
\label{subsec:overview}

Our benchmark encompasses substantial diversity in both models and
tasks. The 40 evaluated models span the following architectural
families: Transformer-encoder ($n=15$), Transformer-decoder ($n=13$),
Mamba ($n=4$), Hybrid-Mamba-MoE ($n=2$), Hyena ($n=2$),
CNN-Transformer hybrids ($n=2$, including \textsc{Enformer} and
\textsc{SPACE}), Graph-Transformer ($n=1$, \textsc{DeepGene}), and
StripedHyena ($n=1$, \textsc{Evo-1-131k}). Model sizes range from
under $2$M to $7$B parameters, with tokenization strategies including
BPE ($n=15$), $k$-mer ($n=11$), single-nucleotide ($n=11$), and three
custom schemes (\textsc{BioFM-265M}'s BioToken, \textsc{LucaOne}'s
mixed nucleotide--amino-acid vocabulary, and \textsc{Genomics-FM}'s
ensemble of BPE and $k$-mer). Pretraining corpora vary across
multi-species genomes ($n=19$), human-only ($n=13$), eukaryotic gene
sequences ($n=2$), plant genomes ($n=2$), human-mouse epigenomic
profiles ($n=2$), prokaryotic sequences ($n=1$), and multi-species
microbial genomes ($n=1$).

The 100 downstream tasks are organized into 13 categories reflecting
distinct aspects of genomic function: Histone Modifications (30 tasks),
Promoters (22), Enhancers (8), DNA Methylation (8), Splice Sites (7),
lncRNA (6), Mouse Enhancers (5), TF Binding (5), Species
Classification (3), Regulatory (2), Virus/Phage (2),
Coding/Non-coding (1), and Chromatin Accessibility (1). The full task
taxonomy is given in Appendix~\ref{subsec:task_taxonomy}. This
distribution enables fine-grained analysis of model capabilities
across regulatory, structural, and evolutionary prediction problems.

\subsection{Aggregate Performance Patterns}
\label{subsec:aggregate}

\begin{figure}[H]
    \centering
    \includegraphics[width=0.9\linewidth]{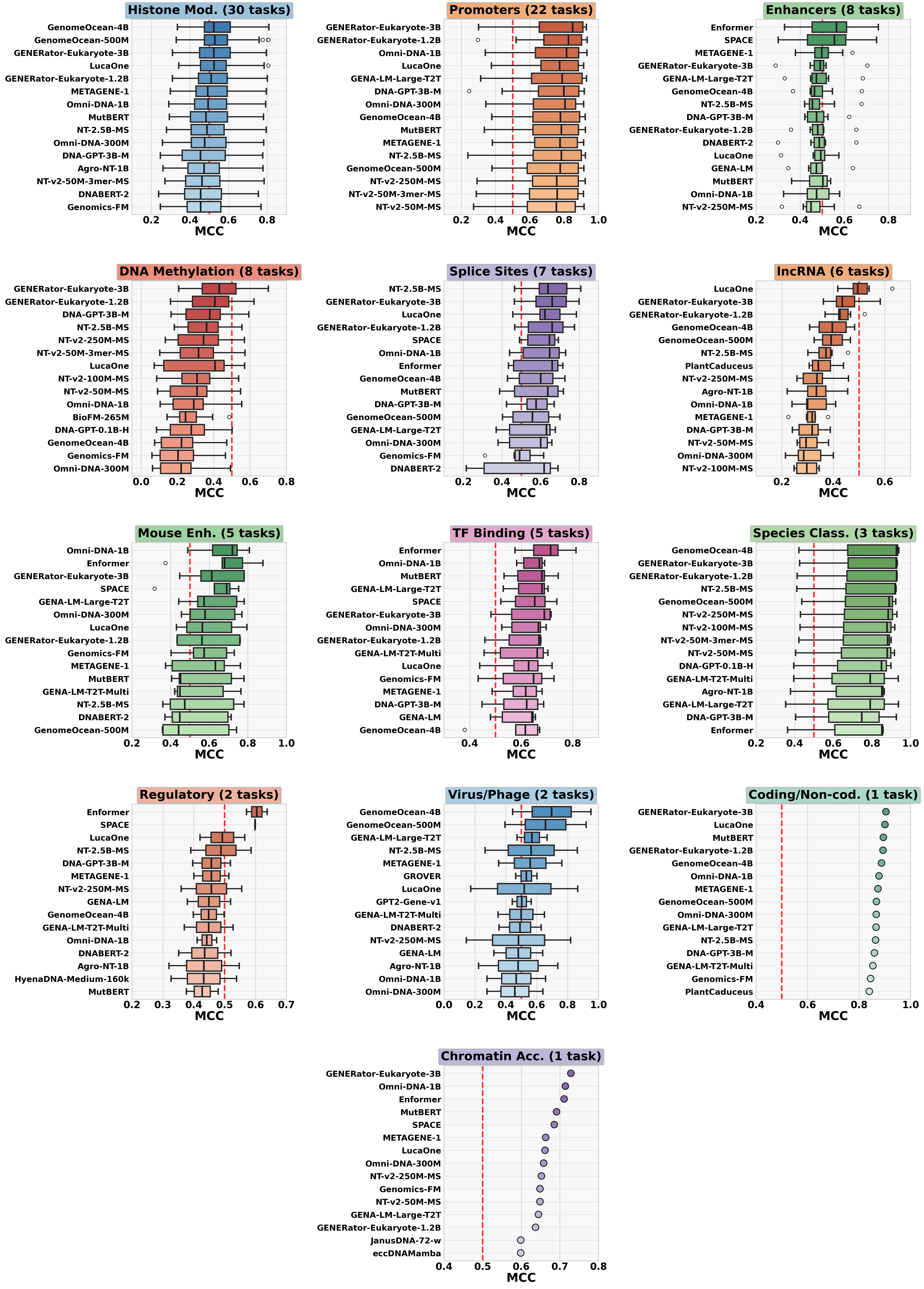}
\caption{\textbf{Category-wise MCC distributions across top-performing models.}
For each functional category, the figure reports the top-15 models ranked by mean MCC. 
Boxplots show the distribution of per-task MCC values within the category: boxes denote the interquartile range, central lines indicate medians, whiskers show the non-outlier range, and points mark outlier tasks. 
For single-task categories, individual MCC values are shown. 
The dashed red line marks MCC $=0.5$.}
    \label{fig:top10_boxplot}
\end{figure}

\begin{figure}[H]
    \centering
    \includegraphics[width=1\linewidth]{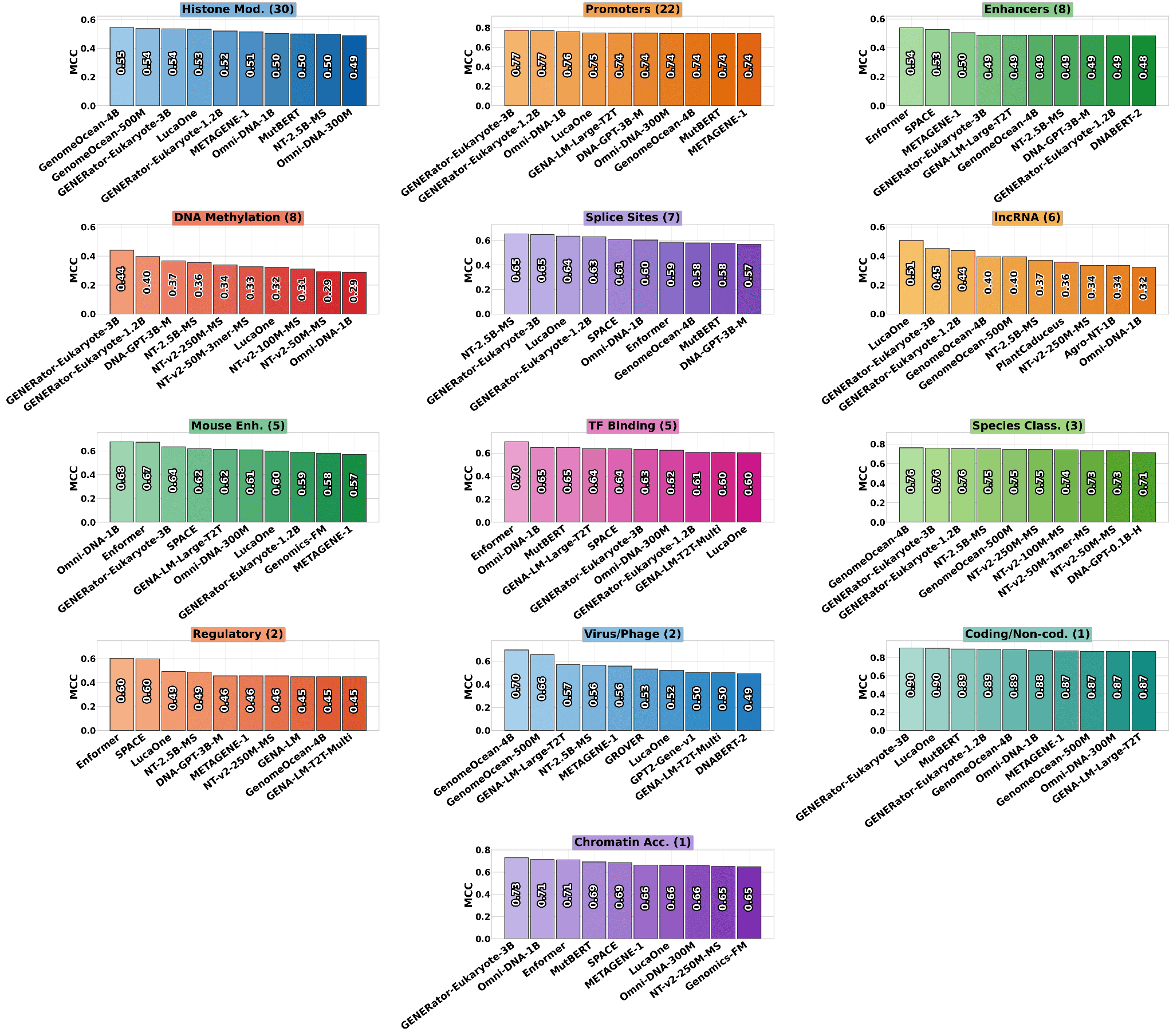}
\caption{\textbf{Top-10 model performance across task categories.}
Mean MCC is shown for the 10 best-performing models within each of the 13 functional task categories.
Models are ranked independently within each category by category-level mean MCC, highlighting task-specific leaders and performance differences across genomic prediction settings.}
\label{fig:top10_category_bars}
    \label{fig:top10_barplot}
\end{figure}

\begin{figure}[h]
    \centering
    \includegraphics[width=0.8\linewidth]{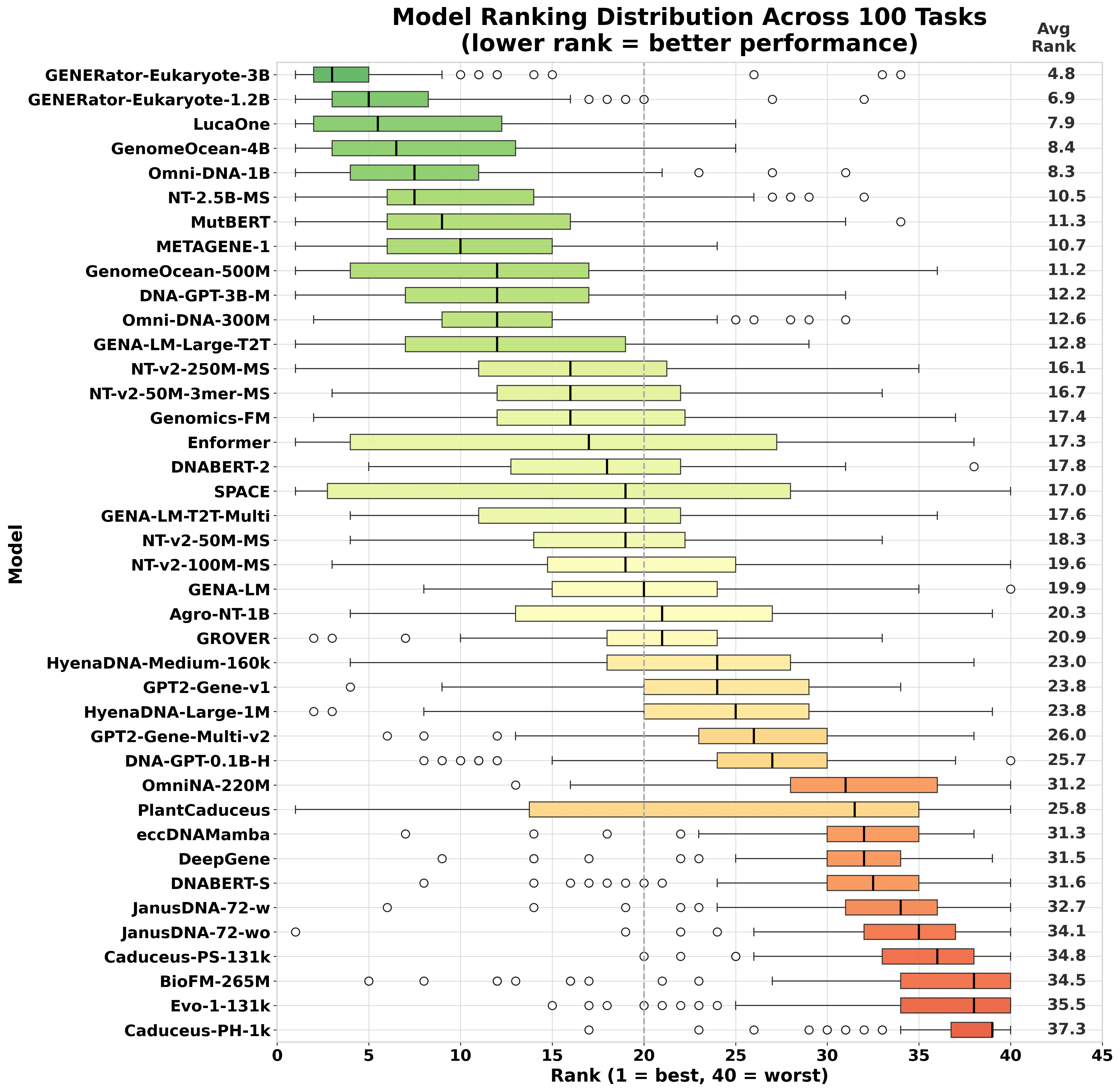}
\caption{\textbf{Distribution of model ranks across benchmark tasks.}
For each model, the boxplot shows the distribution of its task-level ranks across 100 benchmark tasks, where lower rank indicates better performance. 
Models are ordered by median rank, and the right column reports the average rank across all tasks. 
The leading models combine low median rank with relatively compact rank distributions, indicating consistently strong cross-task performance. 
In contrast, models with wide interquartile ranges or numerous outliers show substantial task-dependent variability, suggesting that aggregate performance can obscure pronounced category- and task-specific strengths.}
    \label{fig:rank_distribution}
\end{figure}

\begin{figure}[h]
    \centering
    \includegraphics[width=0.8\linewidth]{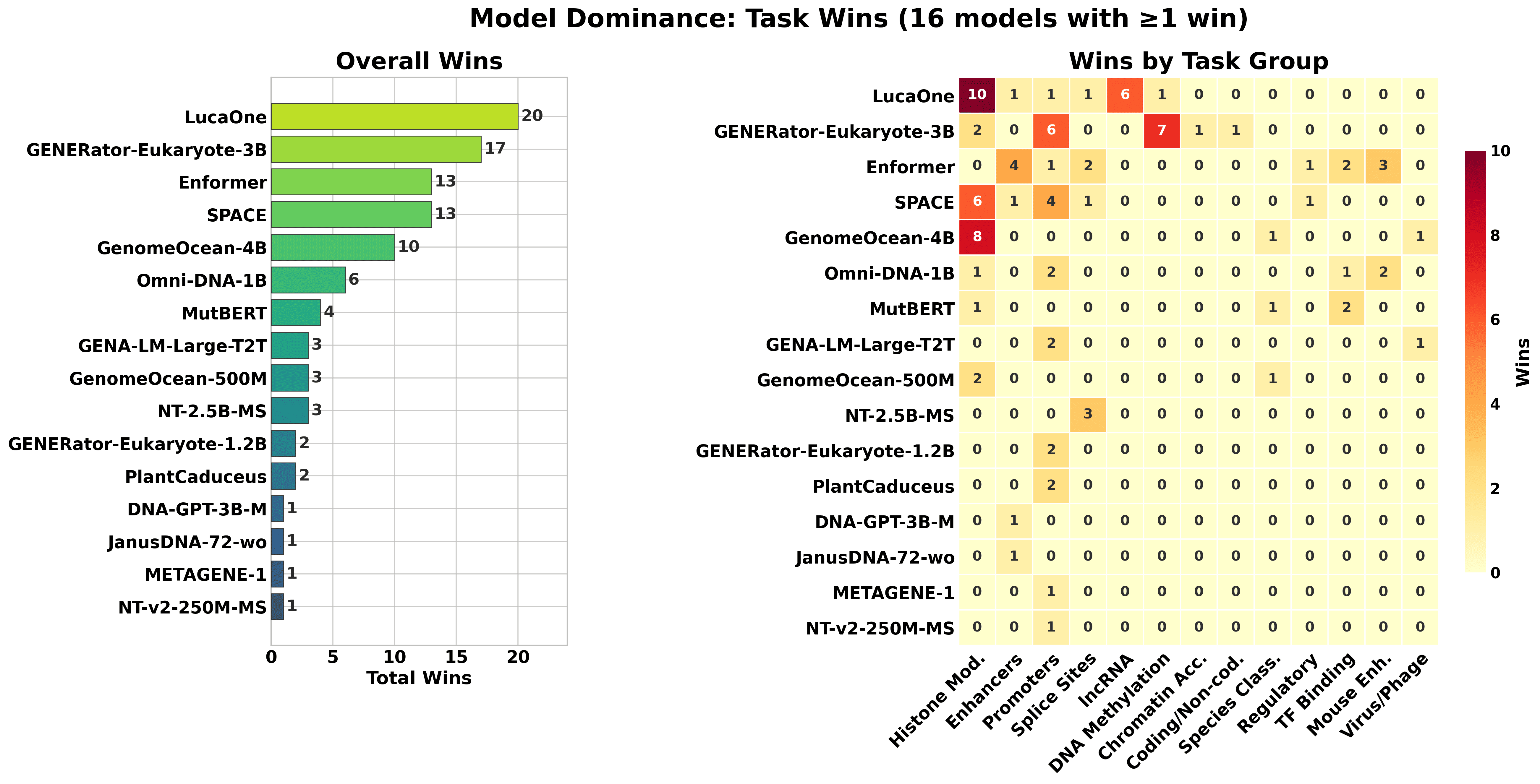}
\caption{\textbf{Distribution of task-level wins across models.}
The figure reports the number of benchmark tasks, out of 100, on which each model achieves the highest MCC. 
Only models with at least one task-level win are included. 
The left panel summarizes total wins per model, while the right panel decomposes these wins by functional task category. 
The dispersed pattern of wins across models and categories indicates that benchmark performance is strongly task-dependent and that no single model dominates uniformly across genomic prediction settings.}
\label{fig:task_wins}
    \label{fig:model_wins}
\end{figure}

Before examining specific architectural and design factors, we first
characterize overall model performance across the benchmark.
Figure~\ref{fig:top10_boxplot} provides boxplot distributions of MCC
scores within each task category for the top-15 models per category,
illustrating that performance variance differs substantially across
categories. 
Figure~\ref{fig:top10_barplot} presents mean MCC scores for the top-10
models across all 13 task categories, revealing substantial variation
in category-specific performance even among leading models.
Figure~\ref{fig:rank_distribution} displays the distribution of model
rankings across all 100 tasks, highlighting that even top-ranked
models exhibit considerable variance in per-task rankings. 
Finally, Figure~\ref{fig:model_wins} shows the distribution of task-level wins
(achieving the highest MCC on a given task) and indicates that no
single model dominates the benchmark: the top model wins only $20$ of
$100$ tasks, with the remaining wins distributed across $15$ additional
models. This fragmentation underscores the importance of task-aware
model selection.


\subsection{Performance Landscape by Model Capacity}
\label{subsec:capacity}

We stratify models into four capacity tiers to examine the
relationship between parameter count and downstream performance.
Tiny models ($<$100M parameters, $n=11$) achieve a mean macro-MCC of
$0.443$, with \textsc{MutBERT} (86M) leading the tier at $0.529$.
Small models (100M--500M, $n=16$) show modest improvement to $0.485$
mean macro-MCC, topped by \textsc{GENA-LM-Large-T2T} ($0.552$) and
\textsc{Enformer} ($0.549$). Medium-scale models (500M--2B, $n=6$)
reach $0.526$ mean macro-MCC, with \textsc{GENERator-Eukaryote-1.2B}
achieving the tier-best $0.581$. Large models ($\geq 2$B, $n=7$)
attain $0.533$ mean macro-MCC, led by \textsc{GENERator-Eukaryote-3B}
at $0.605$.

These tier-level statistics reveal a substantial scaling pattern:
the Spearman correlation between $\log_{10}(\text{parameter count})$
and macro-MCC is $\rho = 0.573$ ($p < 0.001$; see
Section~\ref{sec:results} and Figure~\ref{fig:frontier}). Models
above $1$B parameters achieve mean macro-MCC of $0.538$ compared to
$0.463$ for models below $200$M, a gap of $+0.075$ macro-MCC. While
this association is statistically robust, it does not preclude
substantial within-tier variation: as shown in
Section~\ref{sec:results}, $25$ in-domain models demonstrate cases
where a model at least $5\times$ smaller outperforms a larger
counterpart, indicating that scale alone is not a sufficient
predictor of category-level performance.

\subsection{Architecture Comparison via Controlled Experiments}
\label{subsec:architecture}

To isolate architectural effects from confounding factors, we identify
model pairs sharing pretraining data and tokenization strategy while
differing in architecture. The full enumeration of controlled pairs
is given in Appendix~\ref{subsec:controlled_pairs}. This controlled
comparison reveals consistent patterns favoring attention-based
architectures over the evaluated state-space alternative under matched
conditions.

\textsc{Omni-DNA-1B} (Transformer-decoder) outperforms
\textsc{eccDNAMamba} (Mamba) by $+0.149$ macro-MCC
($0.568$ vs.\ $0.419$), with both models trained on multi-species
data using BPE tokenization. A second matched pair shows the same
direction: \textsc{GenomeOcean-500M} (Transformer-decoder) exceeds
\textsc{eccDNAMamba} by $+0.131$ macro-MCC ($0.550$ vs.\ $0.419$)
under identical pretraining and tokenization conditions. This pattern,
observed across two independent matched pairs, is consistent with
attention-based context modeling providing benefits over the evaluated
state-space architecture in this setting.

Within the Transformer family, encoder models exceed decoders across
all six matched Decoder-vs-Encoder pairs (Appendix~\ref{subsec:controlled_pairs},
Table~\ref{tab:pairs_arch}), with the strongest gap of $+0.127$ macro-MCC
for \textsc{GENA-LM-Large-T2T} over \textsc{OmniNA-220M} ($0.552$ vs.\
$0.425$) under matched multi-species/BPE conditions. However, as
discussed in Section~\ref{sec:results}, the per-category
encoder--decoder ranking is task- and setting-dependent.

\subsection{Tokenization Strategy Effects}
\label{subsec:tokenization-app}

Tokenization represents a fundamental design choice that determines
how nucleotide sequences are discretized for model consumption. Our
controlled comparisons reveal architecture- and setting-dependent
effects rather than a universally optimal strategy
(Appendix~\ref{subsec:tokenization_controlled}).

\textsc{GPT2-Gene-v1} (BPE) exceeds \textsc{BioFM-265M} (BioToken
custom scheme) by $+0.134$ macro-MCC ($0.476$ vs.\ $0.342$), with both
sharing a Transformer-decoder architecture and human pretraining; the
same pattern holds for \textsc{GPT2-Gene-Multi-v2} (BPE) over
\textsc{BioFM-265M} at $+0.125$ macro-MCC. Under matched multi-species
pretraining and Transformer-decoder architecture, the only available
matched pair shows BPE exceeding $k$-mer: \textsc{GenomeOcean-4B}
exceeds \textsc{DNA-GPT-3B-M} by $+0.032$ macro-MCC. Under matched
human pretraining, single-nucleotide tokenization (\textsc{MutBERT})
exceeds both BPE baselines, by $+0.038$ over \textsc{GROVER} and
$+0.033$ over \textsc{GENA-LM}.

These mixed results indicate that tokenization interacts with other
design choices -- particularly pretraining data composition and model
architecture -- in ways that preclude a single global ordering. The
broader picture, including all 11 tokenization-controlled pairs, is
discussed in Section~\ref{sec:results} and enumerated in
Appendix~\ref{subsec:tokenization_controlled}.

\subsection{The Scale--Performance Paradox}
\label{subsec:paradox}

A striking finding emerges from cross-scale comparisons: numerous
smaller models substantially outperform larger counterparts. Among
the $36$ in-domain models (excluding prokaryotic-only, microbial-only,
and plant-specific pretraining), we identify $25$ instances where a
model achieves superior aggregate macro-MCC despite being at least
$5\times$ smaller than its comparison target. Within the full
40-model set, the count rises to $67$, reflecting the additional
contribution of cross-domain cases.

The most dramatic example involves \textsc{MutBERT} (86M parameters,
macro-MCC $= 0.529$) outperforming \textsc{Evo-1-131k} (7B parameters,
macro-MCC $= 0.298$) by $+0.231$ macro-MCC despite an
$81\times$ size disadvantage. The pattern repeats across cross-domain
comparisons with \textsc{Evo-1-131k} as the larger model:
\textsc{Omni-DNA-300M} exceeds \textsc{Evo-1-131k} by
$+0.240$ macro-MCC at a $23\times$ size differential;
\textsc{GenomeOcean-500M} achieves a $+0.252$ macro-MCC advantage at
$14\times$ smaller scale; and \textsc{Omni-DNA-1B} outperforms by
$+0.270$ macro-MCC at $7\times$ fewer parameters. In-domain
reversals (architectural rather than domain-driven) are illustrated
in the main paper by \textsc{MutBERT} outperforming \textsc{eccDNAMamba}
by $+0.110$ macro-MCC at a $6.2\times$ size ratio
(Section~\ref{sec:results}).

The cross-domain reversals are best interpreted as evidence of
domain mismatch rather than as a universal critique of scaling.
\textsc{Evo-1-131k} was pretrained exclusively on prokaryotic
sequences, whereas $12$ of $13$ GENEB categories evaluate eukaryotic
genomic functions (Domain Mismatch paragraph,
Section~\ref{sec:results}); the size disadvantages reported above
therefore conflate scale with domain alignment. The systematic
analysis of pretraining data effects, restricted to controlled pairs
matched on architecture and tokenization, is reported in
Appendix~\ref{subsec:controlled_pairs}.

\subsection{Few-Shot Learning Dynamics}
\label{subsec:fewshot}

We evaluate model robustness under data-limited conditions through
systematic few-shot experiments. Across all $40$ models, mean
macro-MCC degrades from $0.488$ (full training data) to $0.253$
($10$-shot) to $0.106$ ($1$-shot), corresponding to a $78.2\%$
performance reduction in the extreme low-data regime
(Figure~\ref{fig:fewshot}).

Degradation patterns exhibit some heterogeneity across model capacity
tiers but no monotonic trend. Tiny models ($<$100M, $n=11$) decline
from $0.443$ to $0.109$ ($75.4\%$ degradation); small models
(100M--500M, $n=16$) show a steeper decline from $0.485$ to $0.093$
($80.9\%$); medium models (500M--2B, $n=6$) decline from $0.526$ to
$0.121$ ($77.1\%$); and large models ($\geq 2$B, $n=7$) decline from
$0.533$ to $0.121$ ($77.4\%$). The lack of a clear size-robustness
trend at the tier level is consistent with the model-level pattern
reported below.

Counterintuitively, the most few-shot-robust models in absolute terms
are those with the weakest full-data performance. The five smallest
absolute drops are observed for \textsc{Evo-1-131k}
($\Delta = 0.196$), \textsc{Caduceus-PH-1k} ($0.220$),
\textsc{JanusDNA-72-wo} ($0.272$), \textsc{Caduceus-PS-131k}
($0.272$), and \textsc{JanusDNA-72-w} ($0.275$), all of which rank
among the weakest models in full-shot evaluation. Conversely, the
strongest full-shot performers exhibit the largest drops:
\textsc{GENERator-Eukaryote-3B} ($\Delta = 0.489$),
\textsc{GENERator-Eukaryote-1.2B} ($0.463$), \textsc{LucaOne}
($0.461$), and \textsc{NT-2.5B-MS} ($0.456$), all exceeding $0.42$
in absolute drop.

This inverse relationship should not be read as evidence of greater
representational robustness in the lower-performing models: a small
absolute drop reflects a low full-shot ceiling that leaves limited
room for further degradation, not recovery of useful signal under
1-shot supervision (Section~\ref{sec:results}, Few-Shot Robustness
paragraph). The practical implication is that aggregate few-shot
rankings conflate task tractability with model quality and should be
interpreted alongside full-shot performance, as discussed in
Section~\ref{sec:results}.

\subsection{Pretraining Corpus Effects on Aggregate Performance}
\label{subsec:pretraining}

Grouping the 40 evaluated models by pretraining corpus type yields
an ordering by mean macro-MCC that is consistent with the controlled
pretraining-data comparisons reported in
Appendix~\ref{subsec:controlled_pairs}. Eukaryotic gene-focused
pretraining yields the highest aggregate performance
(\textsc{GENERator-Eukaryote-3B}, \textsc{GENERator-Eukaryote-1.2B};
$n=2$, mean macro-MCC $= 0.593$), followed by human-mouse epigenomic
profiles (\textsc{Enformer}, \textsc{SPACE}; $n=2$, $0.544$) and
broad multi-species corpora ($n=19$, $0.525$). Plant-specific
pretraining (\textsc{PlantCaduceus}, \textsc{Agro-NT-1B}; $n=2$,
$0.463$), human-only ($n=13$, $0.433$), and multi-species microbial
(\textsc{DNABERT-S}; $n=1$, $0.427$) corpora form an intermediate
band, with prokaryotic pretraining (\textsc{Evo-1-131k}; $n=1$,
$0.298$) yielding the lowest aggregate macro-MCC.

We emphasize that this ordering is descriptive and confounded by
architecture, tokenization, and parameter scale: only the controlled
pairs reported in Appendix~\ref{subsec:controlled_pairs} support
clean attribution to pretraining corpus. Within those controlled
comparisons (matched architecture and tokenization), the largest
effect observed in GENEB is the multi-species vs.\ microbial
contrast: \textsc{Genomics-FM} (multi-species) exceeds
\textsc{DNABERT-S} (multi-species-microbial) by $+0.088$ macro-MCC
under matched Transformer-encoder/$k$-mer conditions, and
\textsc{NT-v2-100M-MS} similarly exceeds \textsc{DNABERT-S} by
$+0.081$ macro-MCC. The eukaryotic-genes vs.\ broad multi-species
contrast, supported by a single matched pair
(\textsc{GENERator-Eukaryote-3B} vs.\ \textsc{DNA-GPT-3B-M}; both 3B,
Transformer-decoder, $k$-mer), shows a $+0.063$ macro-MCC advantage
for the eukaryotic-genes corpus, but rests on insufficient data to
support strong claims (Transfer Learning Analysis paragraph,
Section~\ref{sec:results}).

The ranking of aggregate means above is consistent with the
biological expectation that taxonomic alignment between pretraining
and downstream tasks supports transfer, since $12$ of $13$ GENEB
categories evaluate eukaryotic genomic functions. Models pretrained
on prokaryotic or microbial corpora are correspondingly disadvantaged
under aggregate evaluation, as discussed in the Domain Mismatch
paragraph (Section~\ref{sec:results}).

\subsection{Synthesis: Design Principles for DNA Foundation Models}
\label{subsec:synthesis}

Our systematic analysis yields several actionable principles for DNA
foundation model development. First, architectural choice matters but
interacts with other factors: Transformer-based models consistently
outperform the evaluated state-space alternative under controlled
conditions, while the encoder vs.\ decoder distinction is task- and
setting-dependent (Section~\ref{sec:results}, Architecture Comparison
paragraph). Second, tokenization effects are context-dependent and
cannot be optimized in isolation from architecture and pretraining
choices. Third, scale shows a substantial but non-deterministic
association with performance (Spearman $\rho = 0.573$, $p < 0.001$):
in-domain cross-scale reversals are common, and scale cannot
compensate for data-domain mismatch -- an 86M-parameter model trained
on human sequences (\textsc{MutBERT}) outperforms a 7B-parameter
model trained on prokaryotic data (\textsc{Evo-1-131k}) by $+0.231$
macro-MCC.

Most critically, pretraining corpus composition is a substantial
contributor to downstream performance. The descriptive ranking of
aggregate means -- eukaryotic genes $>$ human-mouse profiles $>$
multi-species $>$ plant-specific $>$ human-only $\approx$ microbial
$>$ prokaryotic -- reflects taxonomic alignment with evaluation
tasks, although the ordering is confounded by architecture,
tokenization, and parameter scale, and only the controlled pairs
reported in Appendix~\ref{subsec:controlled_pairs} support clean
attribution to corpus type. This pattern suggests that practitioners
should prioritize domain-appropriate pretraining alongside
architectural and scale considerations when computational resources
are constrained.

\subsection{Transfer Learning Analysis}
\label{subsec:transfer_learning}

The diversity of pretraining data sources in our benchmark enables
systematic investigation of cross-domain transfer dynamics. We
analyze how representations learned from different taxonomic
domains transfer to downstream tasks, revealing both positive and
negative transfer phenomena with substantial practical
implications. All values in this section are macro-averaged MCC
(per-category averaging across the 13 functional categories of
GENEB) unless otherwise stated; the full enumeration of matched
pairs underlying the controlled comparisons appears in
Appendix~\ref{subsec:controlled_pairs}.

\subsubsection{Methodological Framework}

To isolate pretraining-data effects from confounding architectural
and tokenization factors, we rely on controlled-pair comparisons.
The \textsc{GENA-LM} pair provides an ideal natural experiment:
\textsc{GENA-LM} (human-only) and \textsc{GENA-LM-T2T-Multi}
(multi-species) share identical Transformer-encoder architecture
and BPE tokenization, differing principally in pretraining corpus.
This controlled setup enables more direct attribution of performance
differences to data composition. The full 31-pair controlled-comparison
set is detailed in Appendix~\ref{subsec:controlled_pairs}.

\subsubsection{Controlled Comparison: Human versus Multi-Species Pretraining}

The \textsc{GENA-LM} comparison reveals systematic advantages for
multi-species pretraining across the task spectrum. Multi-species
training yields superior per-category mean MCC in $11$ of $13$
task categories. The largest advantages emerge for Chromatin
Accessibility ($\Delta = +0.123$ MCC: $0.583$ vs.\ $0.461$),
Mouse Enhancers ($+0.067$: $0.548$ vs.\ $0.480$), and Species
Classification ($+0.033$: $0.707$ vs.\ $0.674$). Human-only
pretraining shows marginal advantages only for Enhancers
($+0.010$ MCC in favor of human-only) and Regulatory tasks
($+0.001$), with differences within noise margins.

This pattern generalizes beyond the \textsc{GENA-LM} pair. Across
all six available human-vs.\ multi-species controlled comparisons
(Appendix~\ref{subsec:controlled_pairs},
Table~\ref{tab:pairs_pretraining}), multi-species pretraining is
favored for Chromatin Accessibility ($6/6$ pairs), lncRNA
($6/6$ pairs), Splice Sites ($5/6$ pairs), and Mouse Enhancers
($5/6$ pairs). Human-only pretraining shows a small advantage on
Virus/Phage ($3/6$ pairs, mean $\Delta = -0.025$ MCC), likely
reflecting the predominance of human-associated viral sequences
in human-genome training data.

\subsubsection{Negative Transfer: Human to Plant Domains}

The lncRNA task category provides a stringent test of cross-kingdom
transfer, comprising six plant-specific classification tasks
spanning \textit{Glycine max} (soybean),
\textit{Manihot esculenta} (cassava), \textit{Sorghum bicolor}
(sorghum), \textit{Solanum lycopersicum} (tomato),
\textit{Triticum aestivum} (wheat), and \textit{Zea mays} (maize).
Human-trained models exhibit substantial negative transfer on
these tasks, achieving mean lncRNA MCC of only $0.164$ compared
to $0.347$ for plant-trained models -- a deficit of $0.183$ MCC,
or roughly a $112\%$ relative gain from domain-appropriate
pretraining.

Task-level analysis reveals consistent patterns across all six
plant species. On \textit{G.\ max} lncRNA classification,
human-trained models average $0.128$ MCC versus $0.309$ for
plant-trained models. Similar gaps emerge for
\textit{S.\ bicolor} ($0.182$ vs.\ $0.408$),
\textit{M.\ esculenta} ($0.143$ vs.\ $0.408$), and the remaining
three species. Even the best-performing human-trained model
(\textsc{MutBERT}, lncRNA MCC $= 0.260$) substantially
underperforms plant-specialized models (\textsc{PlantCaduceus},
$0.357$; \textsc{Agro-NT-1B}, $0.336$).

Notably, multi-species models achieve intermediate performance
(mean lncRNA MCC $= 0.307$), with \textsc{LucaOne} reaching
$0.508$ -- the best overall result on plant lncRNA tasks. This
suggests that broad taxonomic coverage partially compensates for
the lack of plant-specific pretraining, though dedicated plant
models retain advantages on most individual tasks.

\subsubsection{Positive Transfer: Multi-Species to Human-Specific Tasks}

A counterintuitive finding emerges from analysis of predominantly
human-derived tasks: multi-species pretraining consistently
outperforms human-only pretraining even on task categories whose
underlying datasets come primarily from human genomic data. This
positive transfer phenomenon manifests across all five
human-centric task categories examined here.

For Histone Modifications, multi-species models achieve mean MCC
of $0.476$ compared to $0.400$ for human-trained models
($\Delta = +0.076$). Chromatin Accessibility shows the largest
gap at $+0.096$ ($0.605$ vs.\ $0.509$). Regulatory element
prediction exhibits $+0.056$ advantage ($0.424$ vs.\ $0.369$),
Enhancer detection $+0.051$ ($0.464$ vs.\ $0.413$), and TF
Binding $+0.028$ ($0.570$ vs.\ $0.542$). The consistency of
multi-species advantages across functionally diverse task
categories is consistent with exposure to evolutionarily conserved
sequence patterns during pretraining supporting representation
quality even for species-specific downstream applications.

\subsubsection{Catastrophic Transfer Failure: Prokaryotic to Eukaryotic Domains}

The most extreme transfer failure involves \textsc{Evo-1-131k}, a
7B-parameter model pretrained exclusively on prokaryotic
sequences. Despite its substantial scale, \textsc{Evo-1-131k}
achieves only $0.298$ overall macro-MCC, ranking last ($40$ of
$40$ models) and underperforming \textsc{MutBERT} (86M parameters,
human pretraining) by $0.231$ macro-MCC -- an $81$-fold parameter
disadvantage yielding inferior results.

Category-level analysis reveals the biological basis for this
failure. \textsc{Evo-1-131k} ranks last ($40/40$) on Splice Sites
(MCC $= 0.160$), TF Binding ($0.173$), Species Classification
($0.285$), and DNA Methylation ($0.073$). These failures reflect
fundamental differences between prokaryotic and eukaryotic genomic
organization: prokaryotes lack spliceosomal introns, employ
distinct transcription factor families, and utilize different DNA
methylation machinery. The only category where \textsc{Evo-1-131k}
achieves competitive performance is Coding/Non-coding Classification
(MCC $= 0.719$, rank $36/40$), reflecting the more universal
nature of coding sequence signatures across domains of life.

We emphasize three scope qualifiers for this observation.
\emph{First}, the result reflects performance under frozen
linear-probing of pretrained representations -- the evaluation
protocol applied uniformly across all $40$ benchmark models. It does not preclude
the possibility that task-specific full fine-tuning of
\textsc{Evo-1-131k} on eukaryotic tasks would close part of the
gap, and is not a statement about \textsc{Evo-1-131k}'s
capabilities within its intended prokaryotic application domain,
on which it was not evaluated here. \emph{Second},
\textsc{Evo-1-131k} is the only prokaryotic-only model in GENEB;
broader claims about prokaryotic-to-eukaryotic transfer would
require additional prokaryotic-pretrained models, which we flag as
a coverage limitation. \emph{Third}, the magnitude of the failure
on biologically structured tasks (splicing, DNA methylation) is
consistent with the prior expectation that representations of
prokaryotic sequence statistics carry limited information about
eukaryotic-specific molecular machinery, and we frame this
observation as evidence of domain mismatch rather than a universal
critique of the underlying model architecture or training
methodology.

\subsubsection{Partial Transfer: Microbial to Eukaryotic Domains}

\textsc{DNABERT-S}, pretrained on multi-species microbial genomes,
provides an intermediate case between prokaryotic-only and
eukaryotic multi-species pretraining. With overall macro-MCC of
$0.427$, \textsc{DNABERT-S} substantially outperforms
prokaryotic-only \textsc{Evo-1-131k} ($0.298$) but underperforms eukaryotic multi-species models (mean $0.525$).

Controlled comparison with \textsc{NT-v2-100M-MS} (eukaryotic
multi-species, matched architecture and tokenization) reveals
systematic deficits for the microbial corpus:
\textsc{NT-v2-100M-MS} exceeds \textsc{DNABERT-S} by $+0.081$
overall macro-MCC, with the largest per-category gaps on Splice
Sites ($+0.183$) and Species Classification ($+0.187$). The splice
site deficit again reflects the absence of spliceosomal machinery
in microbial training sequences. Interestingly, \textsc{DNABERT-S}
shows unusual strength on Regulatory tasks ($+0.069$ vs.\
\textsc{Genomics-FM}), potentially reflecting transferable sequence
features in regulatory regions across bacterial and eukaryotic
genomes. This intermediate-corpus observation rests on a single
microbial-pretrained model in GENEB
(Appendix~\ref{app:excluded_models}); broader claims about
microbial-to-eukaryotic transfer await additional matched models.

\subsubsection{Strong Positive Transfer: Eukaryotic Gene-Focused Pretraining}
The \textsc{GENERator} models, pretrained on curated eukaryotic
gene sequences, achieve the strongest overall transfer performance
in our benchmark (mean macro-MCC $= 0.593$). Remarkably, these
models exceed benchmark-wide averages across all $13$ task
categories, with largest advantages on DNA Methylation ($+0.200$
vs.\ benchmark mean), Splice Sites ($+0.198$), lncRNA ($+0.187$),
Species Classification ($+0.137$), Mouse Enhancers ($+0.136$),
and Chromatin Accessibility ($+0.117$).

Controlled comparison between \textsc{GENERator-Eukaryote-3B} and
\textsc{DNA-GPT-3B-M} (both 3B parameters, Transformer-decoder,
$k$-mer tokenization) isolates the effect of gene-focused versus
general multi-species pretraining. The gene-focused approach
yields $+0.063$ overall macro-MCC advantage, with particularly
strong gains on Chromatin Accessibility ($+0.191$), lncRNA
($+0.142$), and Mouse Enhancers ($+0.124$). This pattern is
consistent with curation of pretraining data toward functionally
annotated genomic regions supporting downstream task performance
beyond what raw sequence diversity provides. We emphasize that
this contrast rests on a single matched pair and should be
interpreted with corresponding caution; see the Transfer Learning
Analysis paragraph (Section~\ref{sec:results}) and the
single-pair caveat in Appendix~\ref{subsec:controlled_pairs}.

\subsubsection{Specialized Transfer: Human-Mouse Epigenomic Profiles}

\textsc{Enformer} and \textsc{SPACE}, pretrained on human and mouse
epigenomic profiles, exhibit a distinctive transfer pattern
characterized by strong performance on regulatory tasks but
deficits on sequence-intrinsic features. These models achieve mean
macro-MCC of $0.544$ overall, with substantial advantages on
Regulatory tasks ($+0.194$ vs.\ benchmark mean), Mouse Enhancers
($+0.170$), Chromatin Accessibility ($+0.133$), and TF Binding
($+0.116$).

However, this specialization comes at a cost: human-mouse-profile
models underperform on DNA Methylation ($-0.104$), Coding/Non-coding
Classification ($-0.061$), and lncRNA ($-0.021$). This trade-off
reflects the nature of epigenomic profile pretraining, which
emphasizes chromatin state and regulatory-element patterns while
potentially underweighting primary sequence features. Practitioners
should consider this specialization when selecting models for
specific application domains (see also Practitioner Recommendations,
Section~\ref{sec:results}).

\subsubsection{Transfer Dynamics on Mouse Enhancer Tasks}

The five Mouse Enhancer tasks provide a detailed view of how
pretraining data influences performance on a homogeneous task
category. Human-mouse-profile models (\textsc{Enformer},
\textsc{SPACE}) achieve top-1 performance on tasks $0$
(\textsc{Enformer} MCC $= 0.667$), $2$ ($0.878$), and $3$ ($0.772$),
with substantial margins over human-only models on these tasks.
The remaining two tasks (1 and 4) are won by \textsc{Omni-DNA-1B}
(Transformer-decoder, multi-species, BPE; MCC $= 0.807$ and
$0.488$ respectively). Eukaryotic-gene \textsc{GENERator} models
show consistent strong performance across all five tasks (mean
$0.612$) without claiming any individual top-1 position. Aggregating
by pretraining-corpus group, mean macro-MCC on the Mouse Enhancer
category is $0.646$ for human-mouse-profile models, $0.612$ for
\textsc{GENERator}, $0.518$ for broad multi-species, $0.399$ for
plant-trained, and $0.399$ for human-only models.

The superior performance of human-mouse-profile models on mouse
enhancer tasks -- despite these models not being trained on
general genomic sequences -- demonstrates that task-relevant
pretraining signals can outweigh broader sequence coverage when
the downstream task aligns tightly with the pretraining target.
This pattern is consistent with the practitioner guidance that
model selection should be informed by task--domain alignment
rather than aggregate benchmark performance alone (see
Practitioner Recommendations, Section~\ref{sec:results}).

\subsubsection{Synthesis: Transfer Learning Principles}

Our systematic analysis of cross-domain transfer reveals five
recurring patterns governing DNA foundation model generalization.

First, taxonomic alignment between pretraining and downstream
domains is critical for successful transfer. Prokaryotic
pretraining fails on eukaryotic tasks, while human pretraining
shows negative transfer to plant-specific tasks. The magnitude of
these failures -- 7B parameters underperforming 86M parameters by
$0.231$ macro-MCC -- underscores that domain mismatch cannot be
overcome through scale alone.

Second, taxonomic diversity in pretraining provides positive
transfer even to ostensibly species-specific tasks. Multi-species
models outperform human-only models on all five human-specific
task categories examined here, consistent with broad sequence
exposure during pretraining supporting transferable
representations. This pattern argues against narrow
species-specific pretraining for general-purpose foundation
models.

Third, specialized pretraining creates performance trade-offs
rather than uniform improvements. Human-mouse epigenomic profile
models excel on regulatory tasks ($+0.194$ vs.\ benchmark mean)
but underperform on DNA methylation ($-0.104$). Gene-focused
eukaryotic pretraining achieves the highest overall performance
but relies on curated training data. These trade-offs suggest that
no single pretraining strategy optimally serves all downstream
applications.

Fourth, model scale cannot compensate for fundamental data-domain
mismatch. The $81$-fold parameter advantage of \textsc{Evo-1-131k}
over \textsc{MutBERT} is entirely negated by the
prokaryotic--eukaryotic domain gap. This finding has practical
resource-allocation implications: investment in domain-appropriate
training data may yield higher returns than equivalent investment
in model scale when domain mismatch exists.

Fifth, the observed transfer patterns suggest that model selection
should be informed by task--domain alignment rather than aggregate
benchmark performance alone. Different pretraining strategies
excel on different task categories, motivating per-category
selection guidance: regulatory tasks favor epigenomic-profile
models (\textsc{Enformer}, \textsc{SPACE}); general eukaryotic
tasks favor multi-species or gene-focused models; and for
plant-specific tasks, dedicated plant-trained models substantially
exceed human-trained baselines (mean lncRNA MCC $0.347$ vs.\
$0.164$), while the best individual results come from select
multi-species models such as \textsc{LucaOne} (lncRNA MCC $=
0.508$). The per-category Practitioner Recommendations in
Section~\ref{sec:results} provide operational guidance consistent
with these alignment principles.

\subsection{Specialization Score: Formal Definition and Computation}
\label{subsec:specialization}

For each model $m$ and each task $t$ we compute the per-task rank
$r_{m,t} \in \{1, \ldots, 40\}$ from full-shot MCC, with rank $1$
denoting the best-performing model on task $t$. For a category
$C$ containing tasks $\mathcal{T}_C$, the within-category mean rank
of model $m$ is
\[
\bar{r}_{m,C} \;=\; \frac{1}{|\mathcal{T}_C|}\sum_{t\in\mathcal{T}_C} r_{m,t},
\]
and the corresponding outside-category mean rank is
\[
\bar{r}_{m,\neg C} \;=\; \frac{1}{|\mathcal{T}\setminus\mathcal{T}_C|}\sum_{t\notin\mathcal{T}_C} r_{m,t},
\]
where $\mathcal{T}$ is the full set of $100$ GENEB tasks. The
specialization score of model $m$ on category $C$ is then
\[
\Delta_{m,C} \;=\; \bar{r}_{m,\neg C} \;-\; \bar{r}_{m,C}.
\]
A positive $\Delta_{m,C}$ indicates that model $m$ ranks better on
category $C$ than on the rest of the benchmark; values above $\sim 5$
reflect substantial relative strength. The largest specialization
scores observed in GENEB are
\textsc{BioFM-265M} on DNA methylation ($\Delta = 26.3$),
\textsc{PlantCaduceus} on lncRNA ($\Delta = 19.0$),
\textsc{JanusDNA-72-w} on chromatin accessibility ($\Delta = 17.9$),
and \textsc{eccDNAMamba} on chromatin accessibility ($\Delta = 16.5$).
We use the score as a diagnostic measure throughout
Appendix~\ref{app:results}, particularly when discussing model families
whose category-level strengths differ from their aggregate ranking.

\newpage

\subsection{Task-Category-Specific Analysis}
\label{subsec:task_categories}

Having established general principles governing DNA foundation model performance, we now examine category-specific patterns across the functional task categories of GENEB. This fine-grained analysis reveals substantial heterogeneity in optimal model configurations across functional genomic prediction problems.

\subsubsection{Histone Modifications}

\begin{figure}[h]
    \centering
    \includegraphics[width=1.0\linewidth]{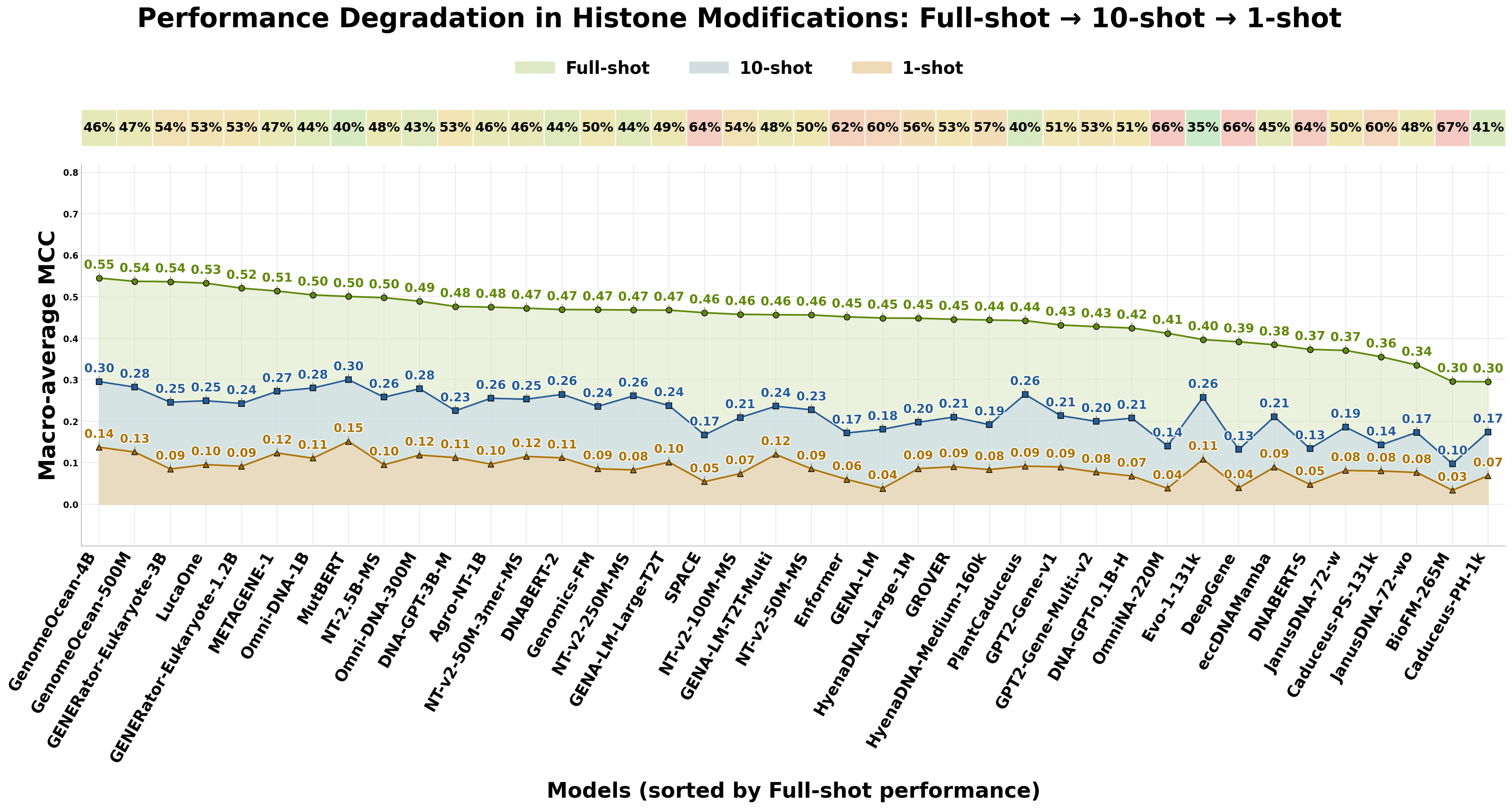}
\caption{\textbf{Few-shot performance degradation on Histone Modifications.}
For each of the 40 models, macro-average MCC across 30 histone modification 
tasks under full-shot, 10-shot, and 1-shot regimes; models ordered by 
full-shot performance. The top band shows the relative drop from full-shot 
to 10-shot per model. Benchmark-wide mean degradation: $51.0\%$ for 10-shot, 
$80.3\%$ for 1-shot. The 1-shot regime collapses to near-random performance, 
while 10-shot retains discriminative signal; full-shot and 10-shot rankings 
differ substantially (Spearman $\rho = 0.77$).}
\label{fig:kshot_histone}
\end{figure}

\begin{figure}[h]
    \centering
    \includegraphics[height=0.3\textheight]{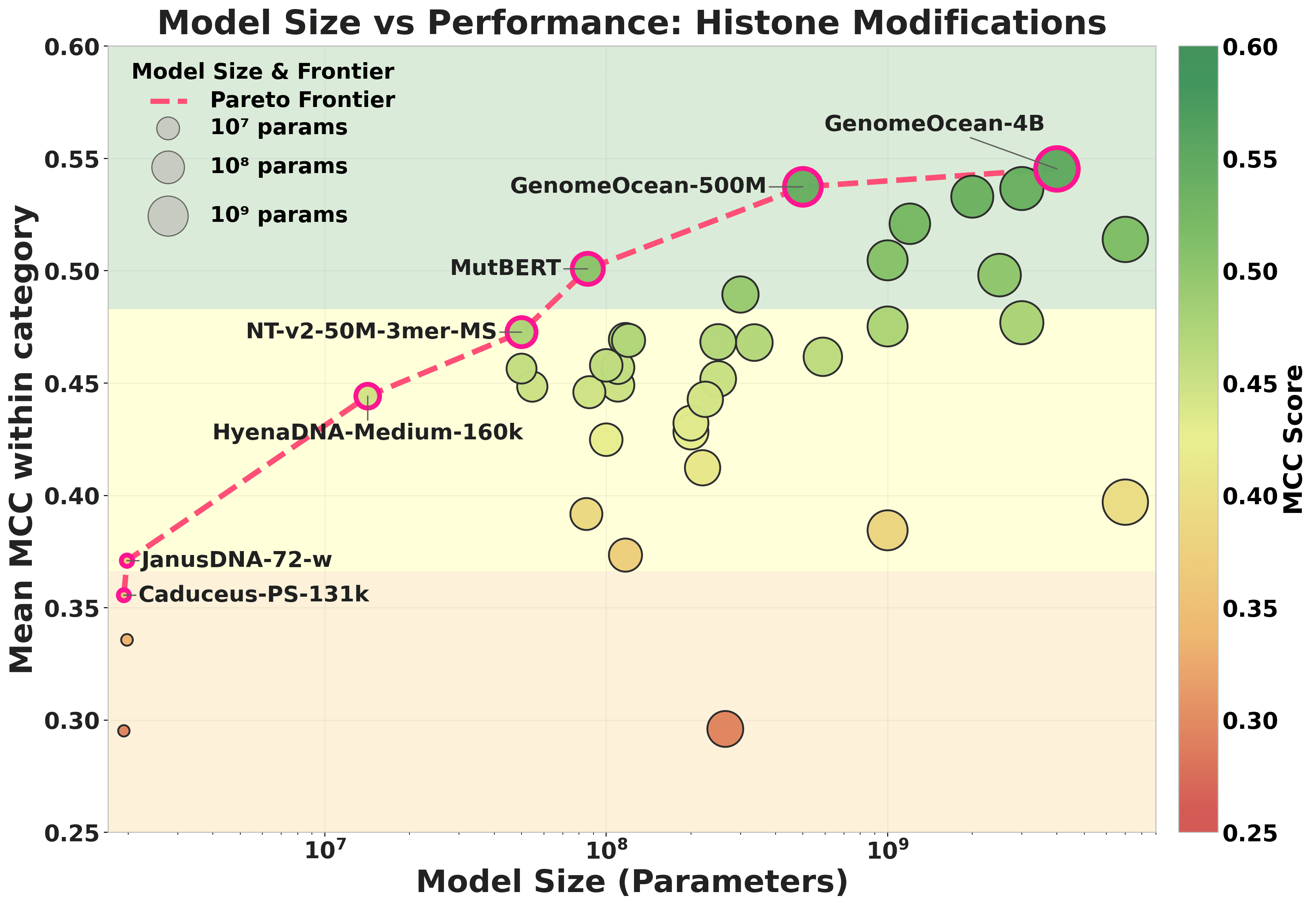}
    \caption{\textbf{Pareto frontier for Histone Modifications:
    mean MCC vs.\ parameter count.}
    Each point represents one of the 40 genomic foundation models,
    with parameter count on a logarithmic x-axis and mean
    full-shot Histone Modifications MCC on the y-axis. Marker size
    and color both encode MCC. The dashed line marks the Pareto
    frontier of best performance--size trade-offs.
    Scale shows a positive but non-deterministic association with
    Histone Modifications performance: several smaller models sit
    on or near the frontier, including \textsc{MutBERT} (86M,
    mean MCC $= 0.501$), which is the strongest sub-100M model on
    this category. Other small models such as
    \textsc{JanusDNA-72-w} and \textsc{JanusDNA-72-wo} (both
    2M parameters) sit well below the frontier, illustrating
    that small parameter count is neither sufficient nor
    consistent for competitive performance on this category.}
    \label{fig:pareto_histone}
\end{figure}

\begin{figure}[H]
    \centering
    \includegraphics[width=1.0\linewidth]{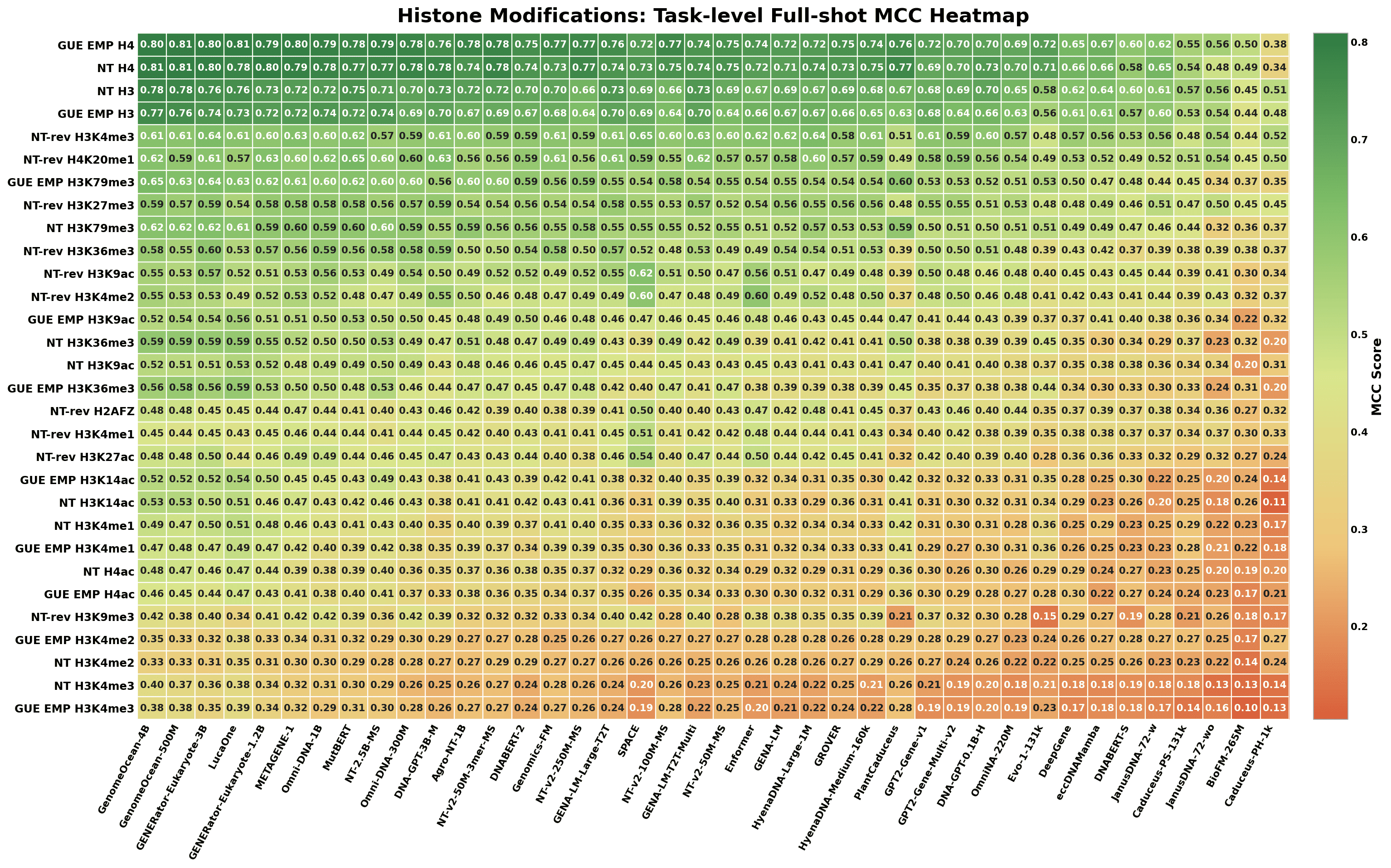}
    \caption{\textbf{Per-task MCC for Histone Modifications.}
    Heatmap shows full-shot MCC for each of the 40 genomic
    foundation models on the 30 histone modification tasks, with
    models sorted by mean Histone Modifications MCC. Cell values
    report per-task MCC, with colors ranging from red/orange for
    lower scores to green for higher scores. Task difficulty varies
    substantially across marks: H4-family tasks (H4, H4ac,
    H4K20me1) are consistently among the easiest, while
    H3K4me2/me3 tasks remain among the hardest across most models.
    Tasks from the NT, NT-revised, and GUE EMP sources show
    broadly similar within-mark patterns, supporting the
    consolidation of these sources into a single category.}
    \label{fig:heatmap_histone}
\end{figure}

The Histone Modifications category ($n=30$ tasks) spans
diverse chromatin marks including H3K4me1, H3K4me2, H3K4me3,
H3K27ac, H3K36me3, and H4 modifications. Task difficulty varies
substantially: the easiest individual task is GUE EMP H4
prediction (mean MCC $= 0.720$ across the $40$ models), while the
most challenging is GUE EMP H3K4me3 prediction (mean MCC
$= 0.243$). Results for the 30 histone modification prediction tasks are
presented in Figures~\ref{fig:kshot_histone}--\ref{fig:heatmap_histone}.

Architecture comparisons under controlled conditions reveal
substantial advantages for Transformer-based models.
\textsc{GenomeOcean-500M} (Transformer-decoder, 500M) exceeds
\textsc{eccDNAMamba} (Mamba, 537M) by $0.153$ MCC
($0.537$ vs.\ $0.384$) at near-matched scale (1.07$\times$
ratio), with both models trained on multi-species data using
BPE tokenization. A within-Transformer contrast
(decoder vs.\ encoder) between \textsc{GenomeOcean-500M} (500M,
decoder, BPE, multi-species) and \textsc{GENA-LM-Large-T2T} (336M,
encoder, BPE, multi-species) shows a $0.069$ MCC advantage for the
decoder model, although the two models differ in parameter count
and the contrast is therefore not strictly matched on scale.

Pretraining data effects follow the general hierarchy established
in Section~\ref{subsec:pretraining}. Eukaryotic gene-focused
\textsc{GENERator-Eukaryote-3B} exceeds multi-species
\textsc{DNA-GPT-3B-M} by $0.060$ MCC ($0.537$ vs.\ $0.477$) in
controlled comparison (both 3B, Transformer-decoder, $k$-mer
tokenization). Multi-species models consistently outperform
microbial-trained alternatives, with \textsc{Genomics-FM}
exceeding \textsc{DNABERT-S} by $0.096$ MCC ($0.469$ vs.\ $0.373$)
under matched Transformer-encoder/$k$-mer conditions. Among
small-scale ($<$100M) models, \textsc{MutBERT} (86M) achieves the
highest Histone Modifications MCC ($0.501$), ranking $8$th of $40$
overall on this category.

Few-shot performance on histone modification tasks degrades
severely. The mean macro-MCC across models drops from $0.447$
(full data) to $0.219$ ($10$-shot) to $0.088$ ($1$-shot),
corresponding to $51.0\%$ and $80.3\%$ relative reductions
respectively. We focus on the $10$-shot regime here as the
$1$-shot regime collapses to near-random performance for all
models (no model exceeds $0.16$ macro-MCC at $1$-shot), limiting
its discriminative value.

At $10$-shot, the ranking of models differs substantially from
full-shot performance (Spearman $\rho = 0.77$ across all 40
models). The top $10$-shot performer is \textsc{MutBERT} (86M
parameters; $10$-shot MCC $= 0.300$, full-shot $= 0.501$), which
ranks $8$th at full-shot, illustrating that compact,
efficiency-oriented models can retain operationally useful
transferability under data-limited conditions where larger
gene-focused models do not (e.g., \textsc{GENERator-Eukaryote-3B}
drops to $10$-shot MCC $= 0.246$ from a full-shot MCC of $0.537$).
Of the top-5 models by full-shot MCC, only $2$ remain in the
top-5 at $10$-shot (\textsc{GenomeOcean-4B},
\textsc{GenomeOcean-500M}). As discussed in
Section~\ref{subsec:fewshot}, this reranking reflects task-
and setting-dependent few-shot dynamics rather than a single
property of model robustness.

\subsubsection{Promoter Recognition}

\begin{figure}[h]
    \centering
    \includegraphics[width=1.0\linewidth]{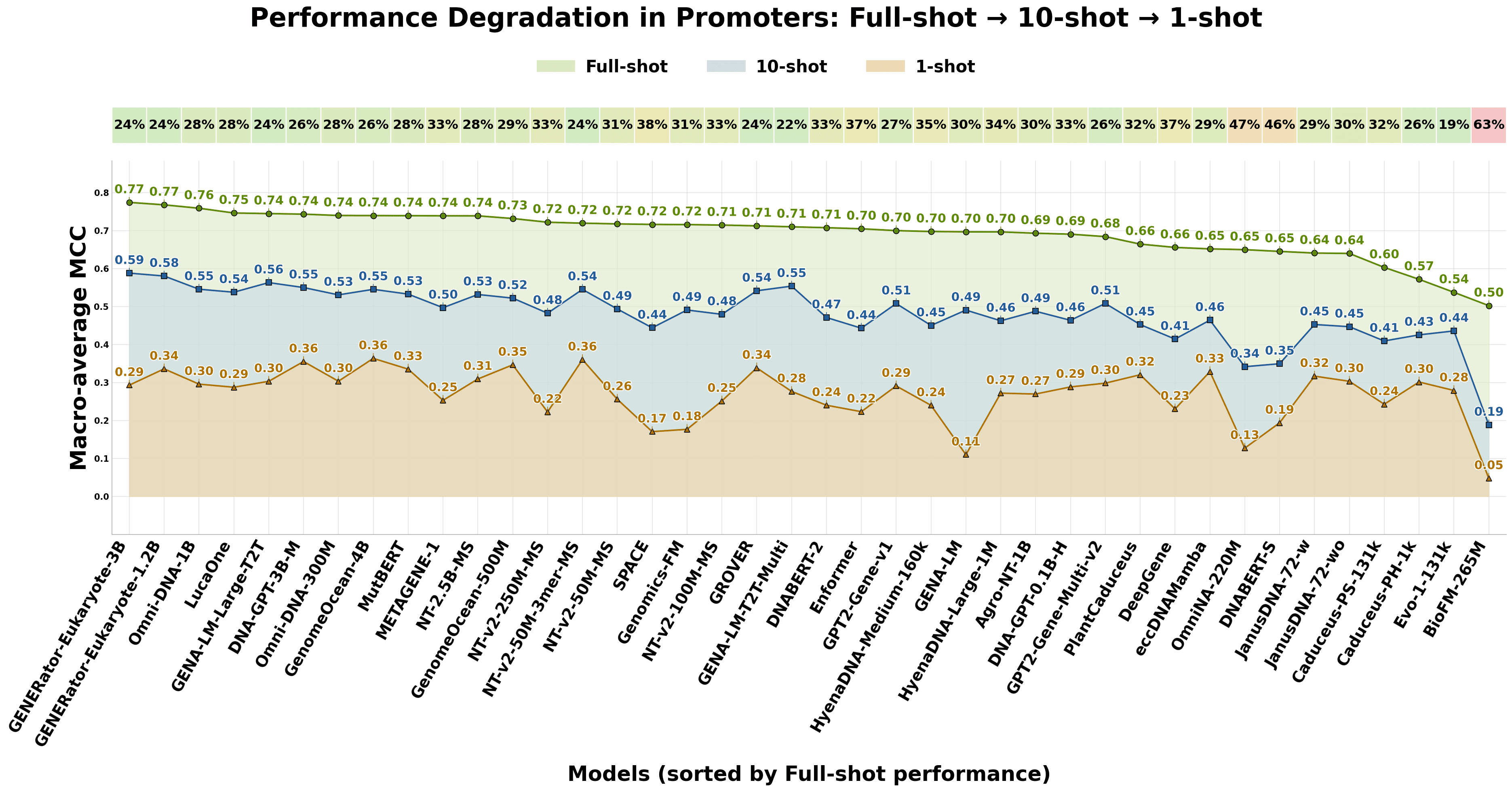}
    \caption{\textbf{Few-shot performance degradation on Promoter
    Recognition.} For each of the 40 models, macro-average MCC
    across 22 promoter prediction tasks under full-shot, 10-shot,
    and 1-shot regimes; models ordered by full-shot performance.
    The top band shows the relative drop from full-shot to 10-shot
    per model. Benchmark-wide mean degradation: $30.7\%$ for
    10-shot, $61.2\%$ for 1-shot. Unlike histone modifications,
    the 1-shot regime retains operationally meaningful signal on
    this category (maximum 1-shot MCC $= 0.363$); full-shot and
    10-shot rankings remain strongly correlated (Spearman
    $\rho = 0.85$).}
    \label{fig:kshot_promoters}
\end{figure}

\begin{figure}[H]
    \centering
    \includegraphics[height=0.3\textheight]{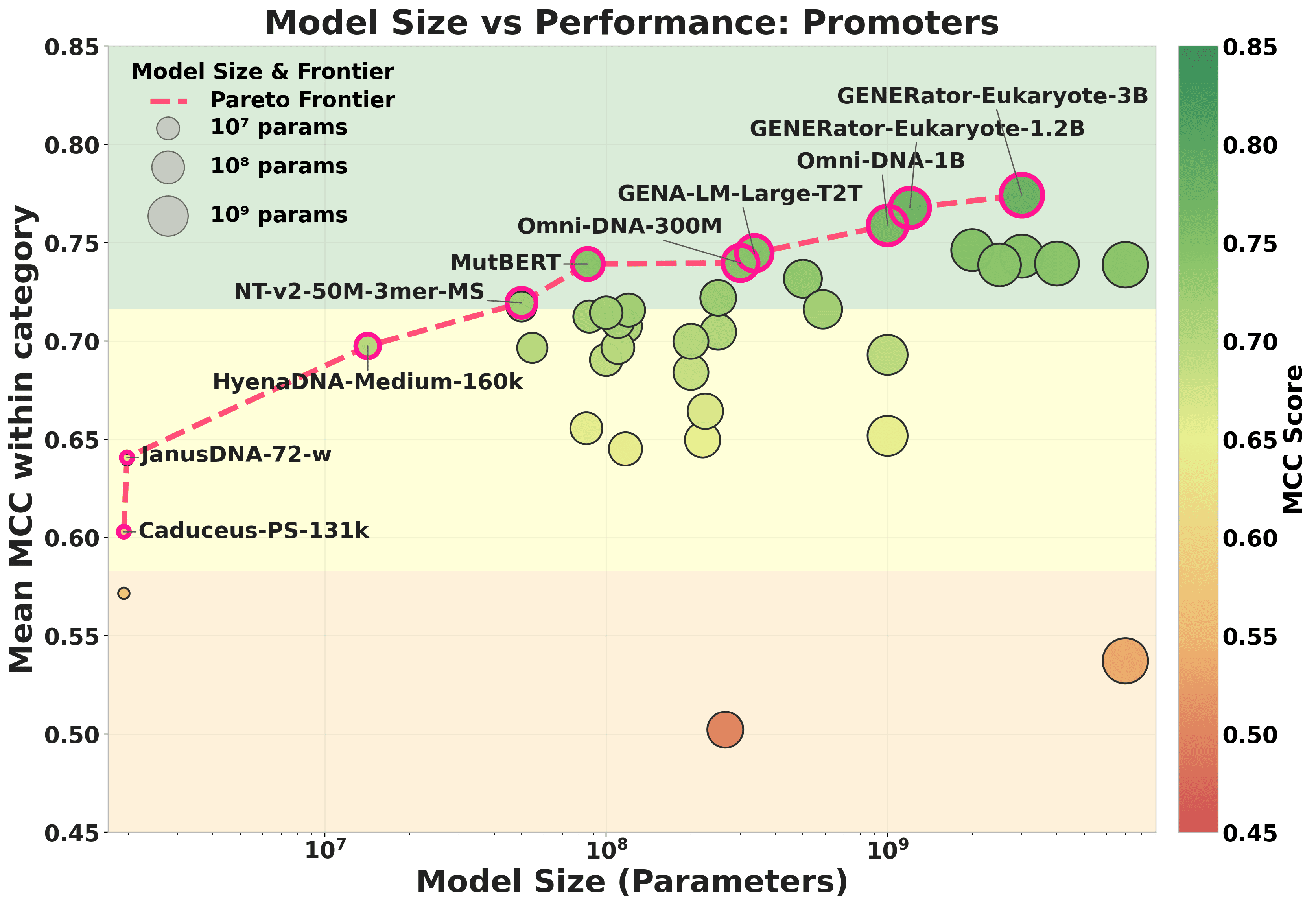}
    \caption{\textbf{Pareto frontier for Promoter Recognition:
    mean MCC vs.\ parameter count.}
    Each point represents one of the 40 genomic foundation models,
    with parameter count on a logarithmic x-axis and mean
    full-shot Promoter MCC on the y-axis. Marker size and color
    both encode MCC. The dashed line marks the Pareto frontier of
    best performance--size trade-offs. Scaling is comparatively
    weak on this category (Spearman $\rho = 0.493$; tier gap $\geq 1$B vs.\ $<$200M of only $0.042$ MCC), consistent with
    promoter sequence features being accessible to smaller models.
    \textsc{MutBERT} (86M, mean MCC $= 0.739$, ranked $9$th overall
    on this category) is the strongest sub-100M model, sitting
    near the Pareto frontier alongside several
    multi-billion-parameter models.}
    \label{fig:pareto_promoters}
\end{figure}

\begin{figure}[H]
    \centering
    \includegraphics[width=1.0\linewidth]{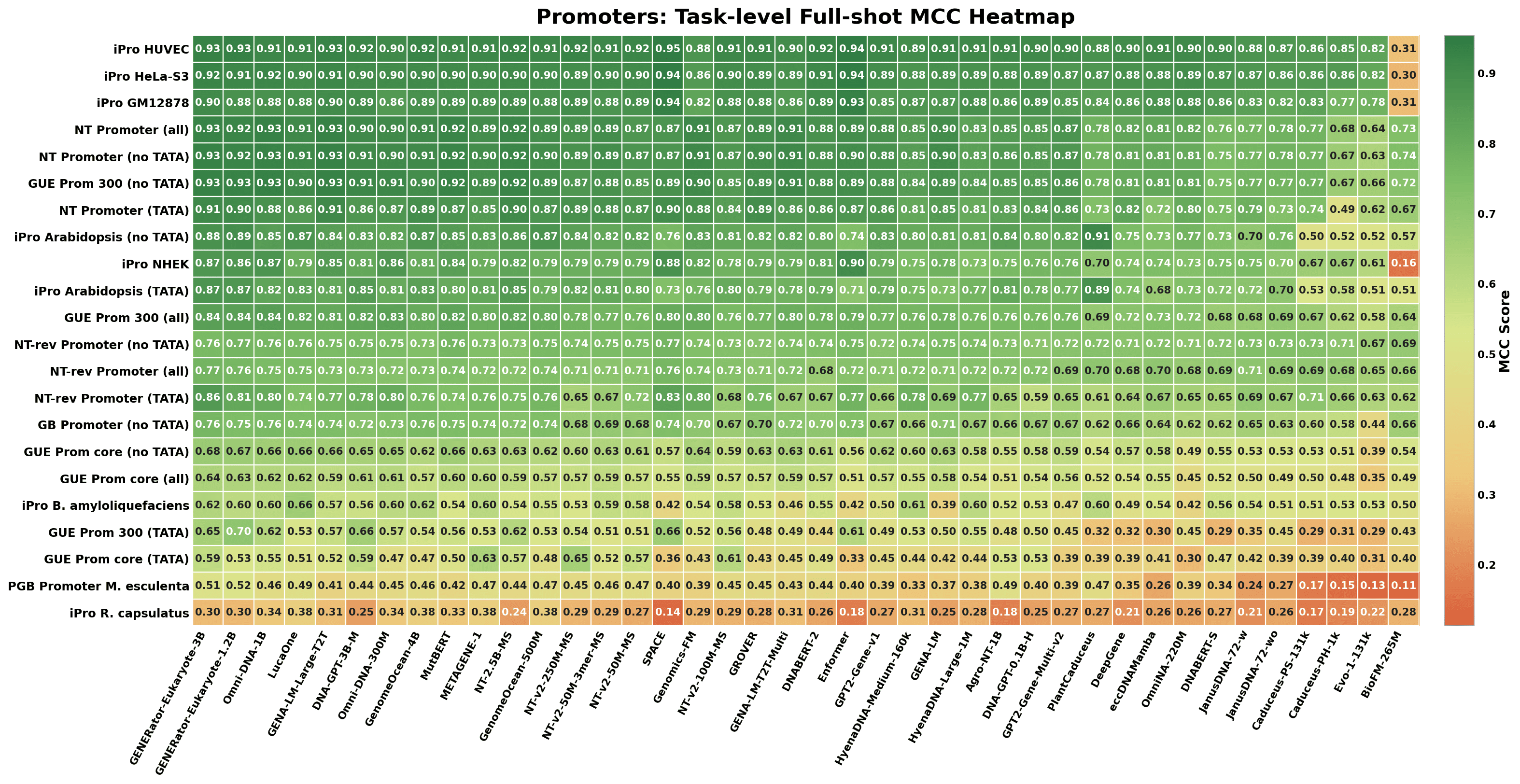}
    \caption{\textbf{Per-task MCC for Promoter Recognition.}
    Heatmap shows full-shot MCC for each of the 40 genomic
    foundation models on the 22 promoter prediction tasks, with
    models sorted by mean Promoter MCC. Cell values report
    per-task MCC, with colors ranging from red/orange for lower
    scores to green for higher scores. Task difficulty varies
    substantially with sequence source and host species:
    cell-type-specific human promoter prediction (HUVEC, HeLa-S3,
    GM12878) is consistently the easiest, while bacterial promoter
    prediction (\textit{R.\ capsulatus},
    \textit{B.\ amyloliquefaciens}) and plant promoter prediction
    (\textit{M.\ esculenta}) are the most challenging.}
    \label{fig:heatmap_promoters}
\end{figure}

Promoter recognition tasks ($n=22$ tasks) span bacterial, plant, and
mammalian systems with substantial difficulty variation.
Cell-type-specific human promoter tasks dominate the easy end
(\textsc{iPro-WAEL} HUVEC: mean MCC $= 0.890$; HeLa-S3: $0.875$;
GM12878: $0.855$), while bacterial promoter prediction for
\textit{R.\ capsulatus} is the most challenging task ($0.274$). Results for the 22 promoter prediction tasks are presented in
Figures~\ref{fig:kshot_promoters}--\ref{fig:heatmap_promoters}.

The scaling relationship is weaker for promoter tasks than for the
benchmark overall. The Spearman correlation between
$\log_{10}(\text{parameter count})$ and mean Promoter MCC is
$\rho = 0.493$ ($p = 0.001$), compared to $\rho = 0.573$ across
all categories (Section~\ref{sec:results}). Models with at least
1B parameters achieve mean MCC of $0.724$ versus $0.682$ for
models below 200M -- a gap of $0.042$ MCC, narrower than the
$+0.075$ tier gap observed at the benchmark-aggregate level. This
compressed performance range suggests that promoter sequence
features are relatively accessible to smaller models.

Architecture comparisons maintain Transformer advantages over the
evaluated state-space alternative: \textsc{GenomeOcean-500M}
(Transformer-decoder, 500M) outperforms \textsc{eccDNAMamba}
(Mamba, 537M) by $0.080$ MCC ($0.732$ vs.\ $0.652$) at
near-matched scale, with both models trained on multi-species
data using BPE tokenization.
Within Transformer-family comparisons,
\textsc{GENA-LM-Large-T2T} (encoder, BPE, multi-species, 336M)
exceeds \textsc{OmniNA-220M} (decoder, BPE, multi-species, 220M)
by $0.095$ MCC ($0.745$ vs.\ $0.650$), although the two models
differ in parameter count and the contrast is therefore not
strictly matched on scale. We note that this single-pair encoder
advantage on Promoters contrasts with the within-Transformer
decoder advantage observed on Histone Modifications, supporting
the broader observation (Section~\ref{sec:results}, Architecture
Comparison paragraph) that the encoder--decoder distinction is
task- and setting-dependent.

Pretraining data comparisons reveal nuanced patterns for promoter
tasks. \textsc{GENA-LM-T2T-Multi} (multi-species) exceeds
\textsc{GENA-LM} (human-only) by only $0.013$ MCC ($0.710$
vs.\ $0.697$) under matched Transformer-encoder/BPE conditions, a
narrower advantage than the average multi-vs.-human pretraining
contrast reported in
Appendix~\ref{subsec:controlled_pairs}.

Few-shot performance on promoter tasks degrades more gracefully
than on histone modifications. Across the 40 models, mean MCC
drops from $0.693$ (full data) to $0.480$ (10-shot) to $0.269$
(1-shot), corresponding to $30.7\%$ and $61.2\%$ relative
reductions respectively. Unlike histone modification, several
models retain operationally meaningful 1-shot performance (maximum
1-shot MCC = $0.363$ for \textsc{GenomeOcean-4B}). At 10-shot, the
top performers remain \textsc{GENERator-Eukaryote-3B} ($0.588$),
\textsc{GENERator-Eukaryote-1.2B} ($0.580$), and
\textsc{GENA-LM-Large-T2T} ($0.563$); the full-shot top-5 and
10-shot top-5 overlap by 3 models, and full-shot vs.\ 10-shot
rankings correlate at Spearman $\rho = 0.85$.

\subsubsection{Enhancer Prediction}

\begin{figure}[H]
    \centering
    \includegraphics[width=1.0\linewidth]{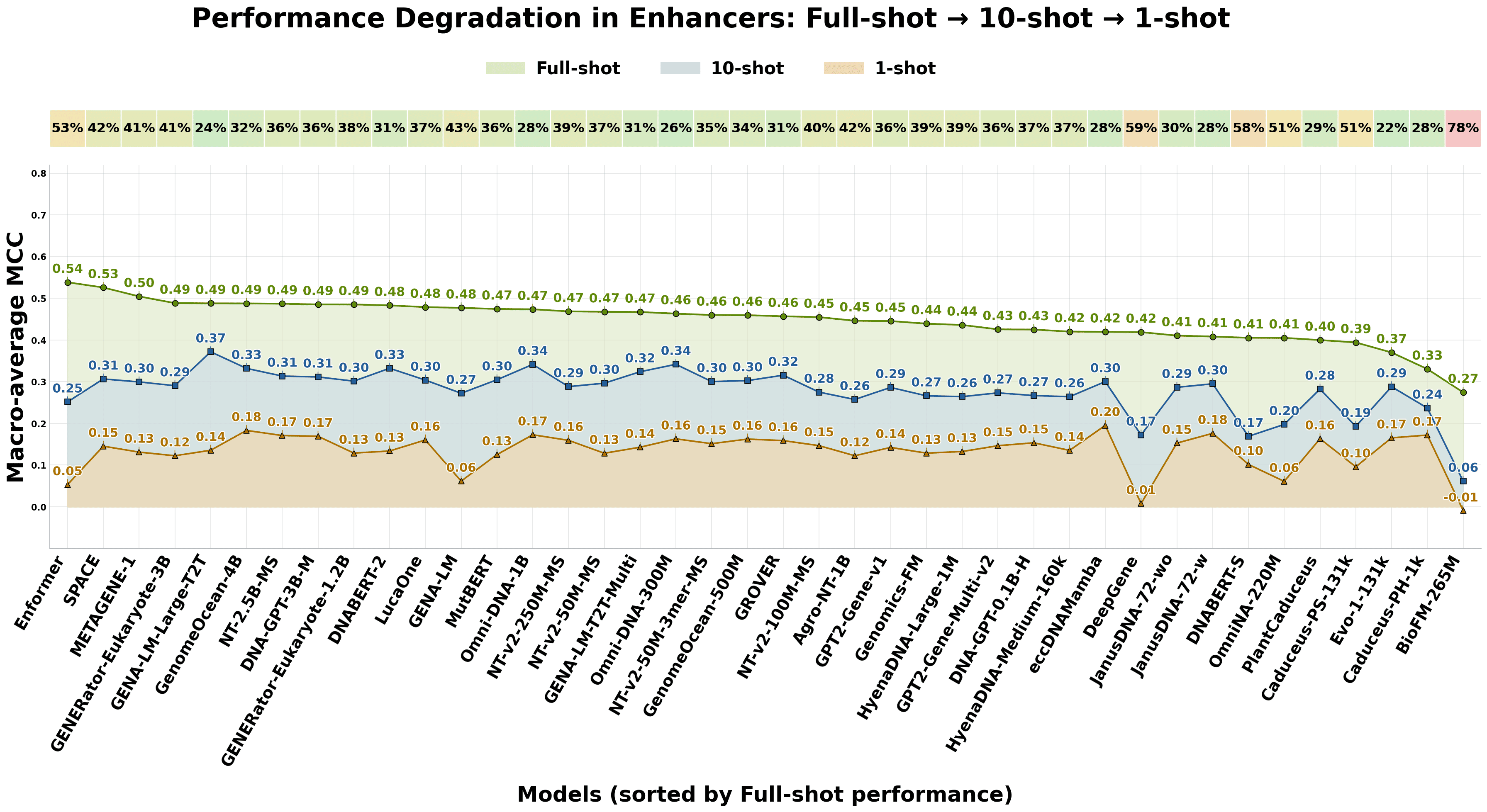}
    \caption{\textbf{Few-shot performance degradation on Enhancer
    Prediction.} For each of the 40 models, macro-average MCC
    across 8 enhancer prediction tasks under full-shot, 10-shot,
    and 1-shot regimes; models ordered by full-shot performance.
    The top band shows the relative drop from full-shot to
    10-shot per model. Benchmark-wide mean degradation: $37.6\%$
    for 10-shot, $70.0\%$ for 1-shot. The 1-shot regime collapses
    to near-random performance (maximum 1-shot MCC $= 0.195$);
    the 10-shot ranking diverges sharply from the full-shot
    ranking (Spearman $\rho = 0.64$), with epigenomic-profile
    specialists losing their full-shot advantage
    (\textsc{Enformer}: full $= 0.539$, 10-shot $= 0.252$;
    \textsc{SPACE}: full $= 0.526$, 10-shot $= 0.307$).}
    \label{fig:kshot_enhancers}
\end{figure}

\begin{figure}[H]
    \centering
    \includegraphics[height=0.3\textheight]{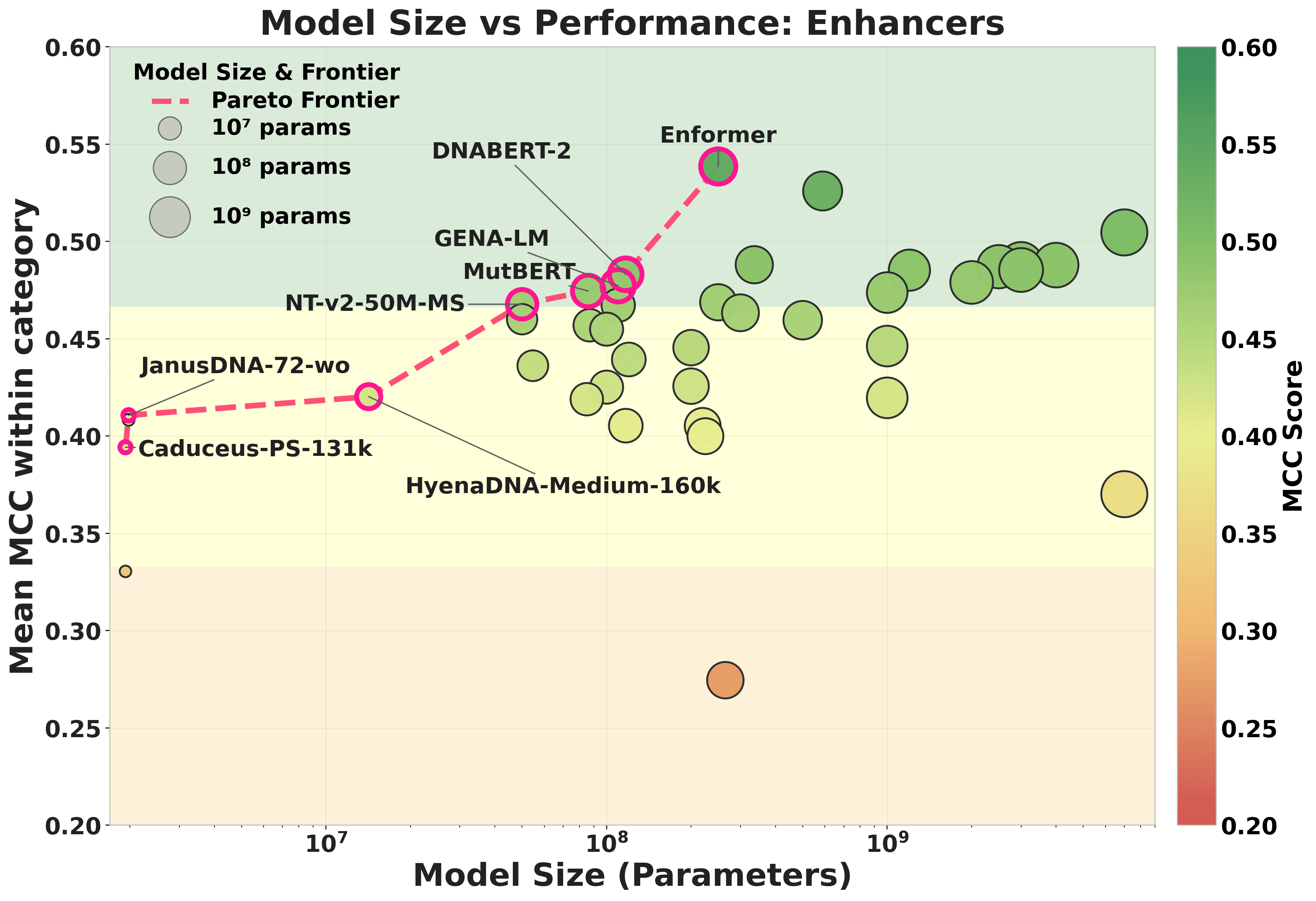}
    \caption{\textbf{Pareto frontier for Enhancer Prediction:
    mean MCC vs.\ parameter count.}
    Each point represents one of the 40 genomic foundation
    models, with parameter count on a logarithmic $x$-axis and
    mean full-shot Enhancer MCC on the $y$-axis. Marker size and
    color both encode MCC. The dashed line marks the Pareto
    frontier of best performance--size trade-offs. Scaling on
    this category is moderate (Spearman $\rho = 0.497$; tier gap $\geq 1$B vs.\ $<$200M of $0.036$ MCC). Specialized
    human-mouse epigenomic-profile models occupy the frontier at
    moderate sizes (\textsc{Enformer}, $252$M, MCC $= 0.539$;
    \textsc{SPACE}, $589$M, MCC $= 0.526$), outperforming
    multi-billion-parameter generalist models. \textsc{MutBERT}
    ($86$M, mean MCC $= 0.474$) is the strongest sub-100M model
    on this category.}
    \label{fig:pareto_enhancers}
\end{figure}

\begin{figure}[H]
    \centering
    \includegraphics[width=1.0\linewidth]{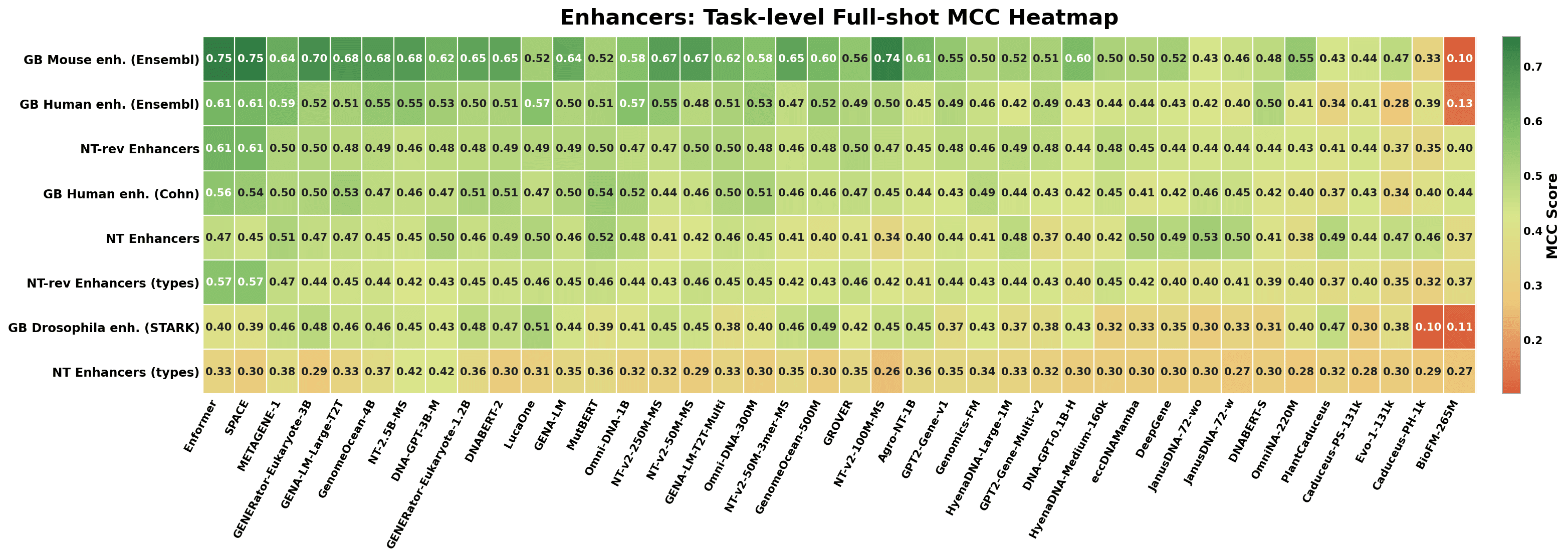}
    \caption{\textbf{Per-task MCC for Enhancer Prediction.}
    Heatmap shows full-shot MCC for each of the 40 genomic
    foundation models on the 8 enhancer prediction tasks, with
    models sorted by mean Enhancer MCC. Cell values report
    per-task MCC, with colors ranging from red/orange for lower
    scores to green for higher scores. Task difficulty varies
    substantially: the multi-class NT Enhancers (types) task is
    the most challenging (mean MCC $= 0.321$), while GB Mouse
    enh.\ (Ensembl) is the easiest (mean MCC $= 0.567$).
    Human-mouse epigenomic-profile models (\textsc{Enformer},
    \textsc{SPACE}) and Transformer-encoder models
    (\textsc{DNABERT-2}, \textsc{GENA-LM}, \textsc{GROVER})
    cluster at the top of the model ordering.}
    \label{fig:heatmap_enhancers}
\end{figure}

Enhancer prediction tasks ($n=8$ tasks) present moderate difficulty
with mean MCC of $0.446$ across the $40$ models. The category
spans human enhancer prediction (NT, NT-rev, and GB Cohn and
Ensembl subsets), enhancer-type classification (NT and NT-rev
multi-class variants), \textit{Drosophila} enhancers (GB Stark),
and a mouse enhancer task (GB Ensembl). Within-category
difficulty varies from a low of $0.321$ (NT Enhancers (types),
multi-class) to a high of $0.567$ (GB Mouse enh.\ (Ensembl)).
Results for the 8 enhancer prediction tasks are presented in
Figures~\ref{fig:kshot_enhancers}--\ref{fig:heatmap_enhancers}.

Human-mouse epigenomic profile models demonstrate the strongest
full-shot performance on enhancer tasks, consistent with their
specialized pretraining on regulatory elements.
\textsc{Enformer} achieves the top mean MCC on the category
($0.539$; rank $1$ of $40$ by mean Enhancer MCC) while ranking
$10$th of $40$ overall on the benchmark, a category-specific
advantage. \textsc{SPACE} follows closely (mean MCC $= 0.526$;
rank $2$ on Enhancers). The advantage reflects the close
relationship between enhancer sequences and the chromatin-state
signals present in human-mouse epigenomic profile pretraining
(Section~\ref{subsec:transfer_learning}).

Category-favoring specialization (defined as the gap between a
model's average per-task rank on the 92 non-enhancer tasks and
its average per-task rank on the 8 enhancer tasks) is observed
across diverse architectures rather than being confined to a
single architectural family. Beyond \textsc{Enformer}
($\Delta = +11.3$) and \textsc{SPACE} ($\Delta = +7.2$), notable
specializers include \textsc{JanusDNA-72-wo}
($\Delta = +7.6$; Hybrid-Mamba-MoE, $2$M parameters),
\textsc{GENA-LM} ($\Delta = +7.2$; Transformer-encoder, human-only,
110M), \textsc{DNABERT-2} ($\Delta = +6.4$;
Transformer-encoder, multi-species, 117M), and \textsc{GROVER}
($\Delta = +5.2$; Transformer-encoder, human-only, 87M).
That \textsc{JanusDNA-72-wo} appears among the top specializers at
$2$M parameters is noteworthy given its very small scale; among
sub-100M models on this category, however, \textsc{MutBERT} (86M,
mean MCC $= 0.474$) achieves the highest full-shot Enhancer MCC.
The architectural diversity of strong specializers argues against
attributing enhancer specialization to any single design choice.

Few-shot performance degrades substantially on enhancer tasks.
Across the 40 models, mean MCC drops from $0.446$ (full data) to
$0.278$ (10-shot) to $0.134$ (1-shot), corresponding to $37.6\%$
and $70.0\%$ relative reductions respectively. The 1-shot regime
collapses to near-random performance (maximum 1-shot MCC $=
0.195$ for \textsc{eccDNAMamba}). The 10-shot ranking differs
sharply from the full-shot ranking (Spearman $\rho = 0.64$;
top-5 overlap of only 1 model out of 5), with both
epigenomic-profile specialists losing their full-shot advantage:
\textsc{Enformer} drops from full-shot mean MCC $0.539$ (rank
$1$) to 10-shot MCC $0.252$ (rank $34$), and \textsc{SPACE}
drops from $0.526$ (rank $2$) to $0.307$ (rank $10$),
illustrating that strong full-data category specialization need
not translate to few-shot transferability. At 10-shot, the top
performers are \textsc{GENA-LM-Large-T2T} ($0.372$),
\textsc{Omni-DNA-300M} ($0.342$), and \textsc{Omni-DNA-1B}
($0.342$).

\subsubsection{DNA Methylation}

\begin{figure}[H]
    \centering
    \includegraphics[width=1.0\linewidth]{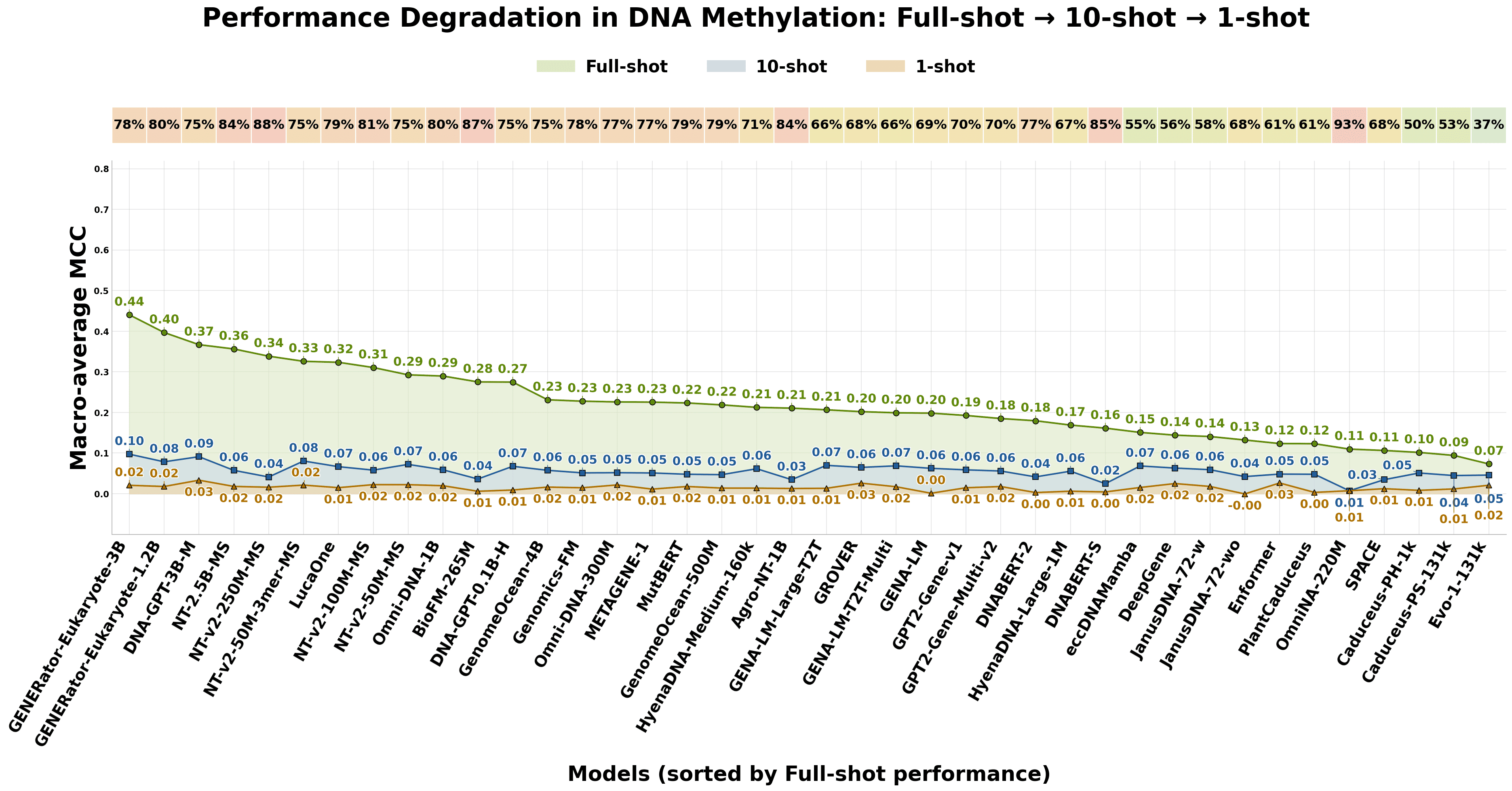}
    \caption{\textbf{Few-shot performance degradation on DNA Methylation.}
    For each of the 40 models, macro-average MCC across 8 DNA
    methylation tasks under full-shot, 10-shot, and 1-shot regimes;
    models ordered by full-shot performance. The top band shows the
    relative drop from full-shot to 10-shot per model. Benchmark-wide
    mean degradation: $74.7\%$ for 10-shot, $93.2\%$ for 1-shot -- 
    among the most severe of any task category. Both 10-shot
    (maximum MCC $= 0.097$) and 1-shot (maximum $= 0.033$) regimes
    collapse to near-random performance for all models; full-shot
    and 10-shot rankings correlate at Spearman $\rho = 0.50$.}
    \label{fig:kshot_methylation}
\end{figure}

\begin{figure}[H]
    \centering
    \includegraphics[height=0.3\textheight]{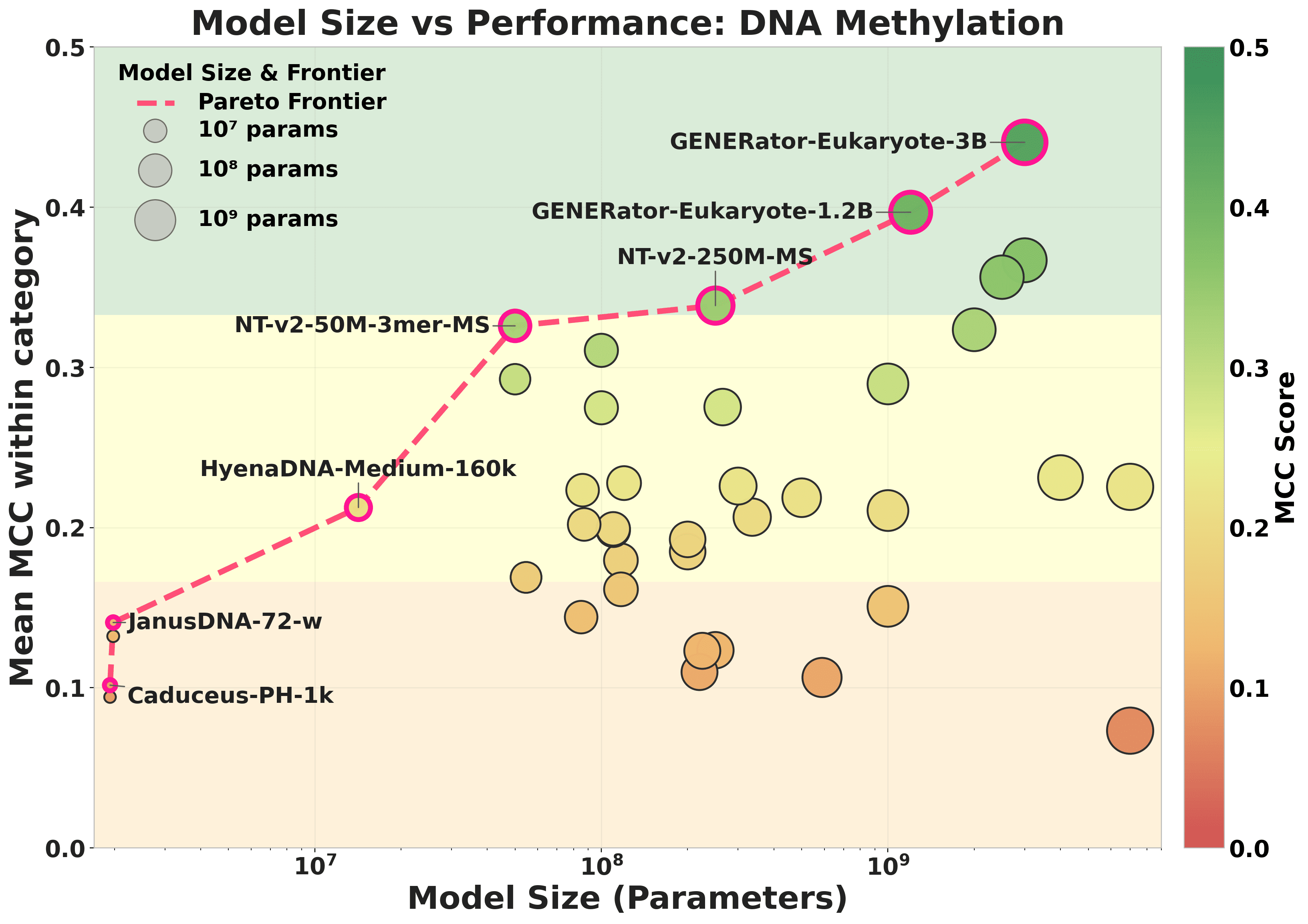}
    \caption{\textbf{Pareto frontier for DNA Methylation: mean MCC vs.\ parameter count.}
    Each point represents one of the 40 genomic foundation models,
    with parameter count on a logarithmic x-axis and mean full-shot
    DNA Methylation MCC on the y-axis. Marker size and color both
    encode MCC. The dashed line marks the Pareto frontier of best
    performance--size trade-offs. \textsc{NT-v2-50M-3mer-MS} (50M,
    mean MCC $= 0.326$) is the strongest sub-100M model on this
    category, ranking $6$th of $40$ overall and sitting near the
    frontier alongside multi-billion-parameter \textsc{GENERator}
    models.}
    \label{fig:pareto_methylation}
\end{figure}

\begin{figure}[H]
    \centering
    \includegraphics[width=1.0\linewidth]{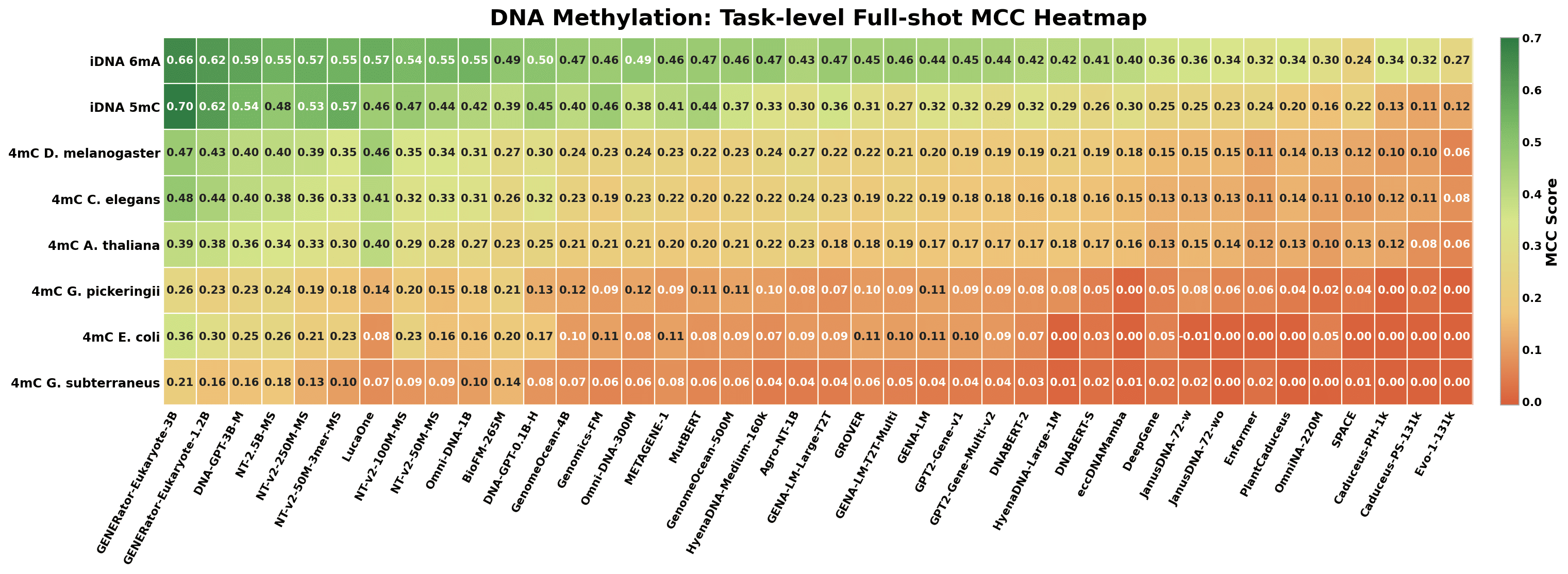}
    \caption{\textbf{Per-task MCC for DNA Methylation.}
    Heatmap shows full-shot MCC for each of the 40 genomic foundation
    models on the 8 DNA methylation tasks (six 4mC, one 5mC, one 6mA),
    with models sorted by mean DNA Methylation MCC. Cell values
    report per-task MCC, with colors ranging from red/orange for
    lower scores to green for higher scores.}
    \label{fig:heatmap_methylation}
\end{figure}

DNA methylation prediction ($n=8$ tasks) reveals the most
pronounced specialization effects in our benchmark. Task
difficulty spans an extreme range: the iDNA-ABF 6mA task
achieves $0.450$ mean MCC while the hardest 4mC task
(\textit{G.\ subterraneus}) approaches random performance at
$0.061$ MCC. Results for the 8 DNA methylation prediction tasks
are presented in
Figures~\ref{fig:kshot_methylation}--\ref{fig:heatmap_methylation}.

Model specialization is particularly striking. \textsc{BioFM-265M}
achieves rank $10.4$ on methylation tasks versus rank $36.7$ across
other categories -- a specialization score of $26.3$, the highest
in our benchmark. Similarly, \textsc{DNA-GPT-0.1B-H} shows rank
$10.6$ on methylation versus $27.0$ elsewhere (score $16.4$). All four NT-v2 multi-species variants demonstrate methylation
specialization (specialization scores ranging from $\Delta = 9.7$
for \textsc{NT-v2-50M-MS} to $\Delta = 12.9$ for
\textsc{NT-v2-100M-MS}).

Scaling on methylation tasks is modest. Spearman correlation between
model size and MCC is $\rho = 0.346$ ($p = 0.029$), weaker than the
benchmark-wide $\rho = 0.573$. This suggests that methylation
prediction depends more on specific sequence features captured
during pretraining than on model capacity.

Tokenization effects diverge from overall patterns. $k$-mer
tokenization shows clear advantages: \textsc{DNA-GPT-3B-M}
($k$-mer) exceeds \textsc{GenomeOcean-4B} (BPE) by $0.136$ MCC
($0.367$ vs.\ $0.231$) despite smaller size.
\textsc{NT-v2-250M-MS} ($k$-mer) outperforms
\textsc{GENA-LM-Large-T2T} (BPE) by $0.132$ MCC ($0.339$
vs.\ $0.207$). This advantage likely reflects the importance of
specific $k$-mer motifs in methylation site recognition.

Few-shot degradation is among the most severe in our benchmark:
$74.7\%$ at 10-shot ($0.219$ to $0.056$) and $93.2\%$ at 1-shot
($0.219$ to $0.015$). Even top-performing models collapse:
\textsc{GENERator-Eukaryote-3B} degrades from $0.440$ to $0.097$ at
10-shot and to $0.021$ at 1-shot. The 10-shot ranking remains
moderately correlated with full-shot (Spearman $\rho = 0.50$), with
\textsc{GENERator-Eukaryote-3B}, \textsc{DNA-GPT-3B-M}, and
\textsc{NT-v2-50M-3mer-MS} occupying the top three 10-shot
positions.

\subsubsection{Splice Site Detection}

\begin{figure}[h]
    \centering
    \includegraphics[width=1.0\linewidth]{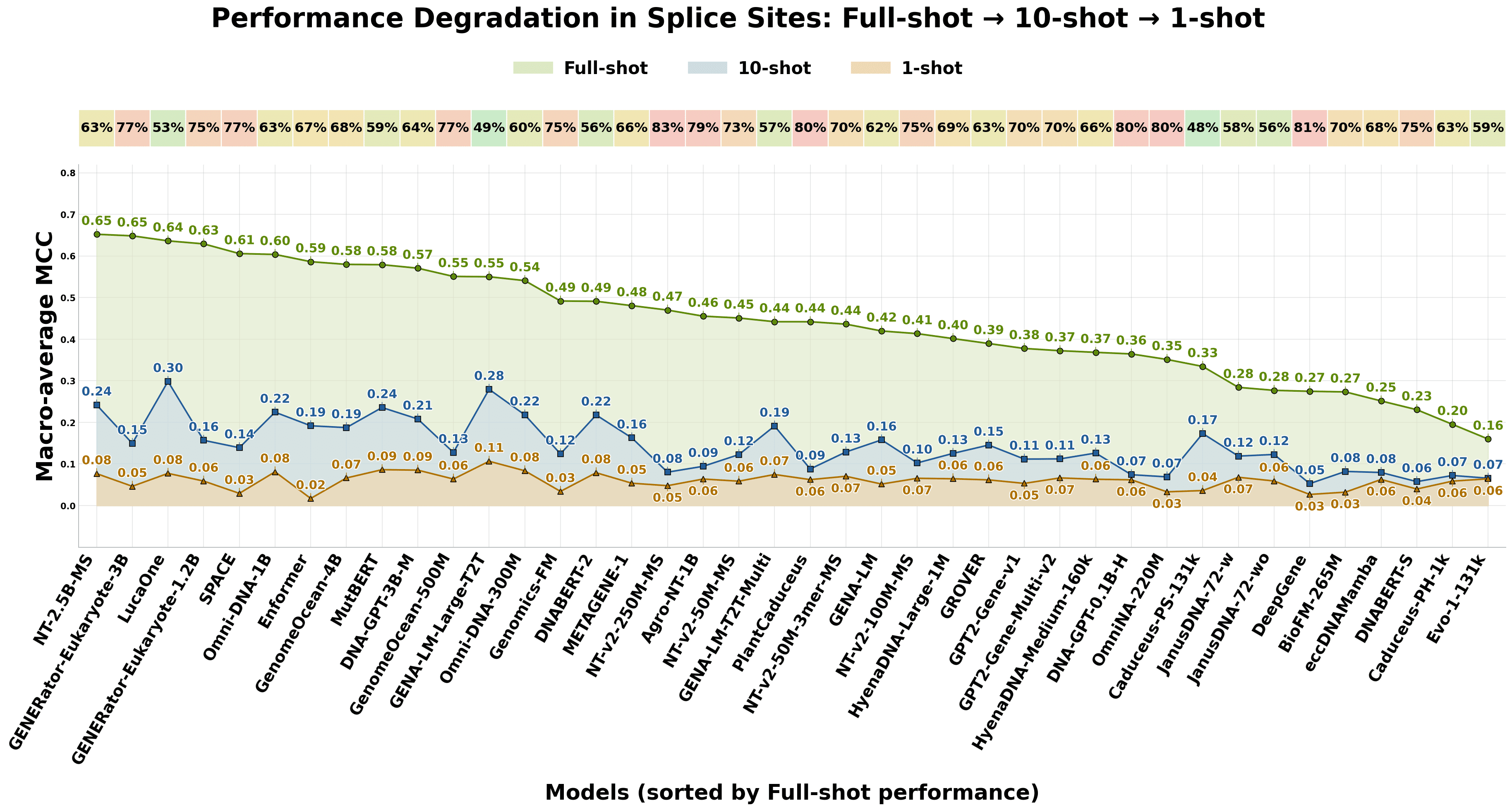}
    \caption{\textbf{Few-shot performance degradation on Splice
    Site Detection.} For each of the 40 models, macro-average MCC
    across 7 splice site tasks under full-shot, 10-shot, and
    1-shot regimes; models ordered by full-shot performance. The
    top band shows the relative drop from full-shot to 10-shot
    per model. Benchmark-wide mean degradation: $67.6\%$ for
    10-shot, $86.4\%$ for 1-shot. The 10-shot regime retains
    discriminative signal (maximum MCC $= 0.298$ for
    \textsc{LucaOne}), while the 1-shot regime collapses to
    near-random performance (maximum $= 0.107$). The 10-shot
    ranking diverges noticeably from full-shot (Spearman
    $\rho = 0.77$, top-5 overlap $2$ of $5$), with compact models
    such as \textsc{MutBERT} entering the 10-shot top-5.}
    \label{fig:kshot_splice}
\end{figure}

\begin{figure}[H]
    \centering
    \includegraphics[height=0.3\textheight]{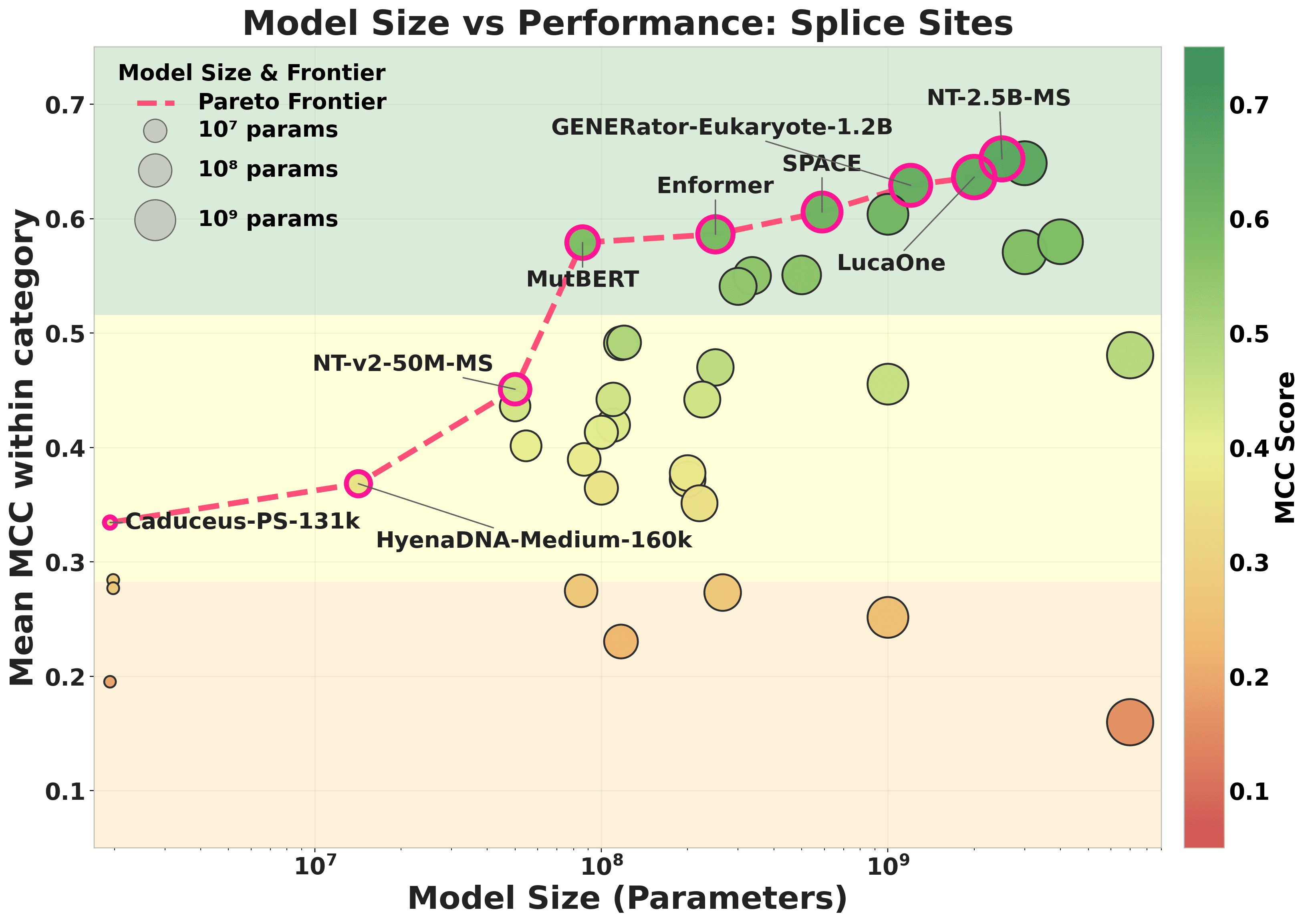}
    \caption{\textbf{Pareto frontier for Splice Site Detection:
    mean MCC vs.\ parameter count.}
    Each point represents one of the 40 genomic foundation
    models, with parameter count on a logarithmic x-axis and mean
    full-shot Splice Sites MCC on the y-axis. Marker size and
    color both encode MCC. The dashed line marks the Pareto
    frontier of best performance--size trade-offs. Scaling on
    this category is strong (Spearman $\rho = 0.547$, $p < 0.001$; $\rho = 0.667$ excluding prokaryotic \textsc{Evo-1-131k}).
    \textsc{MutBERT} (86M, mean MCC $= 0.579$) is the strongest
    sub-100M model on this category, ranking $9$th of $40$
    overall and sitting near the Pareto frontier alongside
    multi-billion-parameter models.}
    \label{fig:pareto_splice}
\end{figure}

\begin{figure}[H]
    \centering
    \includegraphics[width=1.0\linewidth]{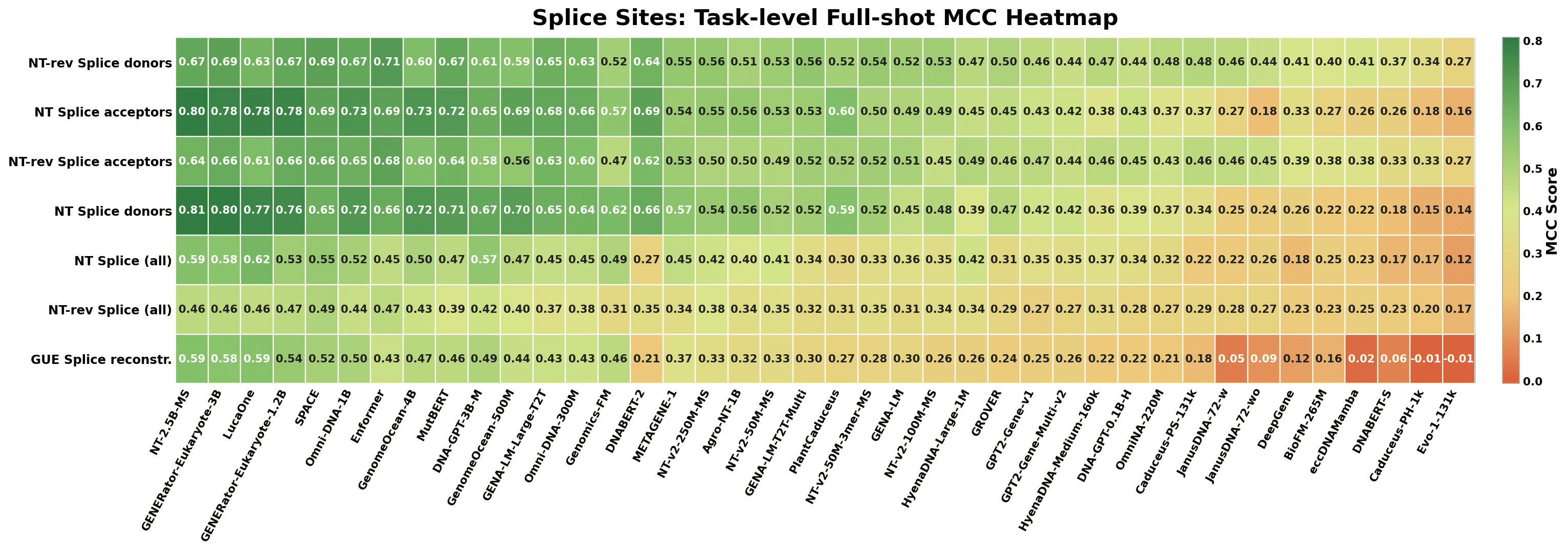}
    \caption{\textbf{Per-task MCC for Splice Site Detection.}
    Heatmap shows full-shot MCC for each of the 40 genomic
    foundation models on the 7 splice site tasks, with models
    sorted by mean Splice Sites MCC. Cell values report per-task
    MCC, with colors ranging from red/orange for lower scores to
    green for higher scores. Donor and acceptor classification
    tasks (NT and NT-revised sources) are consistently easier
    than the joint splice-sites-all task or the GUE reconstructed
    splice site challenge. Eukaryotic gene-focused and
    multi-species models dominate the top of the model ordering,
    with the human-mouse epigenomic-profile model \textsc{SPACE}
    in the top-5 and \textsc{MutBERT} (86M) entering the top-10
    despite its small scale.}
    \label{fig:heatmap_splice}
\end{figure}

Splice site detection ($n=7$ tasks) represents a critical test of
genomic sequence understanding, as accurate splicing prediction
requires recognition of complex sequence patterns spanning donor
and acceptor sites. Task difficulty varies from canonical
donor/acceptor classification (mean MCC $\approx 0.52$) to the
reconstructed splice site challenge (mean MCC $= 0.306$). Results
for the 7 splice site prediction tasks are presented in
Figures~\ref{fig:kshot_splice}--\ref{fig:heatmap_splice}.

The scale-performance paradox manifests dramatically for splice
sites. \textsc{GENERator-Eukaryote-1.2B} (1.2B parameters)
exceeds \textsc{Evo-1-131k} (7B parameters) by $0.469$ MCC ($0.629$
vs.\ $0.160$) -- among the largest pair-wise gaps observed across
any task category. This result directly reflects the fundamental
incompatibility between prokaryotic pretraining and eukaryotic
splicing prediction: prokaryotes lack the spliceosomal machinery
present in eukaryotes, rendering \textsc{Evo-1}'s 7B parameters
essentially uninformative for this task class.

Architecture comparisons reveal one of the largest
Transformer-over-SSM advantages observed in our benchmark.
\textsc{GenomeOcean-500M} (Transformer-decoder, 500M) exceeds
\textsc{eccDNAMamba} (Mamba, 537M) by $0.299$ MCC ($0.551$
vs.\ $0.252$) under matched multi-species/BPE conditions at
near-identical scale (1.07$\times$ ratio); the gap widens to
$0.352$ MCC when the larger \textsc{Omni-DNA-1B} (1B) is used
as the Transformer comparator. These substantial gaps suggest
that attention mechanisms are particularly well-suited for
capturing the long-range dependencies inherent in splice site
recognition.

Tokenization effects favor single-nucleotide approaches.
\textsc{MutBERT} (single-nucleotide) exceeds \textsc{GROVER} (BPE)
by $0.190$ MCC ($0.579$ vs.\ $0.390$) under matched
Transformer-encoder/human-pretraining conditions (both
$\approx 87$M), and exceeds \textsc{GENA-LM} (BPE,
Transformer-encoder, human) by $0.159$ MCC, although the
\textsc{MutBERT}--\textsc{GENA-LM} contrast is not strictly
matched on scale (86M vs.\ 110M). These advantages likely reflect
the importance of precise positional information at splice
junctions, which coarser subword tokenization may obscure.

Pretraining data effects are pronounced. Multi-species
\textsc{Genomics-FM} exceeds microbial \textsc{DNABERT-S} by
$0.261$ MCC ($0.492$ vs.\ $0.230$) -- the largest data-source gap
observed across task categories. Eukaryotic gene-focused
\textsc{GENERator-Eukaryote-3B} outperforms multi-species
\textsc{DNA-GPT-3B-M} by $0.077$ MCC ($0.648$ vs.\ $0.571$) under
matched Transformer-decoder/$k$-mer conditions (both 3B),
confirming the advantage of gene-centric pretraining for
splicing-related tasks.

\subsubsection{Long Non-coding RNA Classification}

\begin{figure}[H]
    \centering
    \includegraphics[width=1.0\linewidth]{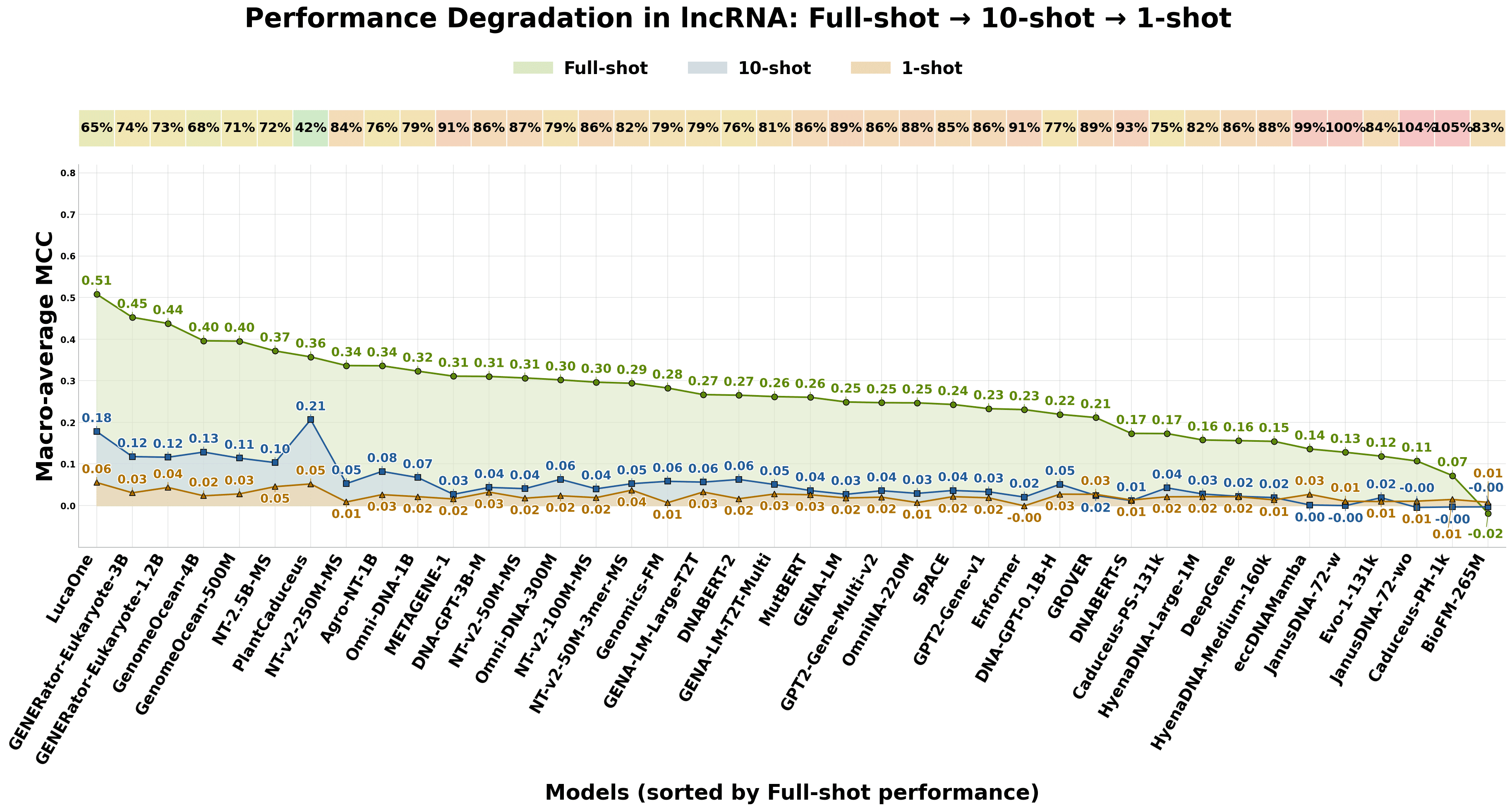}
    \caption{\textbf{Few-shot performance degradation on lncRNA
    Classification.} For each of the 40 models, macro-average MCC
    across 6 plant lncRNA tasks under full-shot, 10-shot, and
    1-shot regimes; models ordered by full-shot performance. The
    top band shows the relative drop from full-shot to 10-shot
    per model. Benchmark-wide mean degradation: $79.8\%$ for
    10-shot, $91.3\%$ for 1-shot. Both regimes collapse to
    near-random performance (10-shot maximum MCC $= 0.206$ for
    \textsc{PlantCaduceus}; 1-shot maximum $= 0.055$). Despite
    the collapse, the 10-shot ranking remains strongly correlated
    with full-shot (Spearman $\rho = 0.90$, top-5 overlap $4$ of
    $5$); notably, the plant-specialized
    \textsc{PlantCaduceus} rises from rank $7$ at full-shot to
    rank $1$ at 10-shot.}
    \label{fig:kshot_lncrna}
\end{figure}

\begin{figure}[H]
    \centering
    \includegraphics[height=0.3\textheight]{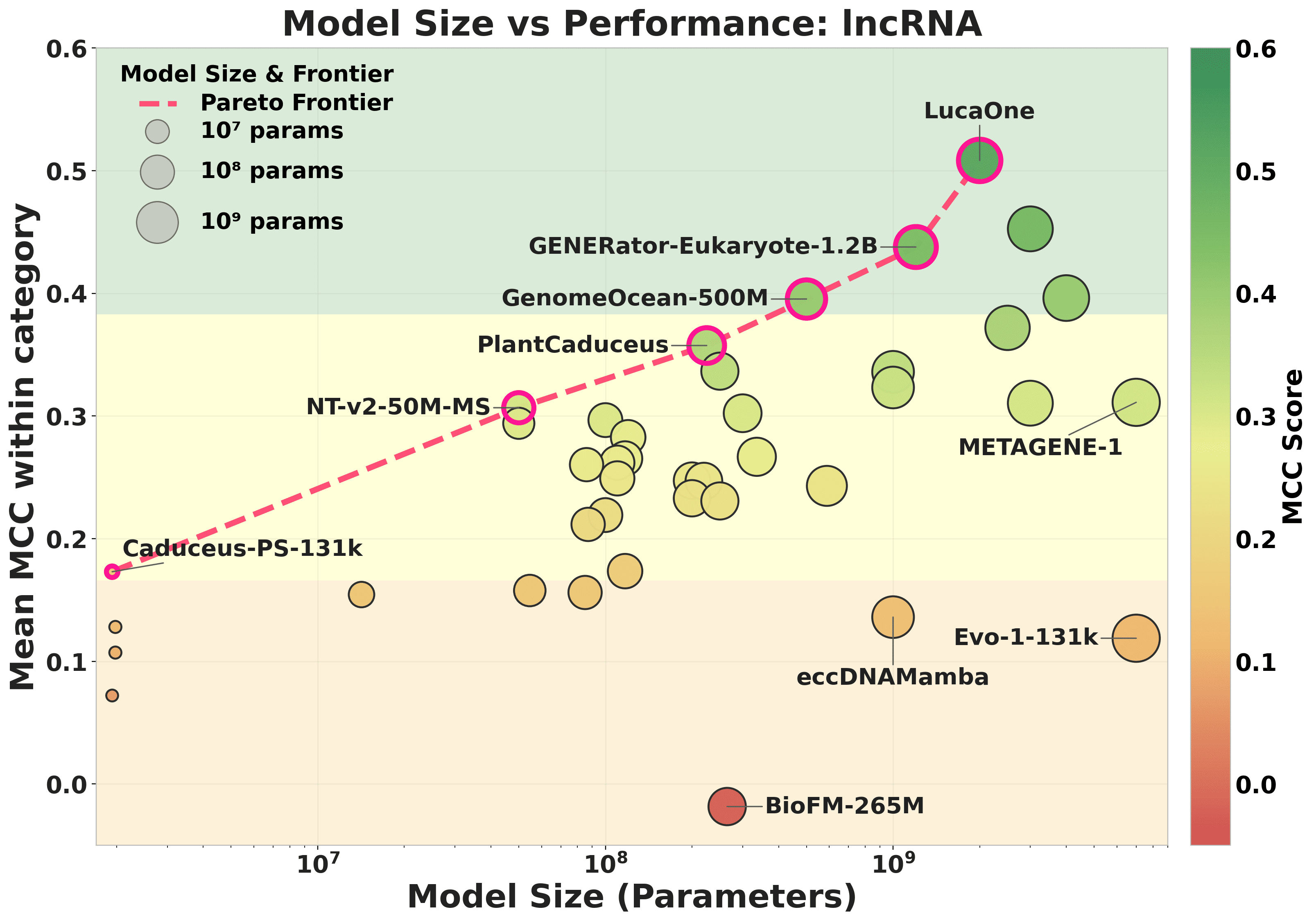}
    \caption{\textbf{Pareto frontier for lncRNA Classification:
    mean MCC vs.\ parameter count.}
    Each point represents one of the 40 genomic foundation
    models, with parameter count on a logarithmic x-axis and
    mean full-shot lncRNA MCC on the y-axis. Marker size and
    color both encode MCC. The dashed line marks the Pareto
    frontier of best performance--size trade-offs. Scaling on
    this category is moderate (Spearman $\rho = 0.575$, $p < 0.001$).
    The frontier reflects the plant-specific nature of the tasks:
    \textsc{PlantCaduceus} (225M, MCC $= 0.357$) and
    \textsc{Agro-NT-1B} (1B, MCC $= 0.336$) sit near the frontier
    despite modest scale, alongside multi-species models with much
    larger parameter counts. Among sub-100M models,
    \textsc{NT-v2-50M-MS} (50M, MCC $= 0.307$) achieves the
    highest lncRNA MCC.}
    \label{fig:pareto_lncrna}
\end{figure}

\begin{figure}[h]
    \centering
    \includegraphics[width=1.0\linewidth]{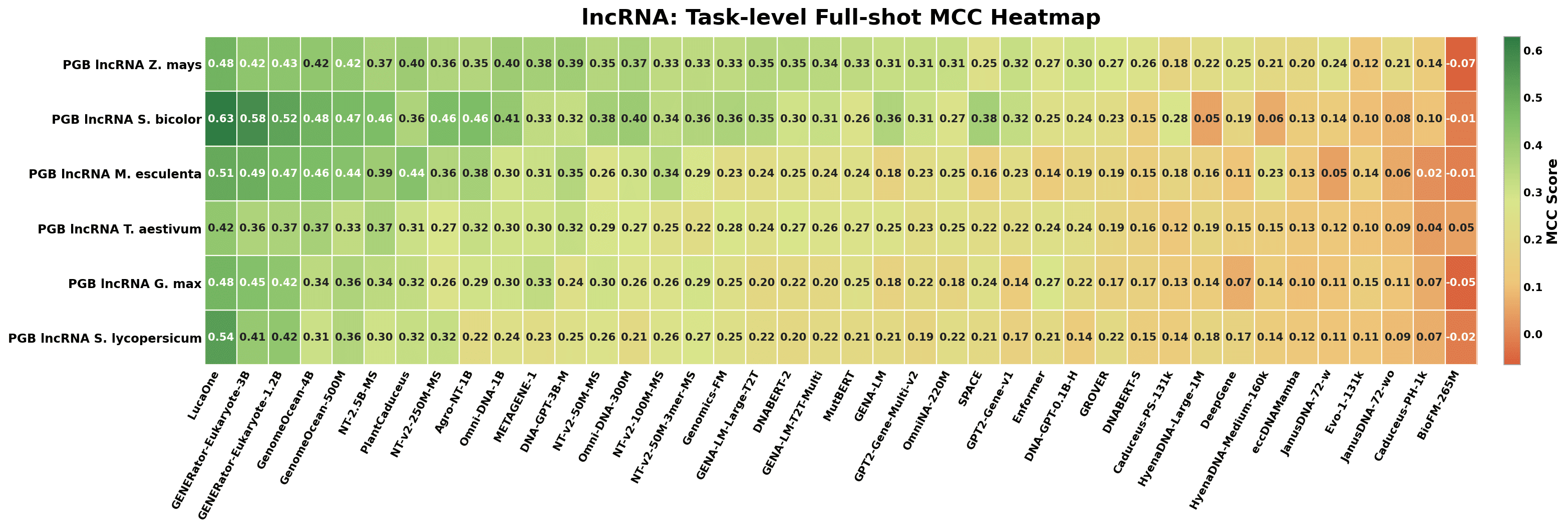}
    \caption{\textbf{Per-task MCC for lncRNA Classification.}
    Heatmap shows full-shot MCC for each of the 40 genomic
    foundation models on the 6 plant lncRNA classification tasks,
    with models sorted by mean lncRNA MCC. Cell values report
    per-task MCC, with colors ranging from red/orange for lower
    scores to green for higher scores. Multi-species
    (\textsc{LucaOne}) and eukaryotic gene-focused
    (\textsc{GENERator}) models dominate the top of the ordering,
    with plant-specialized models (\textsc{PlantCaduceus},
    \textsc{Agro-NT-1B}) close behind. Human-pretrained models
    cluster near the bottom.}
    \label{fig:heatmap_lncrna}
\end{figure}

The lncRNA category ($n=6$ tasks) exclusively comprises plant
species classification problems spanning \textit{Glycine max}
(soybean), \textit{Manihot esculenta} (cassava),
\textit{Sorghum bicolor} (sorghum), \textit{Solanum lycopersicum}
(tomato), \textit{Triticum aestivum} (wheat), and \textit{Zea
mays} (maize). This composition makes lncRNA tasks the primary
testbed for plant-specific transfer learning in our benchmark.
Results for the 6 lncRNA prediction tasks are presented in
Figures~\ref{fig:kshot_lncrna}--\ref{fig:heatmap_lncrna}.

Plant-specialized models demonstrate clear advantages.
\textsc{PlantCaduceus} achieves average per-task rank $8.0$ on
lncRNA tasks versus $27.0$ across the remaining $94$ tasks,
yielding a specialization gap of $\Delta = 19.0$ -- second only
to \textsc{BioFM-265M}'s methylation specialization
($\Delta = 26.3$). \textsc{Agro-NT-1B} shows a similar but smaller
pattern ($\Delta = 10.1$). These results validate the importance
of taxonomically-aligned pretraining for plant genomics
applications.

Human-trained models exhibit substantial negative transfer. The
best human-trained model (\textsc{MutBERT}, MCC $= 0.260$)
substantially underperforms plant-trained alternatives
(\textsc{PlantCaduceus}, MCC $= 0.357$). Across all human-trained
models, mean lncRNA MCC is only $0.164$ compared to $0.347$ for
plant-trained models -- a gap of $0.183$ MCC representing $112\%$
relative improvement. The negative transfer is particularly
striking for \textsc{BioFM-265M}, the only model with a negative
mean lncRNA MCC at full-shot ($-0.018$), indicating worse-than-
random performance on plant lncRNA classification despite the
model's strong methylation specialization.

Notably, the overall lncRNA leader \textsc{LucaOne} (multi-species,
mean MCC $= 0.508$) achieves the best per-category performance,
suggesting that sufficiently diverse pretraining can compensate
for the lack of plant-specific data. \textsc{LucaOne}'s advantage
over \textsc{PlantCaduceus} is consistent across all six
individual plant species (per-task $\Delta$ ranging from $+0.07$
on \textit{M.\ esculenta} to $+0.27$ on \textit{S.\ bicolor}),
indicating that broad taxonomic exposure can outweigh
plant-specific pretraining on this category even at the per-task
level.

Few-shot degradation is severe (79.8\% relative drop at 10-shot,
91.3\% at 1-shot), with all models collapsing to near-random
performance (10-shot maximum MCC $= 0.206$). The 10-shot ranking
remains strongly correlated with full-shot (Spearman $\rho = 0.90$),
but the plant-specialized \textsc{PlantCaduceus} rises from
full-shot rank $7$ to 10-shot rank $1$, while \textsc{LucaOne}
drops to rank $2$ -- indicating that plant-aligned pretraining
provides somewhat greater robustness under data scarcity even
when not yielding the highest full-shot score.

\newpage

\subsubsection{Mouse Enhancer Prediction}

\begin{figure}[h]
    \centering
    \includegraphics[width=1.0\linewidth]{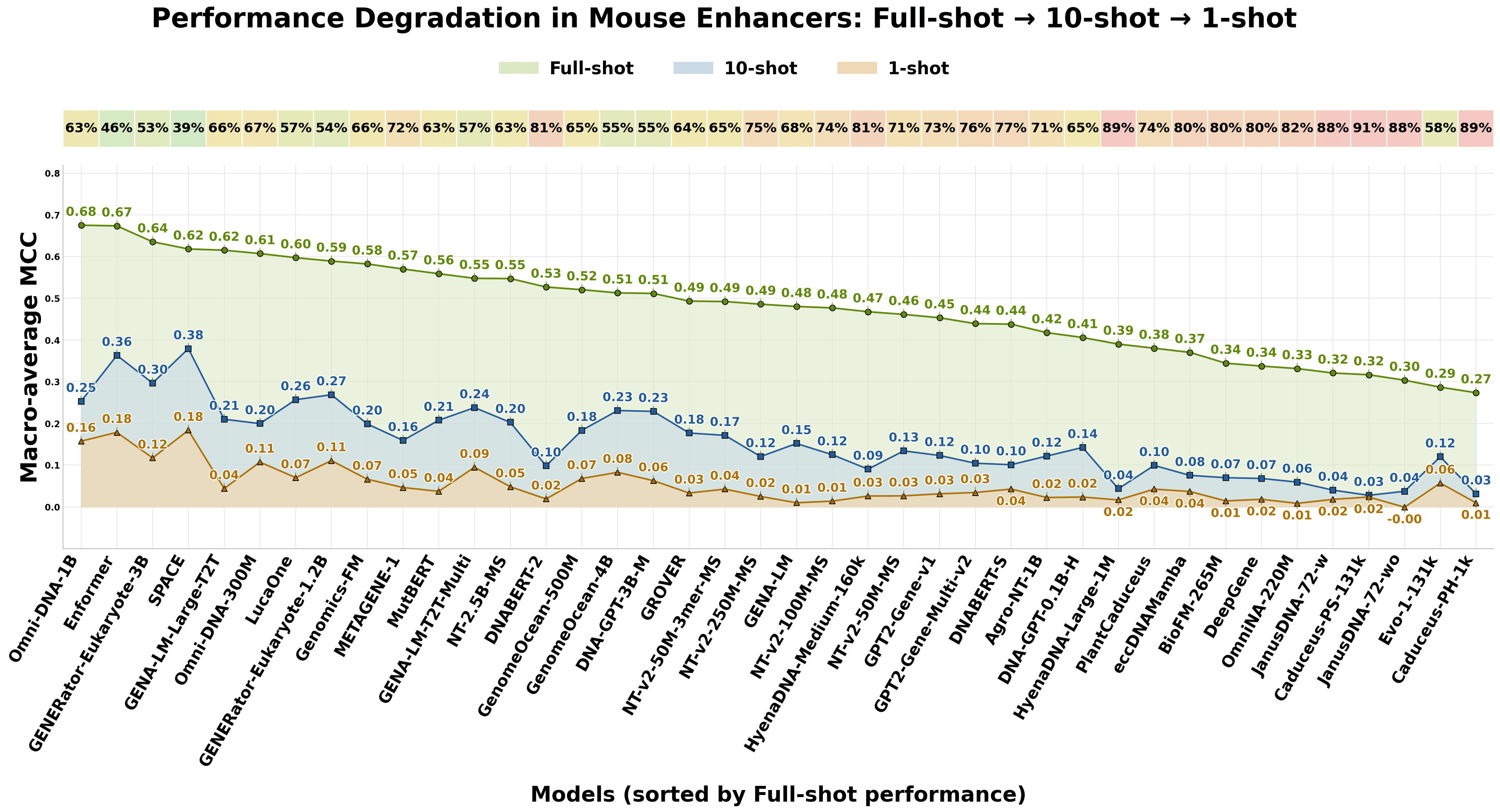}
    \caption{\textbf{Few-shot performance degradation on Mouse
    Enhancer Prediction.} For each of the 40 models, macro-average
    MCC across 5 mouse enhancer tasks under full-shot, 10-shot,
    and 1-shot regimes; models ordered by full-shot performance.
    The top band shows the relative drop from full-shot to 10-shot
    per model. Benchmark-wide mean degradation: $67.4\%$ for
    10-shot, $89.2\%$ for 1-shot. Unlike on the (human-centric)
    Enhancers category, human-mouse epigenomic-profile models
    retain their advantage under few-shot conditions:
    \textsc{SPACE} rises from full-shot rank $4$ to 10-shot rank
    $1$ (10-shot MCC $= 0.379$), and \textsc{Enformer} holds rank
    $2$ at both regimes (10-shot MCC $= 0.363$). Full-shot and
    10-shot rankings remain strongly correlated (Spearman
    $\rho = 0.90$, top-5 overlap $3$ of $5$).}
    \label{fig:kshot_mouse}
\end{figure}

\begin{figure}[H]
    \centering
    \includegraphics[height=0.3\textheight]{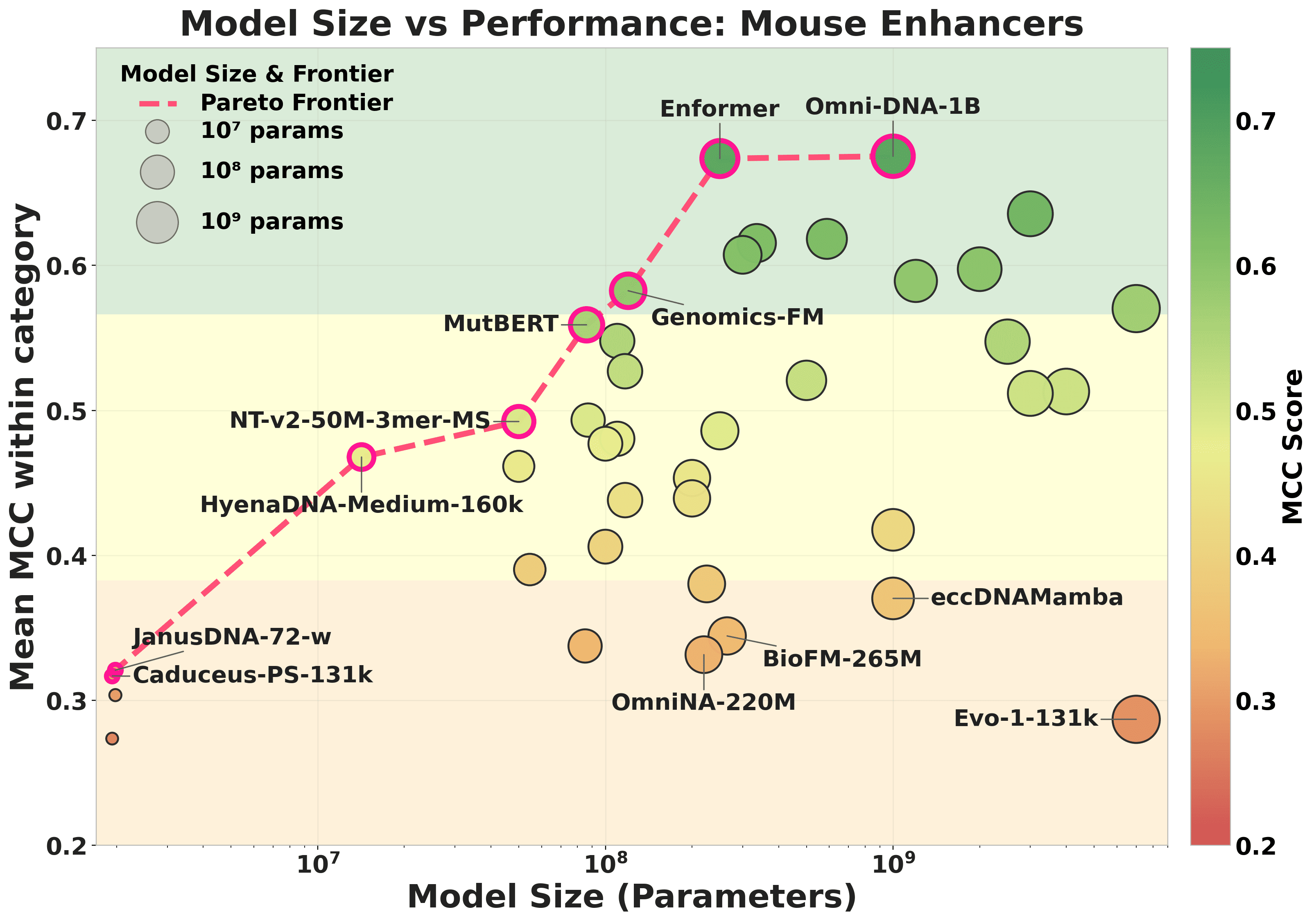}
    \caption{\textbf{Pareto frontier for Mouse Enhancer
    Prediction: mean MCC vs.\ parameter count.}
    Each point represents one of the 40 genomic foundation models,
    with parameter count on a logarithmic x-axis and mean
    full-shot Mouse Enhancer MCC on the y-axis. Marker size and
    color both encode MCC. The dashed line marks the Pareto
    frontier of best performance--size trade-offs. Scaling on
    this category is moderate (Spearman $\rho = 0.483$, $p = 0.002$; $\rho = 0.589$ excluding prokaryotic
    \textsc{Evo-1-131k}). Human-mouse epigenomic-profile models
    sit prominently on the frontier (\textsc{Enformer}, $252$M,
    MCC $= 0.674$; \textsc{SPACE}, $589$M, MCC $= 0.618$), and
    \textsc{MutBERT} (86M, mean MCC $= 0.559$) is the strongest
    sub-100M model on this category.}
    \label{fig:pareto_mouse}
\end{figure}

\begin{figure}[h]
    \centering
    \includegraphics[width=1.0\linewidth]{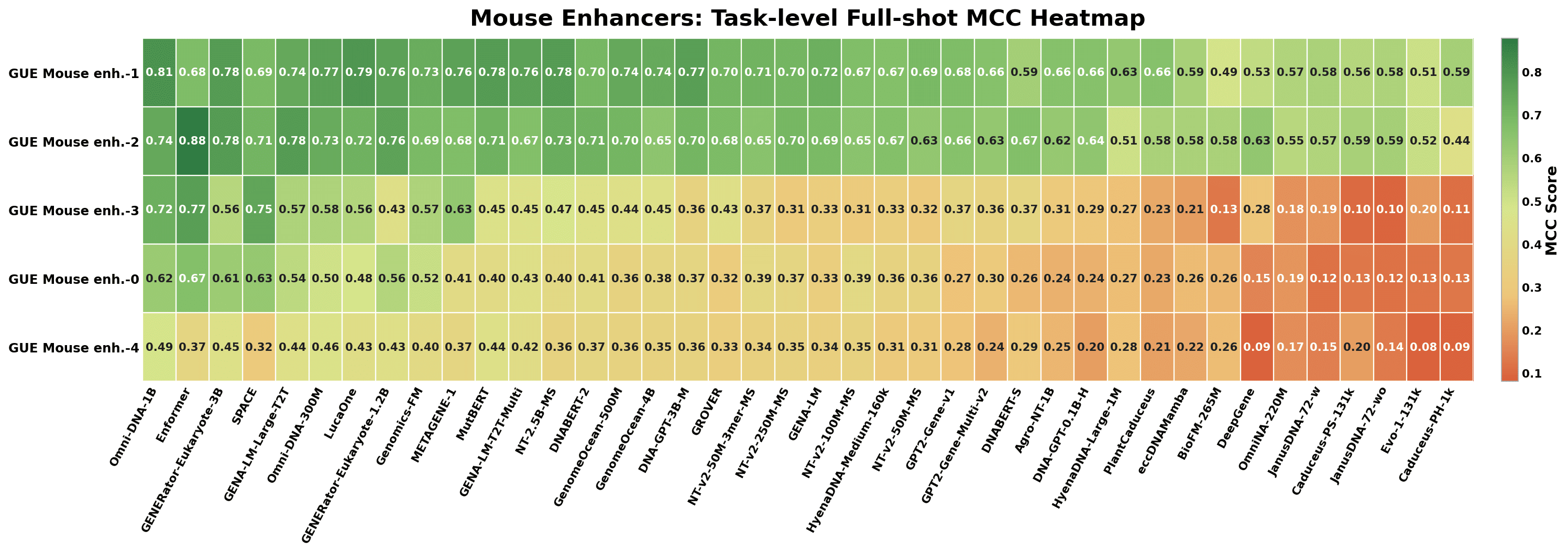}
    \caption{\textbf{Per-task MCC for Mouse Enhancer Prediction.}
    Heatmap shows full-shot MCC for each of the 40 genomic
    foundation models on the 5 mouse enhancer tasks (GUE mouse 0
    through 4), with models sorted by mean Mouse Enhancer MCC.
    Cell values report per-task MCC, with colors ranging from
    red/orange for lower scores to green for higher scores. Tasks
    1 and 2 are uniformly easier across models (mean MCC $> 0.65$)
    than tasks 0, 3, and 4 (mean MCC $< 0.39$). Human-mouse
    epigenomic-profile models (\textsc{Enformer}, \textsc{SPACE}),
    multi-species models (\textsc{Omni-DNA-1B}), and eukaryotic
    gene-focused models (\textsc{GENERator}) dominate the top of
    the model ordering.}
    \label{fig:heatmap_mouse}
\end{figure}

Mouse Enhancer prediction ($n=5$ tasks) evaluates cross-species
regulatory element recognition, providing insight into transfer
learning from human-centric or multi-species pretraining to a
related but distinct mammalian system. Task difficulty varies
substantially across the 5 tasks: task 1 is the easiest (mean MCC
$= 0.680$ across the 40 models) and task 4 is the hardest (mean
MCC $= 0.308$). Results for the 5 mouse enhancer prediction tasks
are presented in
Figures~\ref{fig:kshot_mouse}--\ref{fig:heatmap_mouse}.

Scaling on Mouse Enhancers is moderate and statistically
significant (Spearman $\rho = 0.483$, $p = 0.002$; $\rho = 0.589$
excluding the prokaryotic \textsc{Evo-1-131k}). Models with at
least 1B parameters average $0.534$ MCC versus $0.438$ for models below 200M -- a gap of $0.097$ MCC, larger than the benchmark-wide $+0.075$ tier gap.

Architecture comparisons reveal substantial Transformer-over-SSM
advantages. \textsc{GenomeOcean-500M} (Transformer-decoder,
500M) exceeds \textsc{eccDNAMamba} (Mamba, 537M) by $0.150$
MCC ($0.521$ vs.\ $0.370$) at near-matched scale under matched
multi-species/BPE conditions; the gap widens to $0.305$ MCC
when the larger \textsc{Omni-DNA-1B} (1B) is used as the
Transformer comparator. A within-Transformer encoder-vs-decoder
contrast on the same category shows \textsc{GENA-LM-Large-T2T}
(336M, encoder) outperforming \textsc{OmniNA-220M} (220M,
decoder) by $0.284$ MCC ($0.615$ vs.\ $0.332$), although the
contrast is not strictly matched on scale. This direction is opposite to the
encoder-vs-decoder result we observe on Histone Modifications and
is consistent with the broader observation
(Section~\ref{sec:results}) that the encoder--decoder distinction
is task- and setting-dependent.

Pretraining data effects follow expected patterns with notable
magnitude. The controlled \textsc{GENA-LM} comparison shows
multi-species outperforming human-only by $0.067$ MCC
(\textsc{GENA-LM-T2T-Multi} $= 0.548$ vs.\ \textsc{GENA-LM} $=
0.480$). Eukaryotic gene-focused \textsc{GENERator-Eukaryote-3B}
exceeds multi-species \textsc{DNA-GPT-3B-M} by $0.124$ MCC ($0.636$
vs.\ $0.512$) under matched Transformer-decoder/$k$-mer/3B
conditions, while multi-species \textsc{Genomics-FM} outperforms
microbial \textsc{DNABERT-S} by $0.144$ MCC under matched
Transformer-encoder conditions ($\approx$120M each).

Human-mouse epigenomic profile models show expected advantages
given their training data includes mouse regulatory elements.
\textsc{Enformer} (mean MCC $= 0.674$, rank $2$ of $40$) and
\textsc{SPACE} (mean MCC $= 0.618$, rank $4$ of $40$) rank among
the category leaders, with \textsc{Enformer} achieving the
notable result of outperforming \textsc{Evo-1-131k} (7B
parameters) by $0.387$ MCC despite being $28\times$ smaller.

\newpage

\subsubsection{Transcription Factor Binding}

\begin{figure}[h]
    \centering
    \includegraphics[width=1.0\linewidth]{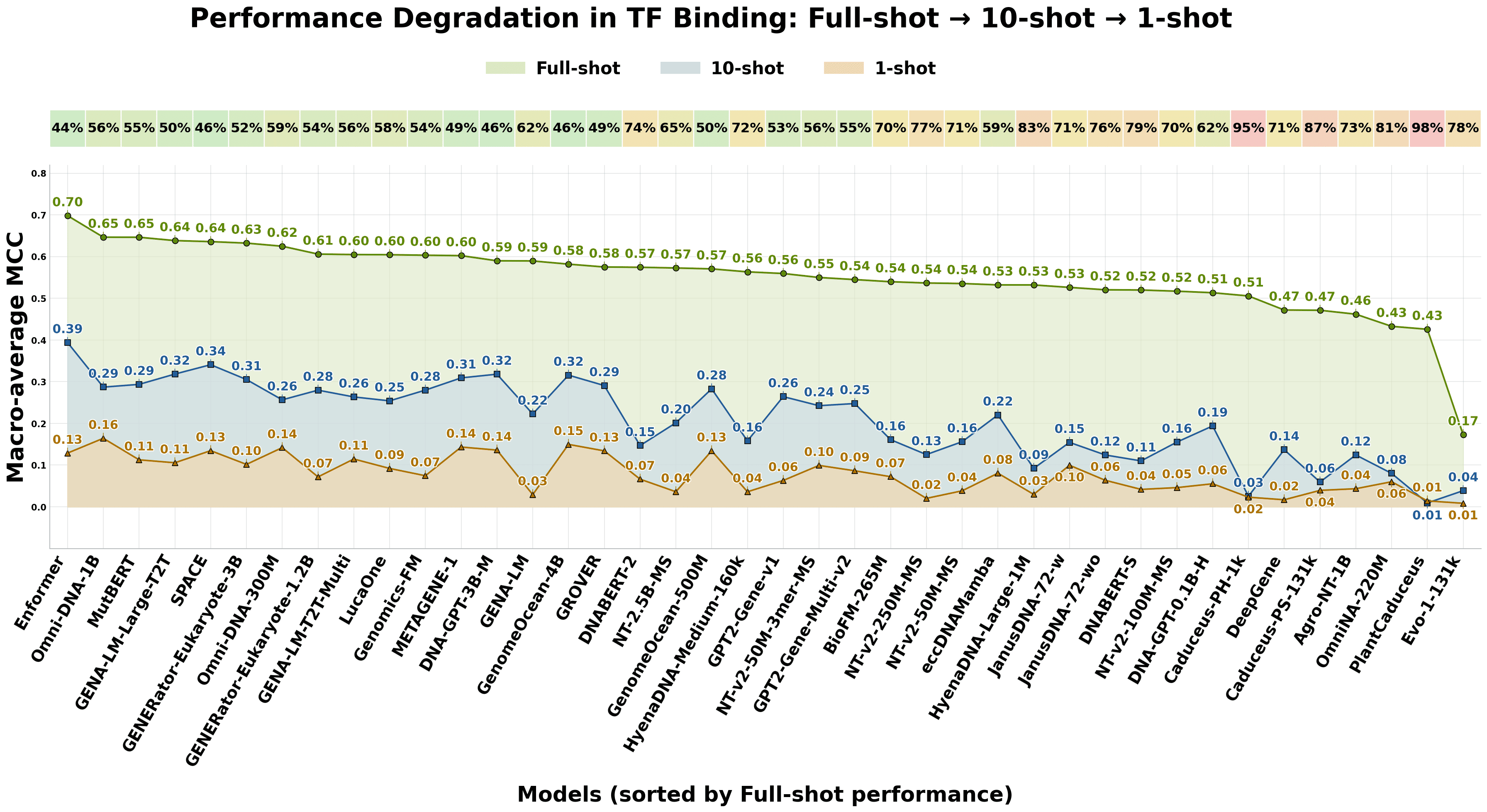}
    \caption{\textbf{Few-shot performance degradation on TF
    Binding.} For each of the 40 models, macro-average MCC
    across 5 TF binding tasks under full-shot, 10-shot, and
    1-shot regimes; models ordered by full-shot performance. The
    top band shows the relative drop from full-shot to 10-shot
    per model. Benchmark-wide mean degradation: $62.6\%$ for
    10-shot, $85.9\%$ for 1-shot. The 10-shot regime retains
    discriminative signal (maximum MCC $= 0.394$ for
    \textsc{Enformer}), while 1-shot collapses to near-random
    performance (maximum $= 0.164$). Full-shot and 10-shot
    rankings remain strongly correlated (Spearman $\rho = 0.88$,
    top-5 overlap $3$ of $5$); notably, \textsc{Enformer}
    retains rank $1$ at both regimes.}
    \label{fig:kshot_tf}
\end{figure}

\begin{figure}[H]
    \centering
    \includegraphics[height=0.3\textheight]{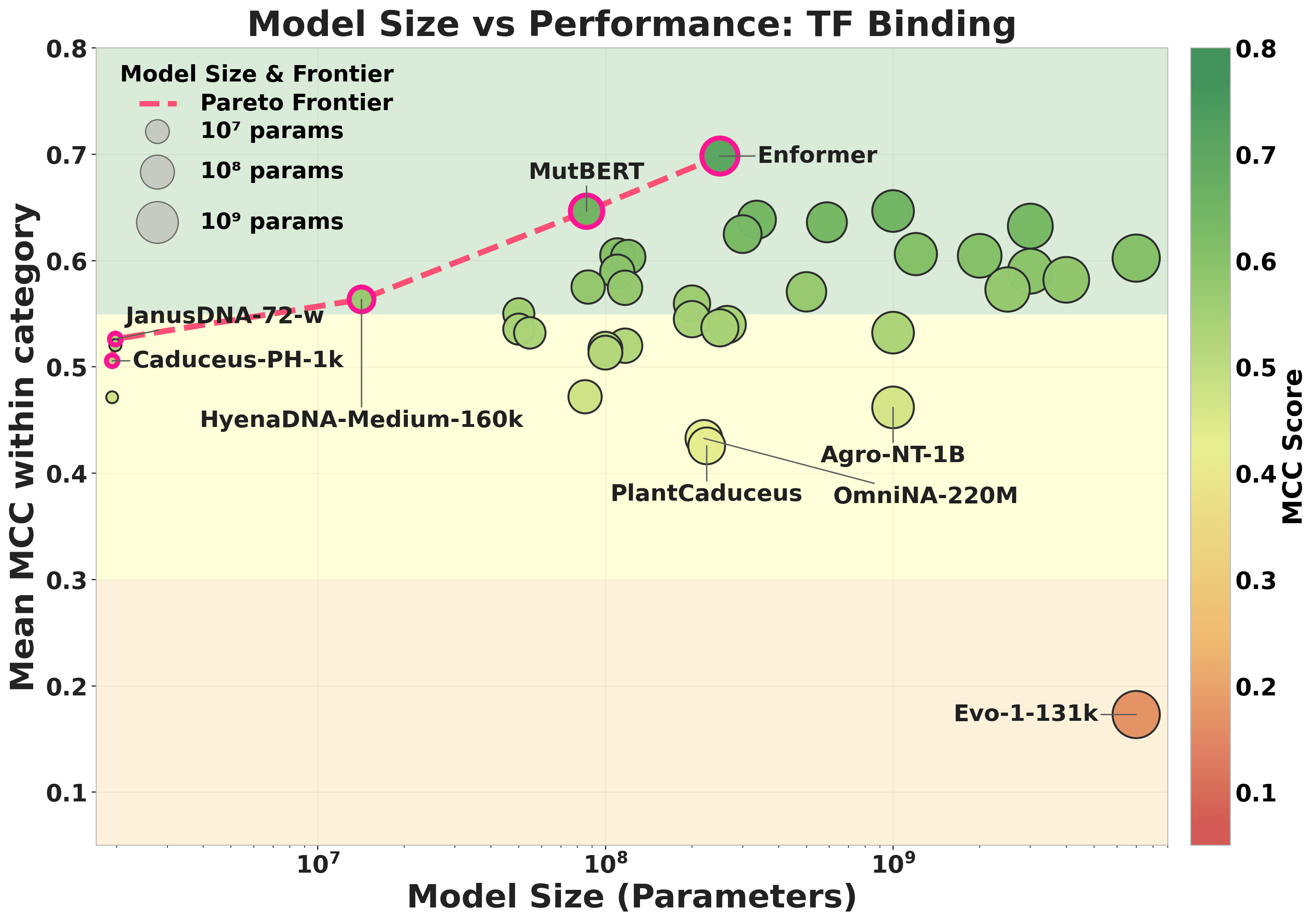}
    \caption{\textbf{Pareto frontier for TF Binding: mean MCC
    vs.\ parameter count.}
    Each point represents one of the 40 genomic foundation
    models, with parameter count on a logarithmic x-axis and
    mean full-shot TF Binding MCC on the y-axis. Marker size
    and color both encode MCC. The dashed line marks the Pareto
    frontier of best performance--size trade-offs. Scaling on
    this category is modest (Spearman $\rho = 0.361$, $p = 0.022$; $\rho = 0.466$ excluding prokaryotic
    \textsc{Evo-1-131k}), with models exceeding 1B parameters
    averaging essentially the same MCC as models below 200M
    ($0.547$ vs.\ $0.546$). \textsc{MutBERT} (86M, mean MCC $=
    0.646$) is the strongest sub-100M model on this category
    and ranks $3$rd of $40$ overall, between
    \textsc{Enformer} (252M) and \textsc{Omni-DNA-1B}.}
    \label{fig:pareto_tf}
\end{figure}

\begin{figure}[H]
    \centering
    \includegraphics[width=1.0\linewidth]{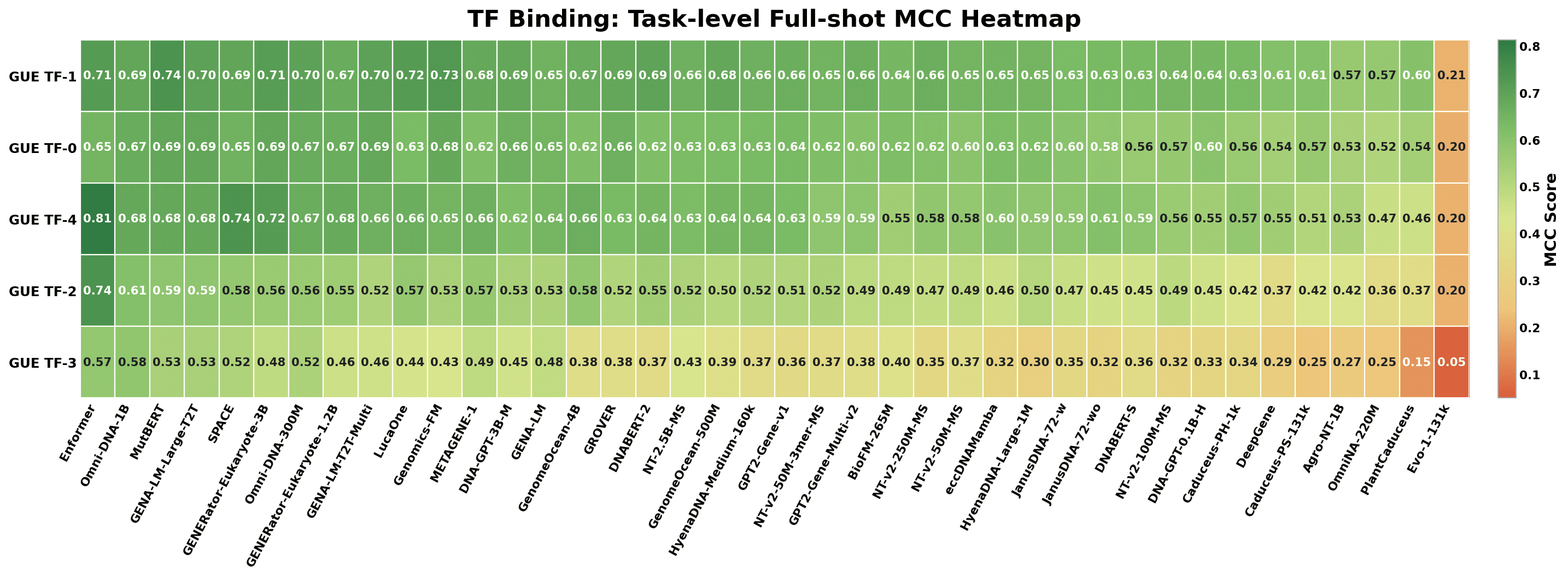}
    \caption{\textbf{Per-task MCC for TF Binding.}
    Heatmap shows full-shot MCC for each of the 40 genomic
    foundation models on the 5 TF binding tasks (GUE human TF
    0 through 4), with models sorted by mean TF Binding MCC.
    Cell values report per-task MCC, with colors ranging from
    red/orange for lower scores to green for higher scores.
    Task difficulty varies from GUE TF-3 (mean MCC $= 0.385$,
    hardest) to GUE TF-1 (mean MCC $= 0.652$, easiest).
    Human-mouse epigenomic-profile models (\textsc{Enformer},
    \textsc{SPACE}), Transformer-encoder models
    (\textsc{GENA-LM-Large-T2T}, \textsc{MutBERT}), and
    multi-species Transformer-decoder models
    (\textsc{Omni-DNA-1B}) dominate the top of the model
    ordering.}
    \label{fig:heatmap_tf}
\end{figure}

TF Binding prediction ($n=5$ tasks) represents a critical
challenge in regulatory genomics, requiring recognition of
sequence motifs that mediate protein-DNA interactions. This
category exhibits one of the most striking
scale-performance paradoxes in our benchmark. Results for the 5
transcription factor binding prediction tasks are presented in
Figures~\ref{fig:kshot_tf}--\ref{fig:heatmap_tf}.

Scaling on TF Binding is modest but statistically detectable
(Spearman $\rho = 0.361$, $p = 0.022$; $\rho = 0.466$ excluding
the prokaryotic \textsc{Evo-1-131k}). Models with at least 1B
parameters average $0.547$ MCC versus $0.546$ for models below
200M -- effectively no tier gap at the extremes, in contrast to
the benchmark-wide $+0.075$ gap. This flat-at-the-extremes
relationship reflects in large part the catastrophic failure of
\textsc{Evo-1-131k} (7B parameters, mean MCC $= 0.173$, rank
$40$ of $40$), which pulls the $\geq 1$B group mean downward.

The scale paradox reaches its apex for TF binding:
\textsc{Enformer} (252M parameters, mean MCC $= 0.698$, rank
$1$ of $40$) exceeds \textsc{Evo-1-131k} by $0.525$ MCC -- the
largest performance gap between any small-vs-large model pair
across our entire benchmark. This $28\times$ size disadvantage
yielding $4\times$ better performance starkly illustrates the
dominance of pretraining data alignment over model scale.

Architecture comparisons favor Transformer-encoder variants on
this category. \textsc{GENA-LM-Large-T2T} (encoder, 336M)
outperforms \textsc{OmniNA-220M} (decoder, 220M) by $0.206$ MCC
($0.638$ vs.\ $0.433$), and the smaller \textsc{GENA-LM-T2T-Multi}
(encoder, 110M) also outperforms \textsc{OmniNA-220M} (decoder,
220M) by $0.172$ MCC. The Transformer-over-SSM gap on this category is comparatively
modest under matched conditions:
\textsc{GenomeOcean-500M} (500M) exceeds \textsc{eccDNAMamba}
(537M) by only $0.038$ MCC ($0.571$ vs.\ $0.532$) at
near-matched scale, broadening to $0.114$ MCC when
\textsc{Omni-DNA-1B} is used as the Transformer comparator. These
pairwise contrasts are consistent with the broader observation
(Section~\ref{sec:results}) that the encoder--decoder ranking
on this benchmark is task- and setting-dependent; for TF
Binding the encoder direction holds across both pairs we examined.

Human-mouse epigenomic profile models again demonstrate
specialized strength. \textsc{Enformer} achieves the highest mean
MCC on the category ($0.698$, rank $1$ of $40$) and
\textsc{SPACE} follows closely ($0.636$, rank $5$ of $40$); on per-task average ranks these correspond to $4.0$ and $7.0$
respectively, reflecting consistent placement among the strongest
models across the 5 TF binding tasks (Enformer in top-5 on
$4$ of $5$ tasks; SPACE in top-11 on all $5$). This performance reflects the direct
relevance of TF binding events to the chromatin-state signals
present in human-mouse epigenomic profile pretraining
(Section~\ref{subsec:transfer_learning}).

\subsubsection{Species Classification}

\begin{figure}[H]
    \centering
    \includegraphics[width=1.0\linewidth]{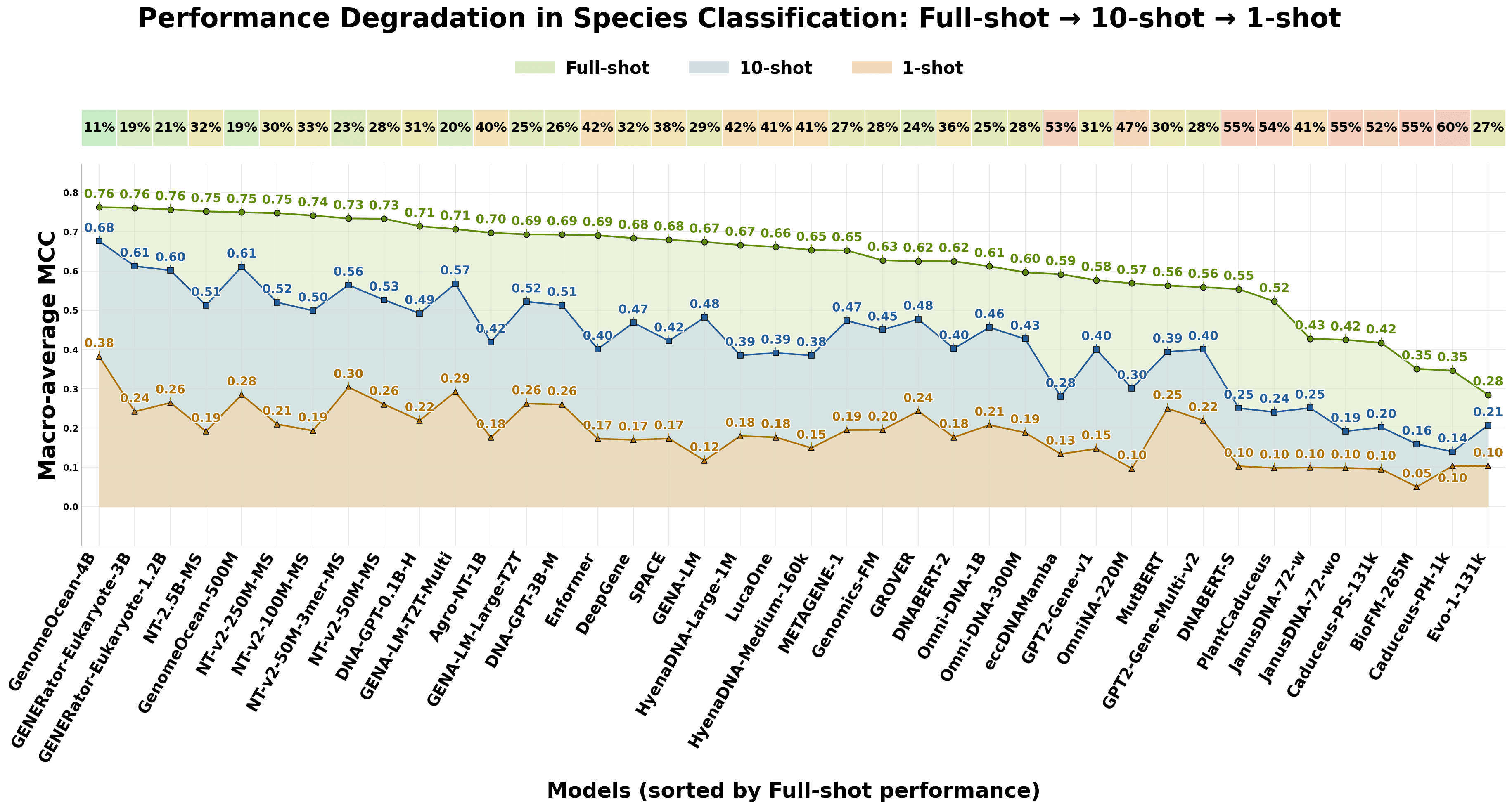}
    \caption{\textbf{Few-shot performance degradation on Species
    Classification.} For each of the 40 models, macro-average MCC
    across 3 species classification tasks under full-shot,
    10-shot, and 1-shot regimes; models ordered by full-shot
    performance. The top band shows the relative drop from
    full-shot to 10-shot per model. Benchmark-wide mean
    degradation: $33.0\%$ for 10-shot, $69.9\%$ for 1-shot --
    among the mildest few-shot degradation of any task category
    in our benchmark. Both 10-shot (maximum MCC $= 0.676$ for
    \textsc{GenomeOcean-4B}) and 1-shot (maximum $= 0.382$, also
    \textsc{GenomeOcean-4B}) regimes retain substantial
    discriminative signal, indicating that species classification
    is relatively robust to data scarcity. Full-shot and 10-shot
    rankings are very strongly correlated (Spearman $\rho = 0.91$,
    top-5 overlap $4$ of $5$).}
    \label{fig:kshot_species}
\end{figure}

\begin{figure}[H]
    \centering
    \includegraphics[height=0.3\textheight]{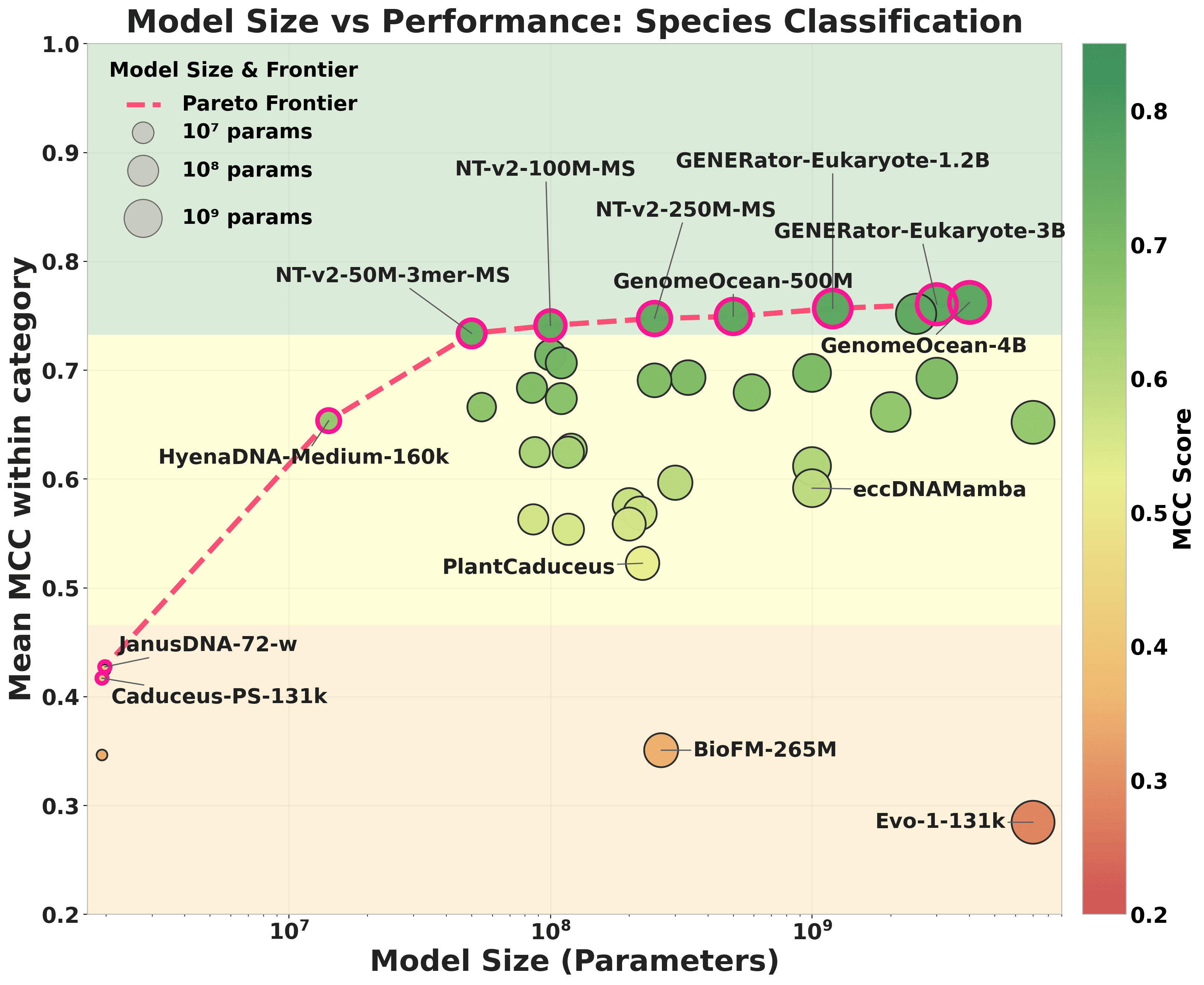}
    \caption{\textbf{Pareto frontier for Species Classification:
    mean MCC vs.\ parameter count.}
    Each point represents one of the 40 genomic foundation
    models, with parameter count on a logarithmic x-axis and
    mean full-shot Species Classification MCC on the y-axis.
    Marker size and color both encode MCC. The dashed line marks
    the Pareto frontier of best performance--size trade-offs.
    Scaling on this category is among the weakest in our
    benchmark (Spearman $\rho = 0.308$, $p = 0.054$; $\rho = 0.409$, $p = 0.009$ excluding the prokaryotic
    \textsc{Evo-1-131k}). \textsc{NT-v2-50M-3mer-MS} (50M, mean
    MCC $= 0.734$) is the strongest sub-100M model on this
    category, sitting near the Pareto frontier alongside
    multi-billion-parameter generalists.}
    \label{fig:pareto_species}
\end{figure}

\begin{figure}[h]
    \centering
    \includegraphics[width=1.0\linewidth]{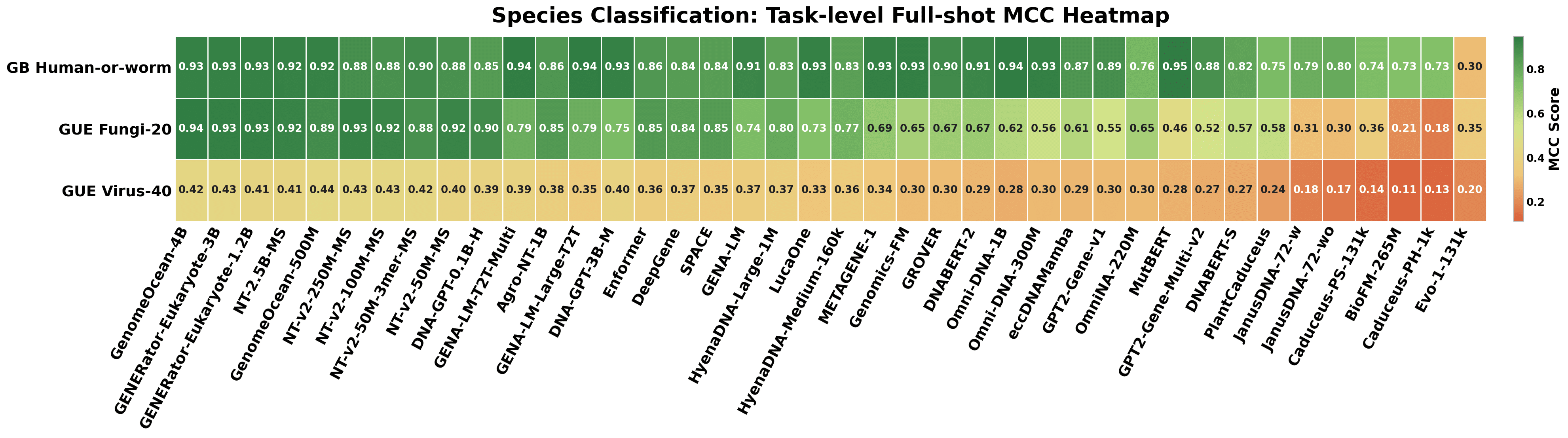}
    \caption{\textbf{Per-task MCC for Species Classification.}
    Heatmap shows full-shot MCC for each of the 40 genomic
    foundation models on the 3 species classification tasks,
    with models sorted by mean Species Classification MCC. Cell
    values report per-task MCC, with colors ranging from
    red/orange for lower scores to green for higher scores. The
    cross-kingdom GB Human-or-worm task is uniformly
    easy across models (mean MCC $= 0.857$), while the
    fine-grained GUE Virus-40 task is uniformly hard
    ($0.323$). Multi-species decoder models
    (\textsc{GenomeOcean-4B}, \textsc{GENERator-Eukaryote-3B})
    and Nucleotide Transformer encoder models cluster at the top
    of the model ordering.}
    \label{fig:heatmap_species}
\end{figure}

Species Classification tasks ($n=3$ tasks) evaluate a model's ability
to distinguish genomic sequences from different organisms,
testing whether pretraining captures species-specific sequence
signatures. The category consists of one cross-kingdom human/worm
classification (GB Human-or-worm) and two fine-grained
taxonomic tasks (GUE Fungi-20 and GUE Virus-40).
Task difficulty varies substantially: the cross-kingdom task
achieves mean MCC $= 0.857$ across the 40 models, while the GUE
Virus-40 task is the hardest at $0.323$. Results for the 3
species classification tasks are presented in
Figures~\ref{fig:kshot_species}--\ref{fig:heatmap_species}.

Scaling on Species Classification is among the weakest in our
benchmark (Spearman $\rho = 0.308$, $p = 0.054$, marginally non-significant; $\rho = 0.409$, $p = 0.009$ when excluding
\textsc{Evo-1-131k}). Models exceeding 1B parameters average $0.663$ MCC versus $0.606$ for models below 200M ($+0.057$ gap, below the benchmark-wide $+0.075$). This compressed
scaling reflects in part the strong performance of small
multi-species Nucleotide Transformer variants on this category,
with \textsc{NT-v2-50M-3mer-MS} (50M) achieving $0.734$ MCC --
within $0.030$ of the top full-shot model \textsc{GenomeOcean-4B}
(4B, MCC $= 0.762$) at $1/80$ the parameter count.

The scale paradox is particularly visible for species
classification, with $122$ pairwise cases where smaller models
outperform larger counterparts by at least $5\times$ size
ratio. \textsc{NT-v2-50M-3mer-MS} (50M) exceeds
\textsc{Evo-1-131k} (7B) by $0.449$ MCC despite a $140\times$
size disadvantage -- one of the largest such gaps observed in
our benchmark, comparable in magnitude to the
\textsc{Enformer}--\textsc{Evo-1-131k} contrast on TF Binding
($0.525$; Section~\ref{subsec:transfer_learning}).

Architecture comparisons favor Transformer-encoder variants under
matched multi-species/BPE conditions. \textsc{GENA-LM-Large-T2T}
(encoder, 336M) outperforms \textsc{OmniNA-220M} (decoder, 220M)
by $0.125$ MCC ($0.693$ vs.\ $0.569$), and the smaller
\textsc{GENA-LM-T2T-Multi} (encoder, 110M) outperforms the same
baseline by $0.138$ MCC despite half the parameter count of
\textsc{OmniNA-220M}. As elsewhere in the benchmark
(Section~\ref{sec:results}), the encoder--decoder ranking is
task- and setting-dependent; for Species Classification the
encoder direction holds across both pairs we examined.

Pretraining data effects are pronounced. Multi-species training
strongly outperforms human-only: \textsc{DNA-GPT-0.1B-H}
(multi-species, 100M, Transformer-decoder, $k$-mer) exceeds
\textsc{GPT2-Gene-v1} (human, 200M, Transformer-decoder, BPE) by
$0.137$ MCC ($0.714$ vs.\ $0.576$); the two models share
architecture family but differ in tokenization, so the contrast is
not a clean pretraining-only comparison. Multi-species also
outperforms microbial: \textsc{NT-v2-100M-MS} outperforms
\textsc{DNABERT-S} (multi-species-microbial) by $0.187$ MCC
($0.741$ vs.\ $0.554$) under matched Transformer-encoder/$k$-mer
conditions (100M vs.\ 117M). These patterns logically reflect
that species classification requires exposure to diverse
taxonomic sequences during pretraining.

The Nucleotide Transformer v2 multi-species family shows clear
specialization for this category, with all four NT-v2
multi-species variants exhibiting positive specialization
(\textsc{NT-v2-100M-MS}: per-task rank $10.7$ on Species
Classification vs.\ $19.9$ elsewhere, $\Delta = 9.2$;
\textsc{NT-v2-250M-MS}: $\Delta = 7.7$;
\textsc{NT-v2-50M-3mer-MS}: $\Delta = 5.9$;
\textsc{NT-v2-50M-MS}: $\Delta = 5.8$).

\subsubsection{Regulatory Element Prediction}

\begin{figure}[h]
    \centering
    \includegraphics[width=1.0\linewidth]{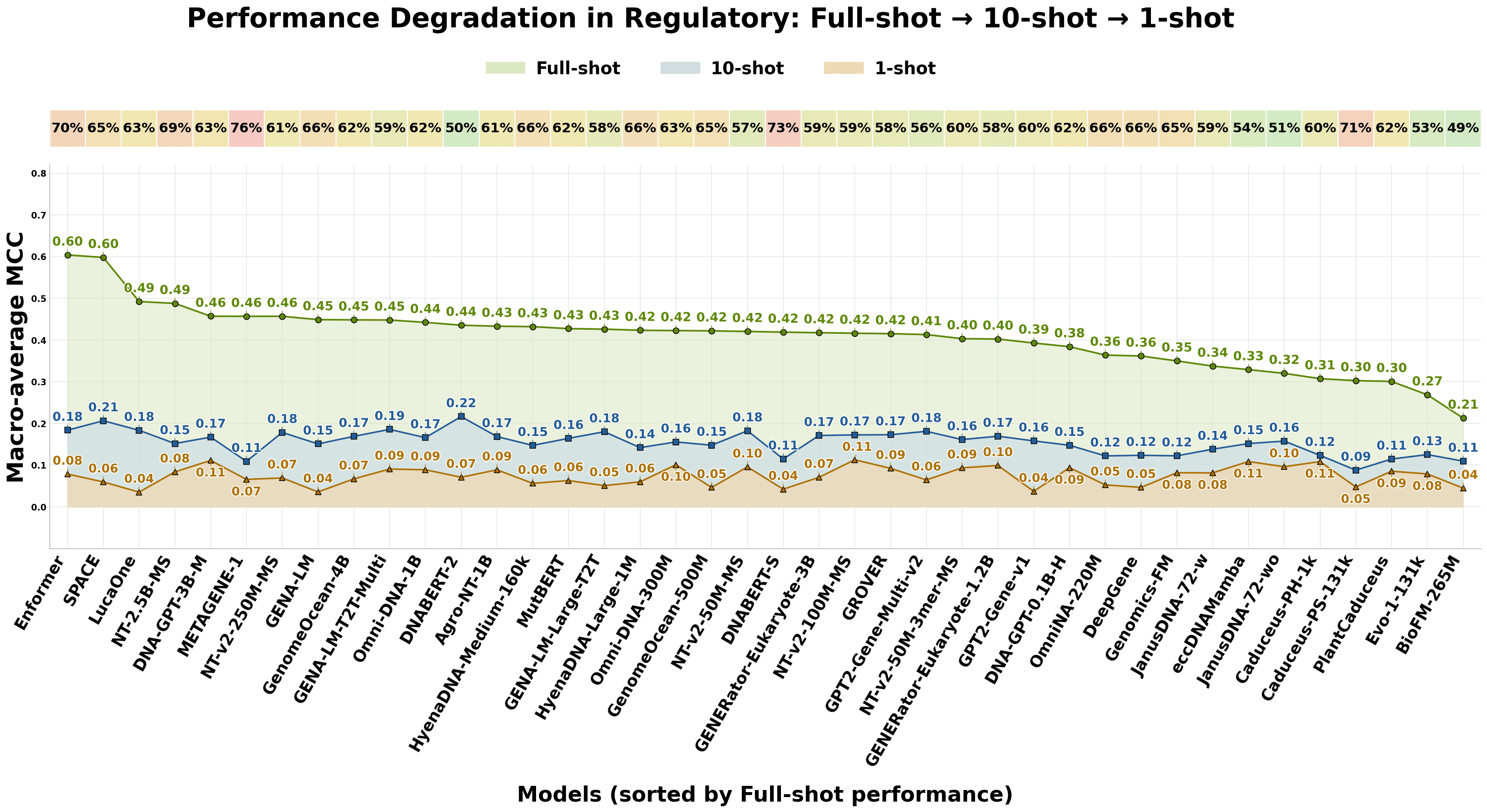}
    \caption{\textbf{Few-shot performance degradation on
    Regulatory Element Prediction.} For each of the 40 models,
    macro-average MCC across 2 regulatory element tasks under
    full-shot, 10-shot, and 1-shot regimes; models ordered by
    full-shot performance. The top band shows the relative drop
    from full-shot to 10-shot per model. Benchmark-wide mean
    degradation: $62.1\%$ for 10-shot, $81.9\%$ for 1-shot. The
    10-shot ranking differs moderately from full-shot (Spearman
    $\rho = 0.58$, top-5 overlap $3$ of $5$):
    \textsc{Enformer} drops from full-shot rank $1$ (MCC $=
    0.604$) to 10-shot rank $4$ (MCC $= 0.184$), while
    \textsc{SPACE} retains its top-2 position (full rank $2$
    $\rightarrow$ 10-shot rank $2$, MCC $= 0.206$). Both 10-shot
    (maximum MCC $= 0.216$ for \textsc{DNABERT-2}) and 1-shot
    (maximum $= 0.113$) regimes collapse to near-random
    performance across most models.}
    \label{fig:kshot_regulatory}
\end{figure}

\begin{figure}[H]
    \centering
    \includegraphics[height=0.3\textheight]{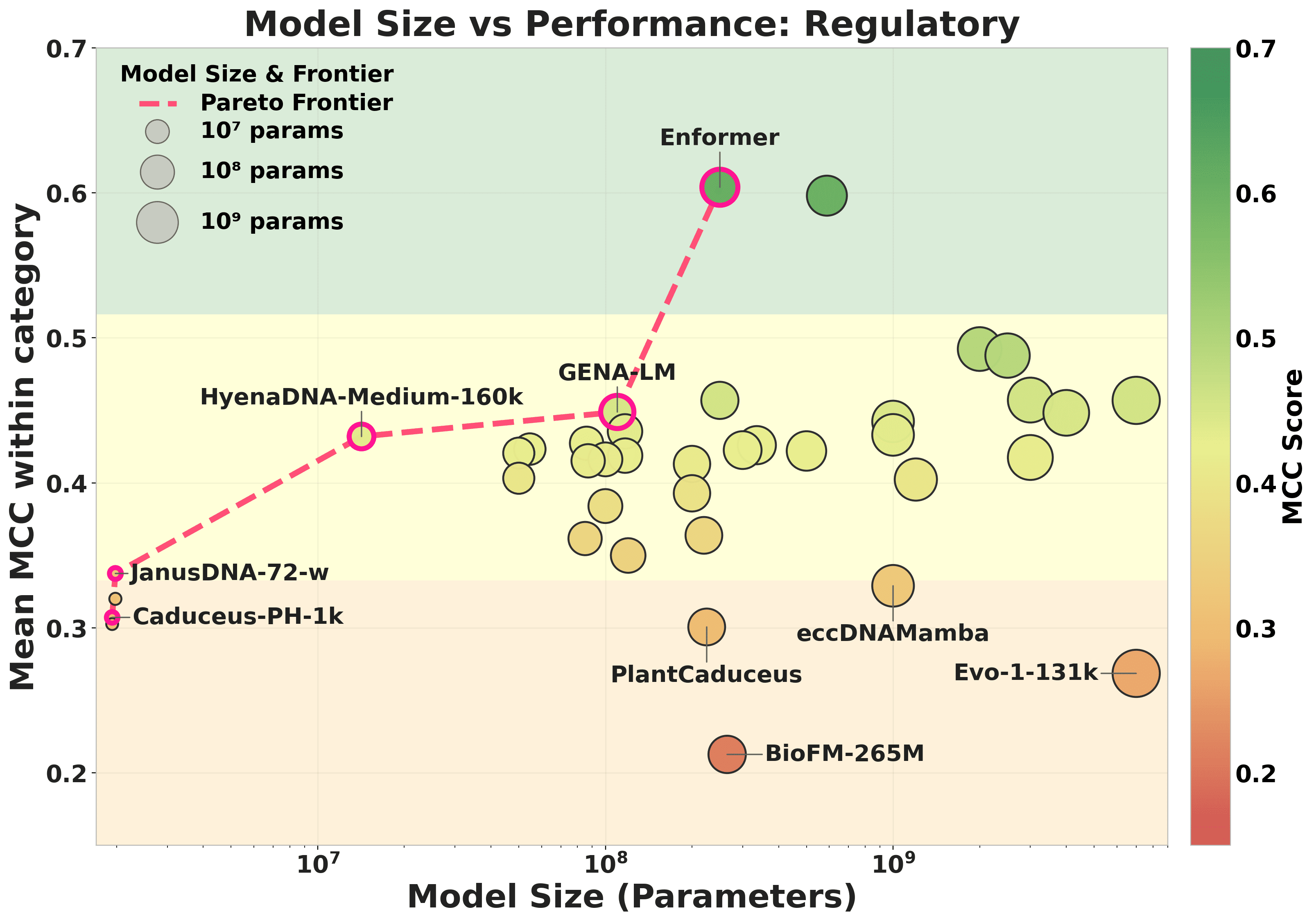}
    \caption{\textbf{Pareto frontier for Regulatory Element
    Prediction: mean MCC vs.\ parameter count.}
    Each point represents one of the 40 genomic foundation
    models, with parameter count on a logarithmic x-axis and
    mean full-shot Regulatory MCC on the y-axis. Marker size and
    color both encode MCC. The dashed line marks the Pareto
    frontier of best performance--size trade-offs. Scaling on
    this category is modest (Spearman $\rho = 0.387$, $p = 0.014$; $\rho = 0.490$, $p = 0.002$ excluding
    prokaryotic \textsc{Evo-1-131k}). Human-mouse
    epigenomic-profile models occupy the frontier at moderate
    sizes (\textsc{Enformer}, $252$M, MCC $= 0.604$;
    \textsc{SPACE}, $589$M, MCC $= 0.598$), outperforming all
    multi-billion-parameter generalists. \textsc{HyenaDNA-Medium}
    (14M, mean MCC $= 0.432$) is the strongest sub-100M model on
    this category.}
    \label{fig:pareto_regulatory}
\end{figure}

\begin{figure}[H]
    \centering
    \includegraphics[width=1.0\linewidth]{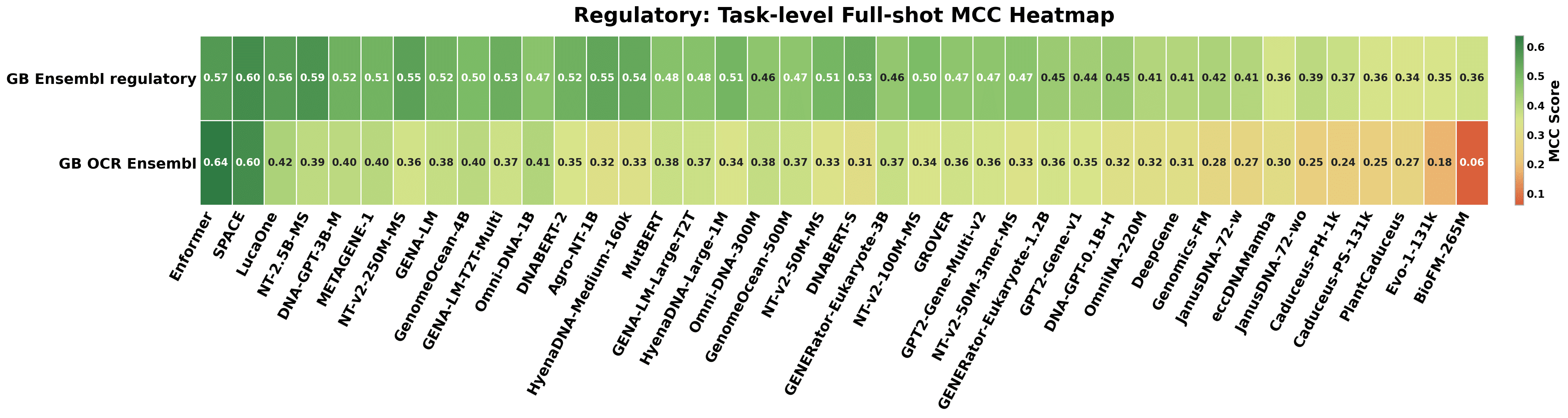}
    \caption{\textbf{Per-task MCC for Regulatory Element
    Prediction.} Heatmap shows full-shot MCC for each of the 40
    genomic foundation models on the 2 regulatory element tasks,
    with models sorted by mean Regulatory MCC. Cell values
    report per-task MCC, with colors ranging from red/orange for
    lower scores to green for higher scores. The
    GB Ensembl regulatory task is easier (mean MCC $=
    0.471$) than GB OCR Ensembl (mean MCC $= 0.344$).
    Human-mouse epigenomic-profile models (\textsc{Enformer},
    \textsc{SPACE}) dominate the top of the model ordering, with
    multi-species models (\textsc{LucaOne}, \textsc{NT-2.5B-MS},
    \textsc{DNA-GPT-3B-M}) following.}
    \label{fig:heatmap_regulatory}
\end{figure}

Regulatory element prediction ($n=2$ tasks: GB Ensembl
regulatory and GB OCR Ensembl) focuses on identifying
genomic regions involved in transcriptional control. This
category reveals one of the largest domain-specific advantages
in our benchmark, with two human-mouse epigenomic-profile models
substantially outperforming all generalists. Results for the 2
regulatory element prediction tasks are presented in
Figures~\ref{fig:kshot_regulatory}--\ref{fig:heatmap_regulatory}.

\textsc{Enformer} and \textsc{SPACE}, pretrained on human-mouse
epigenomic profiles, dominate this category with mean MCC of
$0.604$ and $0.598$ respectively, substantially exceeding all
other models. The next-best performer (\textsc{LucaOne},
multi-species, $2$B, MCC $= 0.492$) trails by over $0.11$ MCC.
This exceptional performance directly reflects the alignment
between pretraining data (regulatory chromatin-state signals
across many cell types) and task requirements (regulatory
element identification).

Scaling on Regulatory is modest but statistically significant
(Spearman $\rho = 0.387$, $p = 0.014$; $\rho = 0.490$, $p = 0.002$ excluding the prokaryotic \textsc{Evo-1-131k}).
Models exceeding 1B parameters average $0.431$ MCC versus $0.392$ for models below 200M ($+0.039$ gap, well below the benchmark-wide $+0.075$). The compressed tier gap reflects the
dominance of medium-scale specialized models
(\textsc{Enformer} at 252M, \textsc{SPACE} at 589M) over larger
general-purpose alternatives, underscoring that
domain-appropriate pretraining trumps scale on this category.

An unusual pattern emerges in pretraining-data comparisons:
\textsc{DNABERT-S} (multi-species microbial) outperforms
\textsc{Genomics-FM} (multi-species, $\approx$120M each,
matched Transformer-encoder) by $0.069$ MCC ($0.419$ vs.\ $0.350$)
-- the only category in our benchmark where this reversal
occurs (in all other 12 categories \textsc{Genomics-FM} leads
this pair). This reversal may indicate that certain regulatory
patterns are shared across prokaryotic and eukaryotic systems,
or that \textsc{DNABERT-S}'s training procedure captures
generalizable regulatory features despite its taxonomic focus;
given $n=2$ tasks and a single pairwise contrast, however, this
remains an observation rather than a strong claim.

\textsc{GENA-LM} shows notable specialization for regulatory
tasks, achieving per-task rank $10.5$ on Regulatory versus $20.0$
across the remaining $98$ tasks (specialization $\Delta = 9.5$).
The specialization is specific to the human-pretrained
\textsc{GENA-LM} base variant: the multi-species
\textsc{GENA-LM-T2T-Multi} shows weaker specialization
($\Delta = 5.8$) and the larger multi-species
\textsc{GENA-LM-Large-T2T} shows negative specialization
($\Delta = -3.2$). The pattern is consistent with the
concentration of well-annotated regulatory elements in human
genomic resources, although with $n=2$ tasks the effect should
be interpreted cautiously.

\newpage

\subsubsection{Virus and Phage Detection}

\begin{figure}[H]
    \centering
    \includegraphics[width=1.0\linewidth]{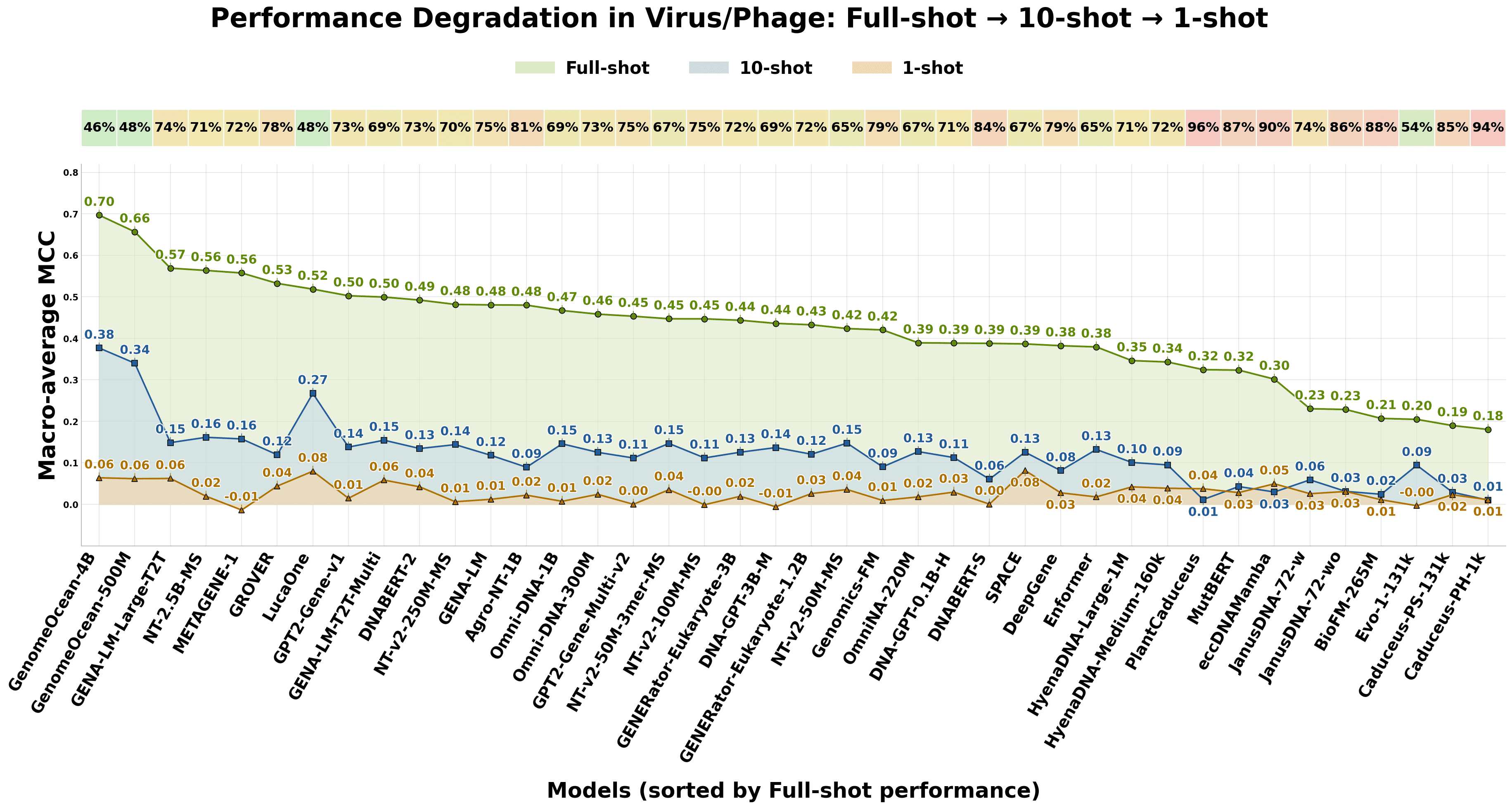}
    \caption{\textbf{Few-shot performance degradation on
    Virus/Phage Detection.} For each of the 40 models,
    macro-average MCC across 2 virus/phage tasks under
    full-shot, 10-shot, and 1-shot regimes; models ordered by
    full-shot performance. The top band shows the relative drop
    from full-shot to 10-shot per model. Benchmark-wide mean
    degradation: $71.3\%$ for 10-shot, $93.5\%$ for 1-shot --
    the largest 1-shot degradation of any task category in our
    benchmark. The 10-shot maximum is $0.377$ for
    \textsc{GenomeOcean-4B}; 1-shot maximum is $0.081$.
    Full-shot and 10-shot rankings remain strongly correlated
    (Spearman $\rho = 0.83$, top-5 overlap $4$ of $5$).}
    \label{fig:kshot_virus}
\end{figure}

\begin{figure}[H]
    \centering
    \includegraphics[height=0.3\textheight]{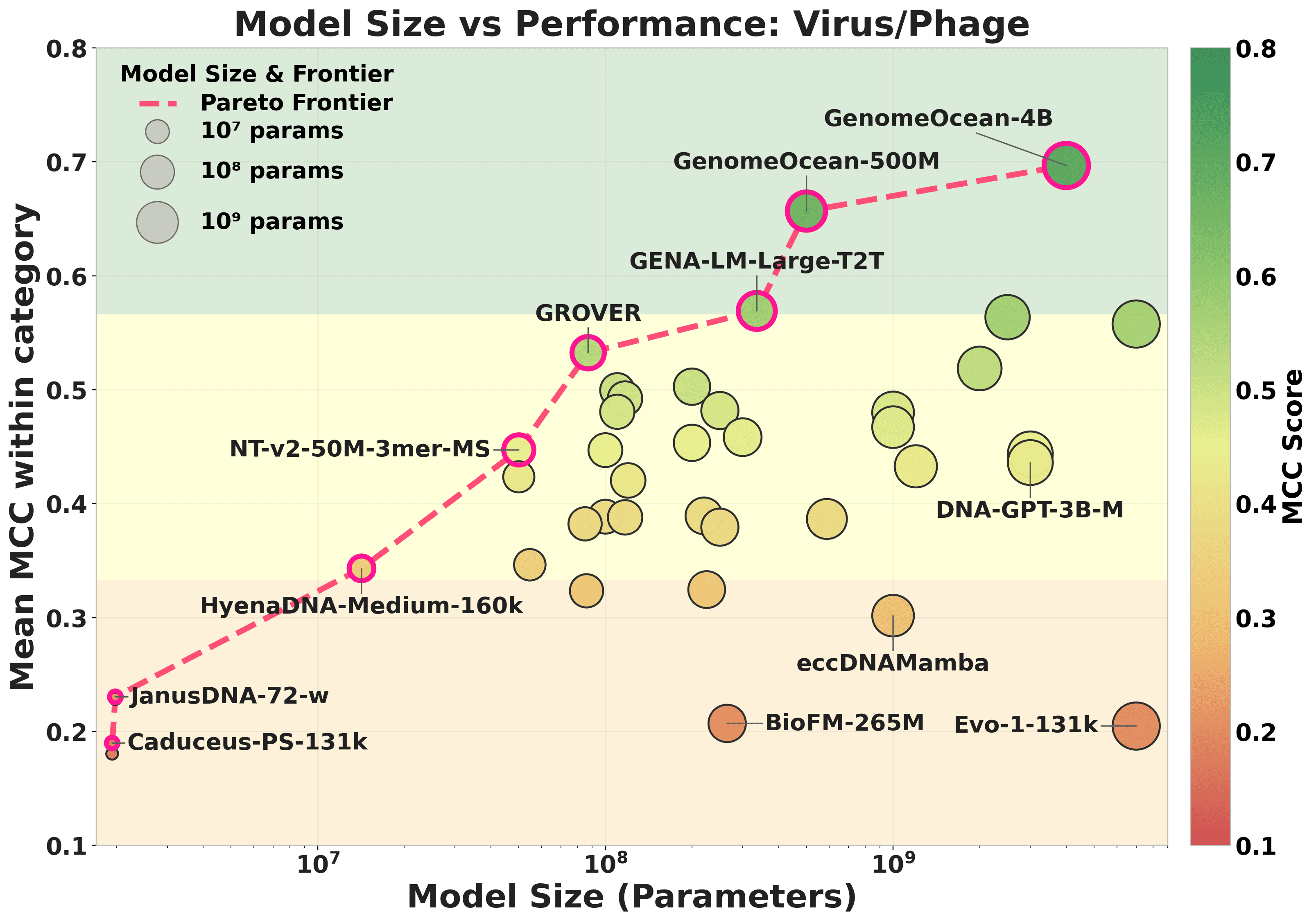}
    \caption{\textbf{Pareto frontier for Virus/Phage Detection:
    mean MCC vs.\ parameter count.}
    Each point represents one of the 40 genomic foundation
    models, with parameter count on a logarithmic x-axis and
    mean full-shot Virus/Phage MCC on the y-axis. Marker size
    and color both encode MCC. The dashed line marks the Pareto
    frontier of best performance--size trade-offs. The
    \textsc{GenomeOcean} multi-species decoders dominate the
    frontier (\textsc{GenomeOcean-4B}, $4$B, MCC $= 0.697$;
    \textsc{GenomeOcean-500M}, $500$M, MCC $= 0.657$).
    \textsc{GROVER} (87M, mean MCC $= 0.532$) is the strongest
    sub-100M model on this category.}
    \label{fig:pareto_virus}
\end{figure}

\begin{figure}[H]
    \centering
    \includegraphics[width=1.0\linewidth]{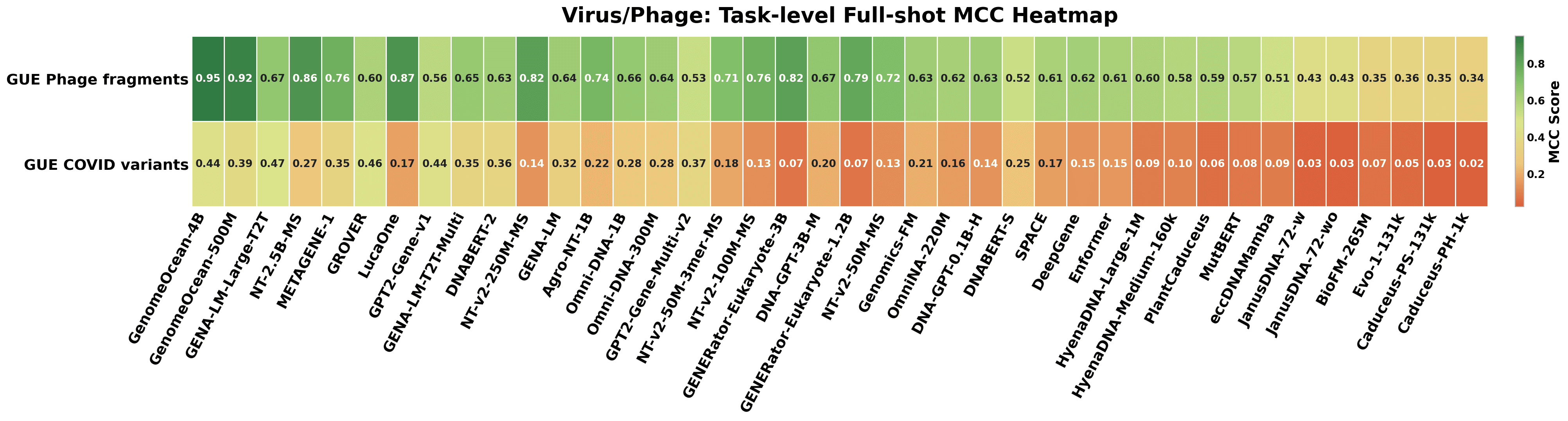}
    \caption{\textbf{Per-task MCC for Virus/Phage Detection.}
    Heatmap shows full-shot MCC for each of the 40 genomic
    foundation models on the 2 virus/phage tasks, with models
    sorted by mean Virus/Phage MCC. Cell values report per-task
    MCC, with colors ranging from red/orange for lower scores
    to green for higher scores. GUE Phage fragments
    (mean MCC $= 0.633$) is substantially easier than
    GUE COVID variants ($0.200$). Multi-species
    decoder models (\textsc{GenomeOcean-4B},
    \textsc{GenomeOcean-500M}) and Transformer-encoder models
    dominate the top of the ordering.}
    \label{fig:heatmap_virus}
\end{figure}

Virus/Phage detection ($n=2$ tasks: GUE Phage fragments
and GUE COVID variants) presents a distinctive
prediction problem involving recognition of viral genomic
signatures. Task difficulty varies dramatically: phage fragment
classification achieves $0.633$ mean MCC while COVID variant
prediction proves challenging at $0.200$ MCC. Results for the
2 virus and phage prediction tasks are presented in
Figures~\ref{fig:kshot_virus}--\ref{fig:heatmap_virus}.

Scaling is statistically significant for this category
(Spearman $\rho = 0.439$, $p = 0.005$), with the
\textsc{GenomeOcean} models achieving strong performance:
\textsc{GenomeOcean-4B} leads the category (MCC $= 0.697$,
rank $1$ of $40$) and \textsc{GenomeOcean-500M} follows at
rank $2$ (MCC $= 0.657$).

A distinctive pattern emerges for this category: human-trained
models show competitive or superior performance compared to
multi-species alternatives. \textsc{GROVER} (human) exceeds \textsc{DNABERT-2} (multi-species) by $0.040$ MCC under matched Transformer-encoder/BPE conditions. This pattern
likely reflects the predominance of human-associated viral
sequences in human genomic training data, providing relevant
exposure to viral integration sites and endogenous retroviral
elements.

Architecture comparisons reveal large Transformer advantages.
\textsc{GenomeOcean-500M} (Transformer-decoder, 500M) exceeds
\textsc{eccDNAMamba} (Mamba, 537M) by $0.355$ MCC ($0.657$
vs.\ $0.302$) at near-matched scale (1.07$\times$ ratio) --
the largest architectural gap observed in our benchmark. \textsc{GENA-LM-Large-T2T}
(Transformer-encoder) outperforms \textsc{OmniNA-220M}
(Transformer-decoder) by $0.180$ MCC ($0.569$ vs.\ $0.389$).

Tokenization effects show BPE advantages over single-nucleotide
approaches, reversing patterns observed for splice sites.
\textsc{GROVER} (BPE) exceeds \textsc{MutBERT}
(single-nucleotide) by $0.209$ MCC ($0.532$ vs.\ $0.323$), and
\textsc{GENA-LM} (BPE) shows similar advantage ($+0.157$ MCC).
This pattern suggests that viral sequence recognition benefits
from the longer-range patterns captured by subword tokenization.

Several models show specialization toward virus/phage detection.
\textsc{Agro-NT-1B} achieves per-task rank $12.5$ on Virus/Phage
versus $20.5$ across the remaining $98$ tasks (specialization
$\Delta = 8.0$), potentially reflecting viral sequences embedded
in plant genomic training data. \textsc{GROVER} ($\Delta = 6.5$)
and \textsc{GENA-LM} ($\Delta = 6.0$) similarly specialize, both
being human-pretrained Transformer-encoders. The
multi-species \textsc{GENA-LM-T2T-Multi} and
\textsc{GENA-LM-Large-T2T} show smaller positive specialization
($\Delta = 5.3$ and $5.4$ respectively); the human-only direction
remains modest within the GENA-LM family.

Few-shot degradation is severe for this category ($93.5\%$ at
1-shot, $71.3\%$ at 10-shot). \textsc{GenomeOcean-4B} and
\textsc{GenomeOcean-500M} show the steepest 1-shot absolute
declines ($0.633$ and $0.595$ respectively) yet retain their
full-shot ranking (top-2 at both regimes). Their 10-shot
relative drops ($46\%$ and $48\%$) are well below the category
mean of $71.3\%$, indicating that despite catastrophic 1-shot
collapse these models retain operationally useful 10-shot
performance. Some rank shifts emerge among the runners-up:
\textsc{LucaOne} rises from full-shot rank $7$ to 10-shot rank
$3$, while \textsc{GROVER} drops from $6$ to $21$.

\newpage

\subsubsection{Coding versus Non-coding Classification}

\begin{figure}[h]
    \centering
    \includegraphics[width=1.0\linewidth]{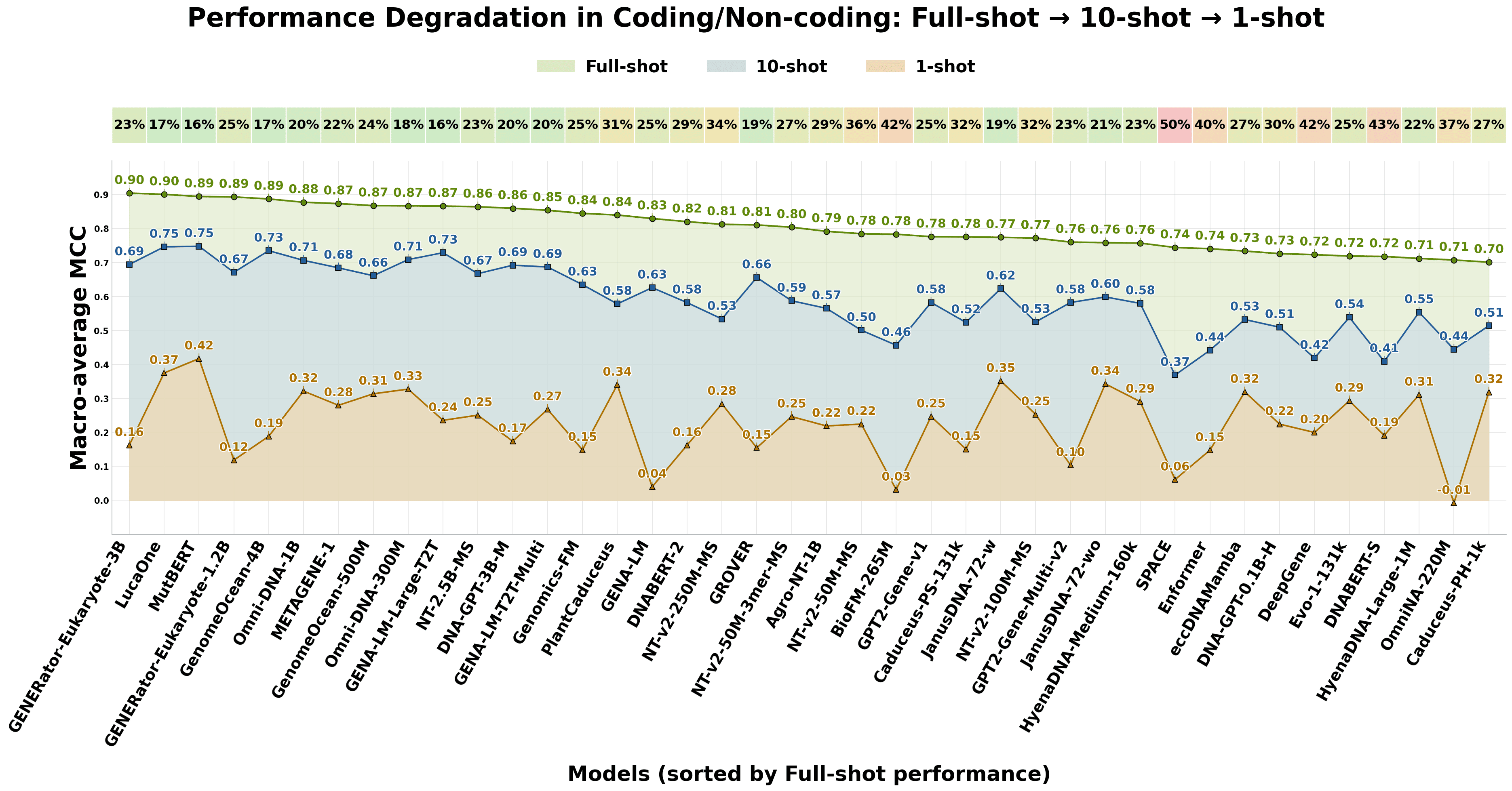}
    \caption{\textbf{Few-shot performance degradation on
    Coding/Non-coding Classification.} For each of the 40 models,
    MCC on the coding vs.\ non-coding task under full-shot,
    10-shot, and 1-shot regimes; models ordered by full-shot
    performance. The top band shows the relative drop from
    full-shot to 10-shot per model. Benchmark-wide mean
    degradation: $26.6\%$ for 10-shot, $71.7\%$ for 1-shot --
    among the mildest 10-shot degradation observed across task
    categories. \textsc{MutBERT} (86M, human, encoder) achieves
    the highest 10-shot MCC ($0.748$) and the highest 1-shot
    MCC ($0.417$), rising from full-shot rank $3$ to 10-shot
    rank $1$. Full-shot and 10-shot rankings remain strongly
    correlated (Spearman $\rho = 0.86$, top-5 overlap $3$ of $5$).}
    \label{fig:kshot_coding}
\end{figure}

\begin{figure}[h]
    \centering
    \includegraphics[height=0.3\textheight]{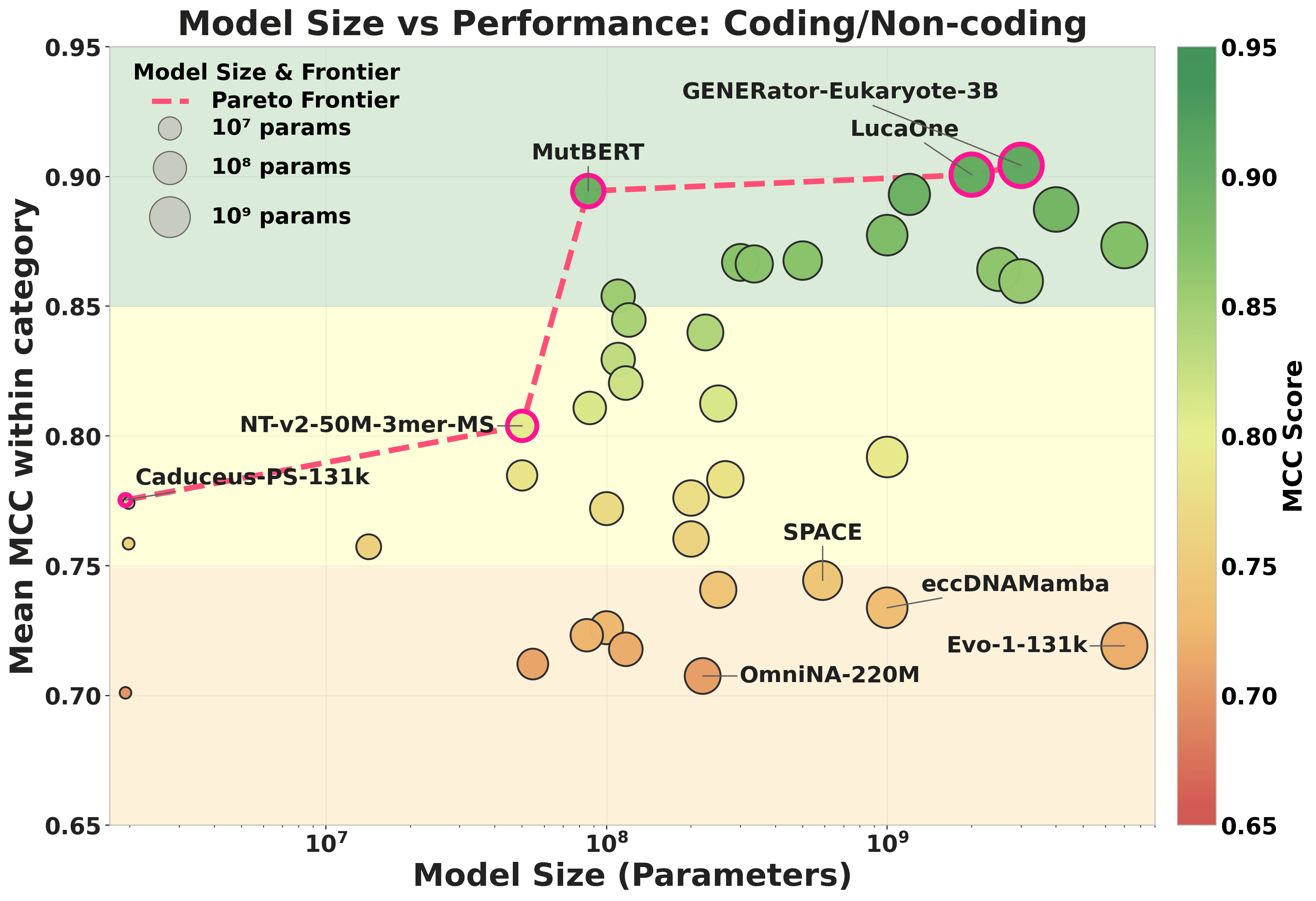}
    \caption{\textbf{Pareto frontier for Coding/Non-coding
    Classification: MCC vs.\ parameter count.}
    Each point represents one of the 40 genomic foundation
    models, with parameter count on a logarithmic x-axis and
    full-shot MCC on the y-axis. Marker size and color both
    encode MCC. The dashed line marks the Pareto frontier of
    best performance--size trade-offs. Eukaryotic gene-focused
    \textsc{GENERator-Eukaryote-3B} (3B, MCC $= 0.904$) leads,
    with multi-species \textsc{LucaOne} (2B, MCC $= 0.901$)
    close behind. \textsc{MutBERT} (86M, MCC $= 0.894$) is the
    strongest sub-100M model on this category and ranks $3$rd
    of $40$ overall, demonstrating that for this task small
    well-designed models remain highly competitive with much
    larger generalists.}
    \label{fig:pareto_coding}
\end{figure}

\begin{figure}[h]
    \centering
    \includegraphics[width=1.0\linewidth]{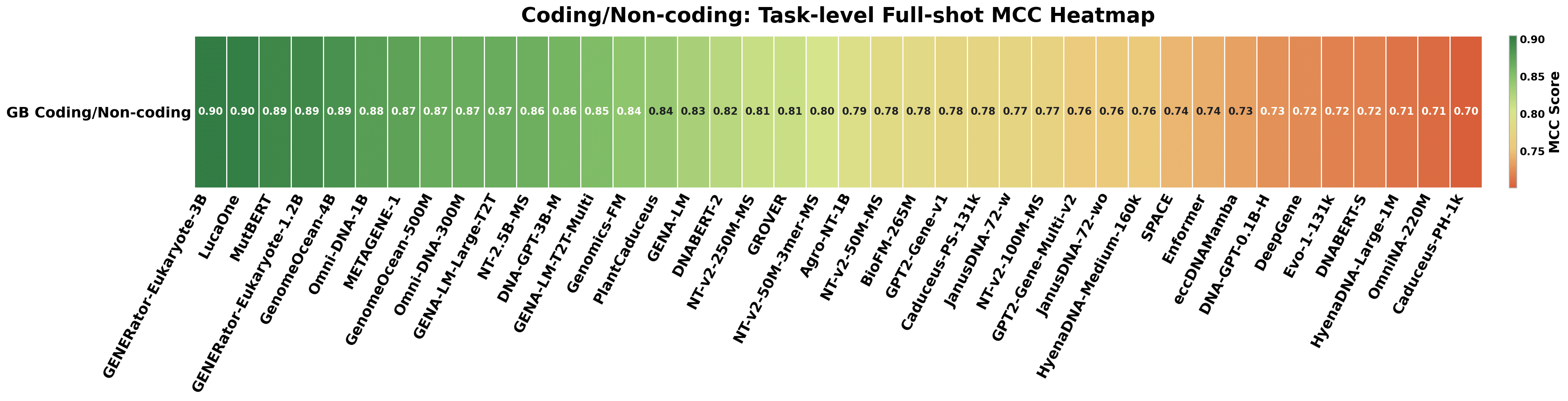}
    \caption{\textbf{Per-model MCC for Coding/Non-coding
    Classification.} Heatmap shows full-shot MCC for each of the
    40 genomic foundation models on the single coding vs.\
    non-coding task, with models sorted by MCC. Cell values
    report per-model MCC, with colors ranging from red/orange
    for lower scores to green for higher scores. Multi-species
    decoder models (\textsc{GENERator-Eukaryote-3B},
    \textsc{LucaOne}), Transformer-encoder models
    (\textsc{MutBERT}), and large multi-species generalists
    cluster at the top of the model ordering.}
    \label{fig:heatmap_coding}
\end{figure}

Coding/Non-coding classification ($n=1$ task: GB
Coding/Non-coding) represents a fundamental sequence annotation
problem with relatively high baseline performance (mean MCC $=
0.803$ across the 40 models). This task evaluates whether models
capture the statistical signatures distinguishing protein-coding
from non-coding sequences. Results for this task are presented
in Figures~\ref{fig:kshot_coding}--\ref{fig:heatmap_coding}.

Coding/Non-coding classification shows significant positive
scaling (Spearman $\rho = 0.486$, $p = 0.002$; $\rho = 0.577$, $p < 0.001$ excluding the prokaryotic \textsc{Evo-1-131k}).
Models exceeding 1B parameters achieve $0.857$ mean MCC compared to $0.781$ for models below 200M, a $+0.076$ gap that matches the benchmark-wide tier difference.

\textsc{GENERator-Eukaryote-3B} leads the category (MCC $=
0.904$), followed closely by \textsc{LucaOne} (MCC $= 0.901$)
and \textsc{MutBERT} (MCC $= 0.894$). The strong performance of
\textsc{MutBERT} (86M parameters, rank $3$ of $40$) demonstrates
that even on a task that responds to scale, well-designed
smaller models remain competitive with multi-billion-parameter
counterparts.

Architecture comparisons favor Transformers on this category.
\textsc{MutBERT} (Transformer-encoder) exceeds
\textsc{HyenaDNA-Large-1M} (Hyena) by $0.182$ MCC ($0.894$
vs.\ $0.712$) under matched human-pretraining and
single-nucleotide tokenization, although the contrast is not
strictly matched on scale (86M vs.\ 55M). \textsc{GenomeOcean-500M} (Transformer-decoder, 500M)
outperforms \textsc{eccDNAMamba} (Mamba, 537M) by $0.134$
MCC ($0.868$ vs.\ $0.734$) at near-matched scale under matched
multi-species/BPE conditions.

Pretraining data comparisons show advantages for both eukaryotic
gene-focused and multi-species approaches over microbial
training. \textsc{Genomics-FM} (multi-species) exceeds
\textsc{DNABERT-S} (multi-species-microbial) by $0.127$ MCC
($0.845$ vs.\ $0.718$) under matched Transformer-encoder
conditions ($\approx$120M each), while
\textsc{GENERator-Eukaryote-3B} (eukaryotic-gene) outperforms
\textsc{DNA-GPT-3B-M} (multi-species) by $0.045$ MCC under
matched Transformer-decoder/$k$-mer/3B conditions.

Notably, \textsc{Evo-1-131k} (prokaryotic pretraining) achieves
acceptable performance on this task (MCC $= 0.719$, rank $36$
of $40$), substantially better than its catastrophic results on
Splice Sites (MCC $= 0.160$) or TF Binding (MCC $= 0.173$). This
relative success may reflect the more universal nature of
coding-sequence signatures across prokaryotic and eukaryotic
genomes, where codon usage patterns and open-reading-frame
statistics share fundamental properties.

\newpage

\subsubsection{Chromatin Accessibility}

\begin{figure}[h]
    \centering
    \includegraphics[width=1.0\linewidth]{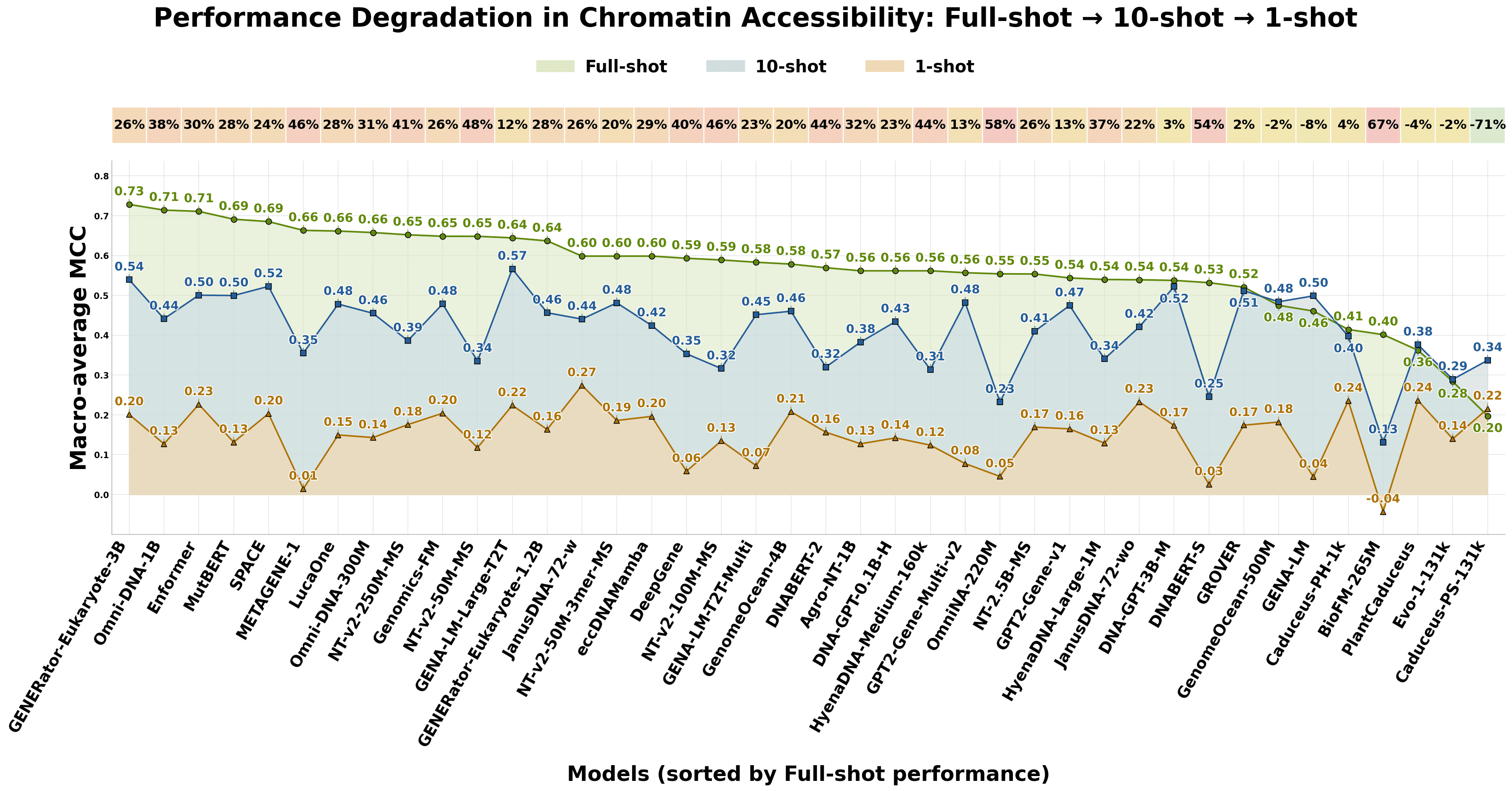}
    \caption{\textbf{Few-shot performance degradation on
    Chromatin Accessibility.} For each of the 40 models, MCC on
    the iDHS DNase-I task under full-shot, 10-shot, and 1-shot
    regimes; models ordered by full-shot performance. The top
    band shows the relative drop from full-shot to 10-shot per
    model. Benchmark-wide mean degradation: $26.8\%$ for 10-shot,
    $73.6\%$ for 1-shot -- among the mildest 10-shot
    degradation observed across task categories. The 10-shot and
    1-shot rankings reshuffle substantially relative to full-shot
    (Spearman $\rho = 0.38$, top-5 overlap $2$ of $5$):
    \textsc{GENA-LM-Large-T2T} rises from full-shot rank $12$ to
    10-shot rank $1$ (MCC $= 0.567$), and \textsc{JanusDNA-72-w}
    (2M parameters) achieves the highest 1-shot MCC ($0.274$).}
    \label{fig:kshot_chromatin}
\end{figure}

\begin{figure}[H]
    \centering
    \includegraphics[height=0.3\textheight]{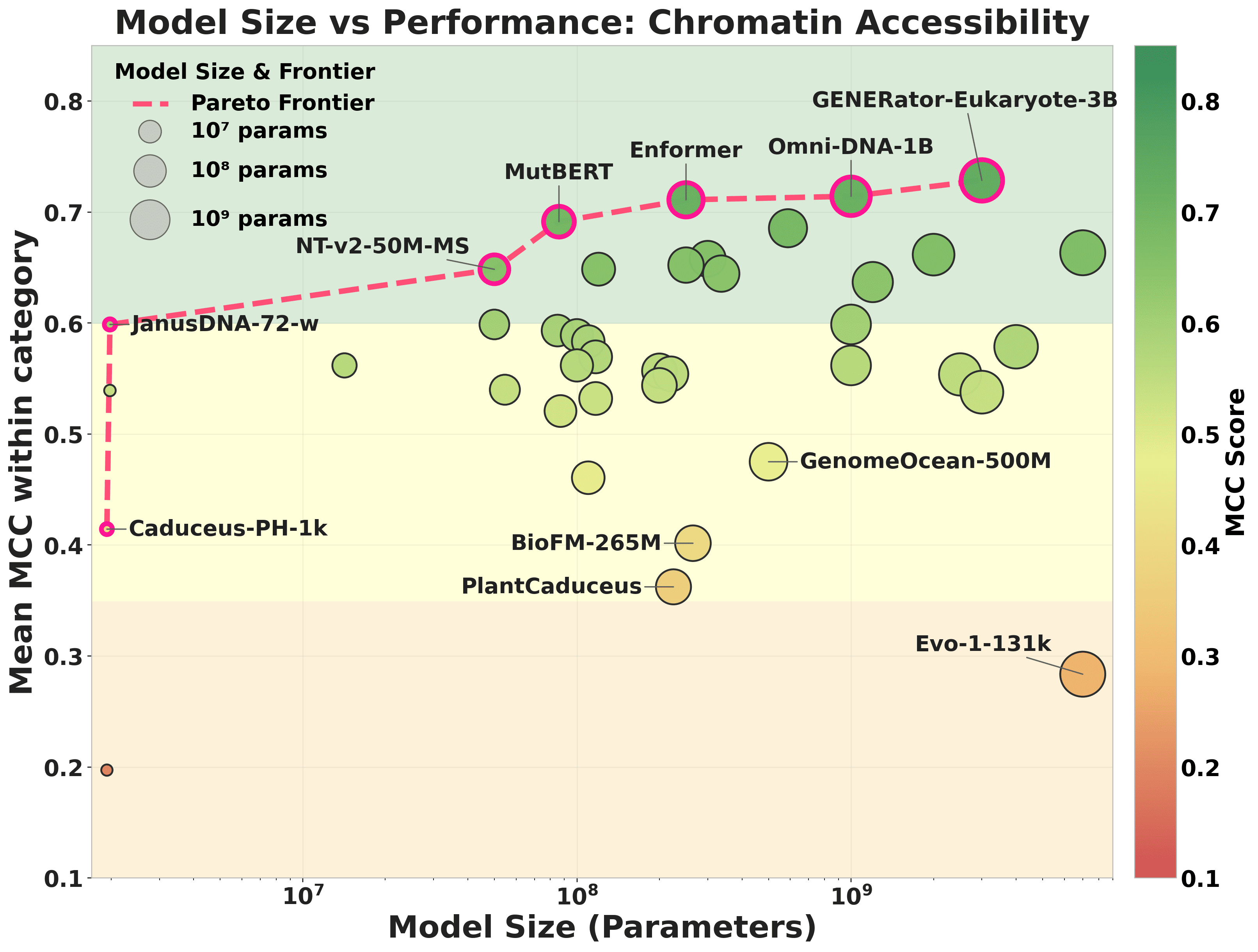}
    \caption{\textbf{Pareto frontier for Chromatin Accessibility:
    MCC vs.\ parameter count.}
    Each point represents one of the 40 genomic foundation
    models, with parameter count on a logarithmic x-axis and
    full-shot MCC on the y-axis. Marker size and color both
    encode MCC. The dashed line marks the Pareto frontier of
    best performance--size trade-offs. \textsc{GENERator-Eukaryote-3B}
    (3B, MCC $= 0.728$) leads, with \textsc{MutBERT} (86M, MCC $=
    0.691$, rank $4$ of $40$) the strongest sub-100M
    Transformer-encoder. Strikingly, \textsc{JanusDNA-72-w}
    (2M, MCC $= 0.599$) sits prominently on the frontier at
    rank $14$ of $40$ -- the smallest model in the benchmark
    outperforming many counterparts that are over $500\times$
    larger.}
    \label{fig:pareto_chromatin}
\end{figure}

\begin{figure}[H]
    \centering
    \includegraphics[width=1.0\linewidth]{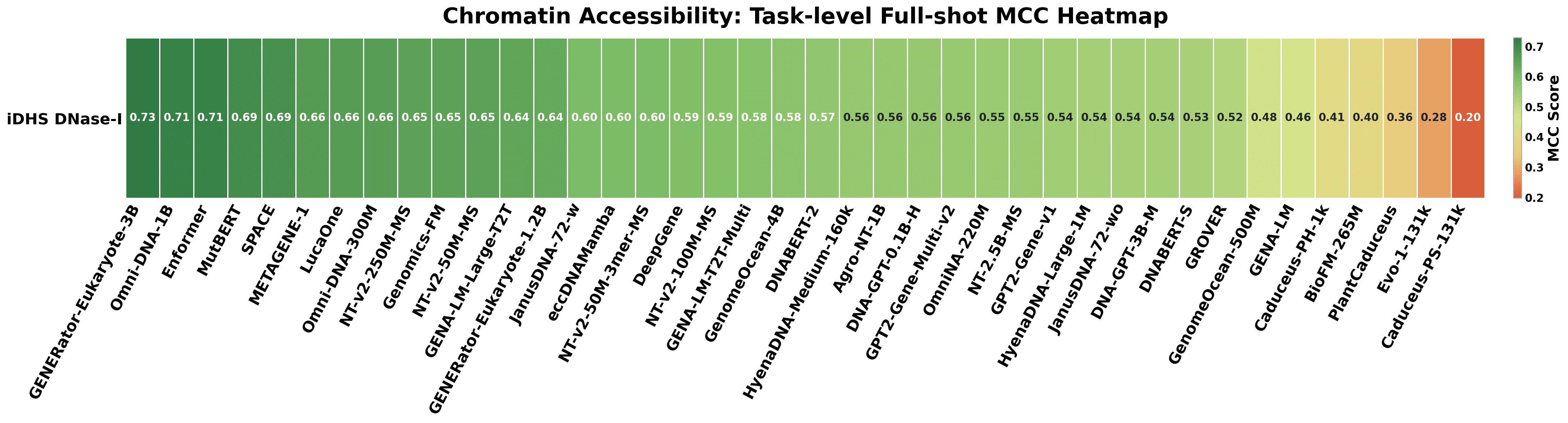}
    \caption{\textbf{Per-model MCC for Chromatin Accessibility.}
    Heatmap shows full-shot MCC for each of the 40 genomic
    foundation models on the single iDHS DNase-I task, with
    models sorted by MCC. Cell values report per-model MCC, with
    colors ranging from red/orange for lower scores to green for
    higher scores. Eukaryotic gene-focused
    (\textsc{GENERator-Eukaryote-3B}), multi-species decoders
    (\textsc{Omni-DNA-1B}), and human-mouse epigenomic-profile
    models (\textsc{Enformer}, \textsc{SPACE}) cluster at the top
    of the model ordering.}
    \label{fig:heatmap_chromatin}
\end{figure}

Chromatin Accessibility prediction ($n=1$ task: iDHS
DNase-I) evaluates recognition of open chromatin regions, a key
determinant of transcriptional potential. This category shows
distinctive patterns in both architecture and tokenization
comparisons, and exhibits the weakest scaling effect of any
category in our benchmark. Results for this task are presented
in Figures~\ref{fig:kshot_chromatin}--\ref{fig:heatmap_chromatin}.

Scaling effects are non-significant on this category (Spearman
$\rho = 0.240$, $p = 0.136$; $\rho = 0.327$, $p = 0.042$
excluding the prokaryotic \textsc{Evo-1-131k}) -- the weakest
scaling relationship of any category in our benchmark. Models
exceeding 1B parameters achieve $0.592$ mean MCC versus $0.547$ for models below 200M (a modest $+0.045$ gap, well below the benchmark-wide $+0.075$). \textsc{GENERator-Eukaryote-3B} leads
the category (MCC $= 0.728$), followed by \textsc{Omni-DNA-1B}
(MCC $= 0.714$) and \textsc{Enformer} (MCC $= 0.711$).

Pretraining-data effects are pronounced. The controlled
\textsc{GENA-LM} comparison shows multi-species exceeding
human-only by $0.123$ MCC (\textsc{GENA-LM-T2T-Multi} $= 0.583$
vs.\ \textsc{GENA-LM} $= 0.461$) -- the largest gap observed
for this model pair across all $13$ task categories (next-largest
is Mouse Enhancers at $0.068$). Eukaryotic gene-focused
\textsc{GENERator-Eukaryote-3B} outperforms multi-species
\textsc{DNA-GPT-3B-M} by $0.191$ MCC ($0.728$ vs.\ $0.538$)
under matched 3B/decoder/$k$-mer conditions -- also the largest
such gap for this controlled pair across the benchmark.

Tokenization comparisons reveal dramatic single-nucleotide
advantages. \textsc{MutBERT} (single-nucleotide) exceeds
\textsc{GENA-LM} (BPE) by $0.231$ MCC ($0.691$ vs.\ $0.461$)
under matched Transformer-encoder/human-pretraining conditions
(86M vs.\ 110M) -- the largest tokenization gap observed for
this controlled pair across the benchmark. \textsc{MutBERT}
similarly outperforms \textsc{GROVER} (BPE) by $0.171$ MCC
($0.691$ vs.\ $0.521$) under matched conditions
($\approx$86--87M each). This substantial advantage suggests
that chromatin accessibility prediction benefits from
fine-grained positional information that coarser tokenization
schemes may obscure.

Architecture comparisons reveal an atypical pattern for this
category. Under matched conditions at near-identical scale
(\textsc{GenomeOcean-500M}, Transformer-decoder, 500M vs.\
\textsc{eccDNAMamba}, Mamba, 537M; $1.07\times$ ratio,
multi-species, BPE), \textsc{eccDNAMamba} outperforms
\textsc{GenomeOcean-500M} by $0.124$ MCC ($0.599$ vs.\ $0.475$)
-- a genuine architectural reversal at matched scale, in
contrast to the Transformer-over-Mamba advantage observed on
every other category. The reversal disappears at larger
Transformer scale: \textsc{Omni-DNA-1B} (1B) exceeds
\textsc{eccDNAMamba} by $0.116$ MCC, indicating that the Mamba
advantage on chromatin accessibility is scale-dependent rather
than a uniform architectural property. Additionally,
\textsc{DeepGene} (Graph-Transformer, 85M) exceeds
\textsc{GENA-LM} (Transformer-encoder, 110M) by $0.133$ MCC.
These atypical patterns suggest that chromatin accessibility
prediction may benefit from architectural features beyond
standard attention mechanisms.

Model specialization is striking for this category. The smallest
model in our benchmark, \textsc{JanusDNA-72-w} (Hybrid-Mamba-MoE,
2M parameters), achieves per-task rank $15.0$ on Chromatin
Accessibility versus $32.9$ across the remaining $99$ tasks
(specialization $\Delta = 17.9$) -- this Mamba/MoE hybrid at
near-toy scale outperforms most multi-billion-parameter
generalists on this task. \textsc{eccDNAMamba} (537M, Mamba)
shows a similar pattern ($\Delta = 16.5$). The relative strength
of these two architectures on chromatin accessibility, despite
their generally weaker overall performance, suggests that
state-space components may capture specific sequence features
relevant to chromatin state prediction; however, other
Mamba-based models in our benchmark (\textsc{Caduceus-PH-1k},
\textsc{PlantCaduceus}, \textsc{Caduceus-PS-131k}) do not show
this specialization, so the effect is not a uniform property of
the Mamba family.

\newpage

\subsection{Task Difficulty and Model Differentiation}
\label{subsec:task_difficulty}

The 100 tasks in our benchmark span a wide difficulty spectrum,
from near-solved problems to challenges that remain largely
intractable for current foundation models. Understanding this
difficulty landscape provides essential context for interpreting
model comparisons and identifying promising directions for future
research.

\subsubsection{Easy Tasks: Approaching Ceiling Performance}

We identify $18$ tasks achieving mean MCC exceeding $0.70$
across all $40$ models, indicating problems where current
approaches converge to robust solutions. Human cell-type-specific
promoter recognition tasks dominate this category: iPro HUVEC
achieves $0.890$ mean MCC, iPro HeLa-S3 reaches $0.875$, and
iPro GM12878 attains $0.855$. The \textsc{SPACE} model achieves
best performance on all three tasks, while \textsc{BioFM-265M}
consistently ranks lowest, highlighting the importance of
appropriate pretraining even for ostensibly solved problems.

General promoter classification tasks (NT Promoter (all), NT
Promoter (no TATA), GUE Prom 300 (no TATA)) cluster at
$0.853$--$0.855$ mean MCC, with \textsc{GENA-LM-Large-T2T} and
\textsc{Omni-DNA-1B} among the top performers. The GB
Human-or-worm species classification task reaches $0.857$ mean
MCC, with \textsc{MutBERT} leading (MCC $= 0.948$) and
\textsc{Evo-1-131k} performing worst -- a pattern reflecting the
prokaryotic model's incompatibility with eukaryotic
discrimination tasks.

The GB Coding/Non-coding task achieves $0.803$ mean MCC,
representing a relatively accessible problem where sequence
composition differences between coding and non-coding regions
provide strong discriminative signal.
\textsc{GENERator-Eukaryote-3B} leads this task, while
\textsc{Caduceus-PH-1k} shows weakest performance.

\subsubsection{Hard Tasks: Persistent Challenges}

We identify $28$ tasks with mean MCC below $0.35$, representing
problems where even state-of-the-art models achieve limited
success. DNA methylation prediction dominates this category,
with 4mC \emph{G.\ subterraneus} achieving only $0.061$ mean MCC
(maximum $0.206$ by \textsc{GENERator-Eukaryote-3B}). 4mC
\emph{E.\ coli} reaches $0.103$ mean MCC and 4mC
\emph{G.\ pickeringii} achieves $0.107$. These results suggest
that current sequence-based approaches struggle to capture the
contextual and enzymatic factors governing methylation site
selection.

Plant lncRNA classification tasks prove similarly challenging,
with PGB lncRNA \emph{S.\ lycopersicum} (tomato) at $0.221$ mean
MCC, PGB lncRNA \emph{G.\ max} (soybean) at $0.228$, and PGB
lncRNA \emph{T.\ aestivum} (wheat) at $0.238$. \textsc{LucaOne}
achieves best performance across all $6$ plant lncRNA tasks,
reaching $0.539$ on \emph{S.\ lycopersicum}, demonstrating that
sufficiently diverse multi-species pretraining can partially
address plant-specific challenges even without dedicated plant
data.

Viral sequence classification presents unexpected difficulty,
with the GUE COVID variants task achieving only $0.200$ mean
MCC. \textsc{GENA-LM-Large-T2T} leads at $0.472$, suggesting
that multi-species genomic pretraining provides some
transferable signal for viral sequence analysis despite the
substantial evolutionary distance.

\subsubsection{High-Variance Tasks: Discriminating Model
Capabilities}

Tasks exhibiting high inter-model variance (standard deviation
$> 0.12$) reveal where architectural and pretraining choices
most strongly differentiate performance. We identify $13$ such
``controversial'' tasks that serve as natural stress tests for
model capabilities.

The GUE Fungi-20 task shows the highest variance (std $=
0.216$, range $= 0.764$ MCC), with \textsc{GenomeOcean-4B}
achieving $0.939$ MCC while \textsc{Caduceus-PH-1k} reaches only
$0.175$. The top performers uniformly employ multi-species or
eukaryotic-gene pretraining (\textsc{GenomeOcean-4B},
\textsc{GENERator-Eukaryote-3B},
\textsc{GENERator-Eukaryote-1.2B}), while bottom performers use
narrower data (\textsc{Caduceus-PH-1k}, \textsc{BioFM-265M},
\textsc{JanusDNA-72-wo}).

Splice site detection tasks exhibit consistently high variance:
NT Splice donors (std $= 0.195$, range $= 0.666$), NT Splice
acceptors (std $= 0.186$, range $= 0.643$), and GUE Splice
reconstr.\ (std $= 0.170$, range $= 0.597$). \textsc{NT-2.5B-MS},
\textsc{LucaOne}, and \textsc{GENERator-Eukaryote-3B}
consistently rank among the top performers across splice tasks,
while \textsc{Evo-1-131k} consistently ranks last -- achieving
negative MCC on GUE Splice reconstr., indicating
worse-than-random performance attributable to prokaryotic
pretraining's incompatibility with spliceosomal machinery.

Mouse enhancer tasks show substantial variance (std $=
0.151$--$0.175$), with \textsc{Enformer} and \textsc{SPACE}
(human-mouse epigenomic profiles) dominating while human-only
trained \textsc{JanusDNA} variants and \textsc{Caduceus} variants
perform poorly. This pattern supports the importance of
taxonomically aligned pretraining for cross-species regulatory
element prediction.

\subsubsection{Patterns in High-Variance Task Performance}

Systematic analysis of top-3 and bottom-3 performers across all
$13$ high-variance tasks reveals striking patterns in
architecture and pretraining data. Among top performers,
Transformer-decoder appears $18$ times and Transformer-encoder
$15$ times, with CNN-Transformer architectures
(\textsc{Enformer}, \textsc{SPACE}) appearing $6$ times. Among
bottom performers, Mamba architectures appear $17$ times,
Hybrid-Mamba-MoE $7$ times, and StripedHyena $6$ times. This
distribution suggests that attention-based architectures
substantially outperform state-space alternatives on the most
discriminating benchmark tasks.

Pretraining data patterns are even more pronounced. Multi-species
training appears in top-3 positions $20$ times and
eukaryotic-gene training $12$ times, while human-mouse-profiles
appears $6$ times. In contrast, human-only training appears in
bottom-3 positions $29$ times -- more than all other categories
combined. Prokaryotic training appears $6$ times in bottom
positions despite comprising only one model
(\textsc{Evo-1-131k}), indicating consistent failure across
high-variance tasks. These patterns provide strong empirical
support for prioritizing taxonomically diverse pretraining over
species-specific approaches.

\end{document}